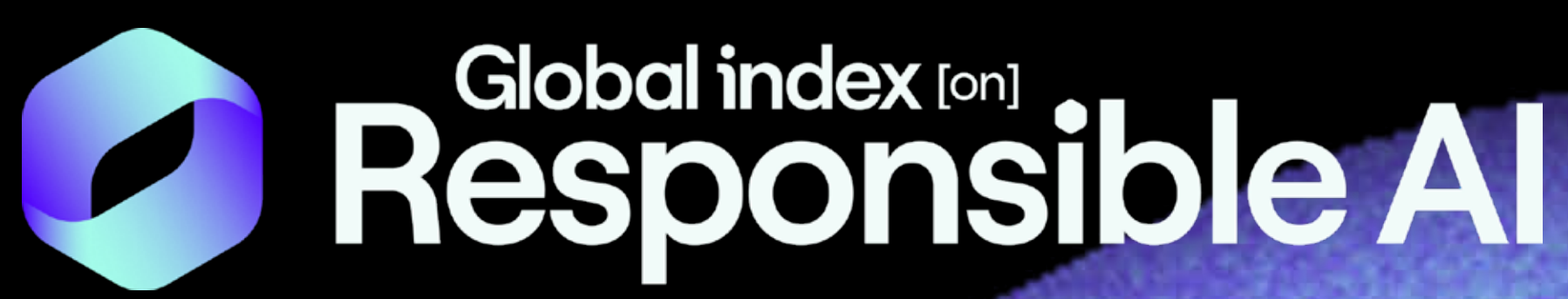

# 2026

2ND EDITION

# Acknowledgements

**Authors**
Rachel Adams
Fola Adeleke
Ayantola Alayande
Selamawit Engida Abdella
Ana Florido
Nicolás Grossman
Leah Junck

**Collaborating Organisations**
- Balkan Investigative Reporting Regional Network (BIRN)
- Caribbean Open Institute, Jamaica (COI)
- East-West Management Institute (EWMI)
- Iniciativa Latinoamericano por los Datos Abiertos, Uruguay (ILDA)
- Local Development Research Institute, Kenya (LDRI)
- LIRNEAsia, Sri Lanka
- MENA Observatory on Responsible AI, Access to Knowledge for Development Center, (A2K4D), Onsi Sawiris School of Business, American University in Cairo (AUC)

The authors would like to thank the support provided by a special group of international experts on AI Governance with whom the GIRAI team built space for discussion, feedback and methodological guidance, while final decisions on the methodology and Index remained the responsibility of the GIRAI team:

- Dr. Aubra Anthony, Teaching Assistant Professor, University of Tennessee
- Dr. Chinasa T. Okolo, Founder and Scientific Director, Technecultura
- Jennifer Louie, AI Trust and Safety Expert, United Nations Development Programme (UNDP)
- Prof. David Leslie, Director of Ethics and Responsible Innovation Research, Public Policy Programme, The Alan Turing Institute
- Prof. Matt Jones, Chief Operating Officer, Responsible AI UK
- Dr. Paola Ricaurte Quijano, Professor and Senior Researcher, Tecnológico de Monterrey
- Prof. Yuval Shany, Accelerator Fellow, Institute for Ethics in AI, University of Oxford

The authors also wish to thank the following people who generously reviewed this report and provided valuable feedback throughout the different stages of the project: Matthew Smitt, Abbey Ghandi, Paula Martins, and Becky Faith. The authors also thank Daniela Benavente, for her methodological support, statistical review of the Index, and generous advice on framing and communicating the findings.

This work was carried out with the aid of a grant from the International Development Research Centre (IDRC), Ottawa, Canada and the Foreign, Commonwealth & Development Office (FCDO) under the AI4Development funding programme. Some further funds were used for the project from the Government of Canada and from the Interamerican Development Bank.

## How to Cite This Report

**The authors wish to thank the following people for their invaluable contributions to the 2nd Edition of the Global Index on Responsible AI — as country-level researchers, coordinators, global reviewers, data analysts, and technical support staff, and as Global Center on AI Governance colleagues and partners.**

Abed Khooli
Adina Florea
Agata Jałosińska
Agustina Brizio
Ahmad AlhajDaoud
Aleksandar Kostadinov
Amadej Petan
Anastasiia Klimashevskaia
Antonia Bogado Rodas
Archana Atmakuri
Artem Kobrin
Aslam Hayat
Asylai Akisheva
Ayantola Alayande
Bewindin Alfred Sawadogo
Bukola Adesokan
Carlos Guerrero Argote
Carolyn Christian
Catherine Setiawan
Chanduni Senara Bandara
Chelsea L. Newhouse
Chinenye Ifebirinachi
Christy Tsang
Clea Strydom
Cosmas Gabin Mbarga Asseng
Cristian León
Dafina Olluri
Daniela Benavente
Daphine Nkunda
Darius Torabi
David Zamora
Duje Prkut
Edgar Brutyan
Eduardo Bejar
Eftichia Soufleri
Erika Marquez
Erjona Huruglica
Eugene Arnaud Yombo Sembe
Fabio Gualandri
Farah Elbehairy
Feng Gao
Fiona Chawana
Gadi Perl
Galina Stoeva
Galiya Yelubayeva
Guya Joel
Herman Ramos
Hugo Miranda
Ilenius Ildephonce
Imane Hmiddou
Irina Buzu
Iris Palma
Isi idemudia
Ivana Jeremic
Jayeola Gbenga
Johannes Bhanye
Jompon Pitaksantayothin
Jordan Seimu
Jorge Ortiz Claverol
Juan David Gutiérrez Rodríguez
Keziah Kithei Munyao
Khadija Aftab
Khadija Imranova
Kunda Nyirongo
Landry Claude Iradukunda
Larissa Galdino de Magalhães Santos
Leena Azzam
Leonardo Stoll De Morais
Lester Ramírez Irías
Lía Hernández
Lila Rao Graham
Lisa Ann Vasciannie
Lucia Cizmaziova
Maathangi Mohan
Madina Tursunova
Maha Jouini
Manase Kudzai Chiweshe
Marcel Mir Teijeiro
Mariam Ayvazyan
Marie Batista Marina Mijuskovic
Marjana Prifti Skenduli
Mark Gaffley
Marwa Soudi
Mary Grace Mirandilla-Santos
Matías Jackson
Matías Valderrama Barragán
Melanie George
Michelle Allen
Mohamed Gele
Muslimbek Buriev
Muhammad Aimal Marjan
Nada Alwadi
Nadine Weheba
Nagham ElHoussamy
Nassibou Bassongui
Nicolas Gholam
Nicole Angel Sánchez Rojas
Nicole Sanchez
Nissily Mushani
Nour Naim
Nowzin Khan
Paola Ricaurte Quijano
Patricia Boshe
Patricio Julián Feldman
Payal Padhy
Peterking Quaye
Pilar Fiuza Casais
Priscilla Ruiz
Guillén
Prof. Dr. Martin Ebers
Rachel Kuo Jing Ying
Rehema Baguma
Rustam Mukhamedov
Ruwangi Fernando
Sally Kuria
Samah Kareem
Samik Kharel
Samuel Segun
Sebastián Hurtado Guevara
Shady Elbassuoni
Sharon Besiwa Essilfie
Sheba C. Okwany
Shurland George
Shyline Muthoni
Steven Akomain
Suraj Juddoo
Suzana Brown
Suzana N. Russell
Svetlana Zens
Tabriz Ahmadli
Terry Ombati
Tetiana Avdieieva
Thy Try
Tolu Okitika
Tong Khanh Linh
Veszna Wessenauer
Víctor González
Wan Zuraini Mahrawi
Yacine Khelladi
Yuan Y. Stevens
Yusra Ahmad
Zilan Qian

# Disclaimer

The data used to produce this report was compiled with the support of partner organisations listed in the Acknowledgements section. However, data and analysis do not necessarily represent the views, positions, or opinions of those individual organisations, and any error or omission is the sole responsibility of the Global Index on Responsible AI project. In particular, the views expressed herein do not necessarily represent those of IDRC or its Board of Governors, the Government of Canada, or FCDO. The data used in this report covers the period 1 November 2023 – 30 September 2025.

# Foreword: Robert Opp

Across the majority of the 170 countries and territories where UNDP works, artificial intelligence has moved, with unusual speed, from a speculative conversation into a concrete governance and delivery challenge. Governments are not only asking what AI might mean for their societies; they are also making operational choices about how AI enters public services, how it is procured, how it reshapes labour markets, and whether people can understand or challenge its use. The 2nd Edition of the Global Index on Responsible AI arrives at precisely this transition from debate to deployment, and that is why it is such a timely addition to the global discussion on AI.

The findings of the report are telling in this regard. Where AI governance frameworks are active, only 55% of those cases have evidence of implementing them – and that drops to 45% in countries in the Global South. This is an indicator of the distance between responsible AI as a language of commitment and responsible AI as a system of protection. Frameworks, principles, and strategies are proliferating, but institutional capacity to translate them into accountability for people is not keeping pace with adoption.

The defining question is no longer whether countries adopt AI, but how AI is integrated, whose priorities it reflects, which risks are surfaced, who benefits as systems scale, and whether institutions retain meaningful agency and oversight as these systems evolve. And while much of the global conversation continues to be shaped by the priorities of those building frontier systems, AI is already being used by farmers consulting agricultural advisories, by health workers and teachers operating AI-enabled platforms, and within judicial, administrative, and financial systems shaping everyday decisions across a wide range of development contexts. The reality is that most governments today are engaging with AI primarily as adopters – and the choices they make in that role are development choices, shaping who is served, who is excluded, and whose interests are reflected as systems scale.

What we are seeing at UNDP through our work (on initiatives such as AI landscape assessments, strategy support, trust and safety, digital public infrastructure, data governance, low-resource languages, and capacity-building) across more than 50 countries is that AI adoption rarely happens through a single national decision or visible deployment moment. It accumulates through operational choices across procurement systems, vendor platforms, enterprise software, and institutional workflows. These choices are often fragmented and difficult to see in their totality. They frequently sit outside the reach of formal governance frameworks as well. As AI diffuses through public and economic systems in this way, the institutions where AI will be operationalised, contested, and trusted in practice will shape societal outcomes as much as the technology itself.

Public-sector AI deserves particular attention in this regard, because it is exercised through public authority and directly affects rights, services, entitlements, and trust in institutions. The Index's finding that fewer than one in five countries require disclosure of their own algorithmic systems is among its most consequential. Responsible AI must apply where public power is exercised, not only where markets are regulated. For developing countries, low implementation scores should not be read simply as lack of ambition. They reflect structural conditions: infrastructure, financing, technical evaluation capacity, institutional bandwidth, and bargaining power within global AI ecosystems. Implementation gaps, in many contexts, are development gaps.

The next phase of responsible AI governance will be less about putting principles in place and more focused on whether countries can build the institutions, safeguards, and coalitions to make those principles real. That means stronger public-sector transparency, procurement accountability, oversight capacity, meaningful redress pathways, multilingual inclusion, and governance embedded into the systems through which AI actually reaches people. It means treating trust and safety not as a question of model performance alone but of whether institutions can prevent, identify, and remedy the harms AI produces in public life. It means recognising that South-South cooperation, adopter-country coalitions, and shared institutional tools will be central to ensuring that global standards are negotiated rather than inherited.

The value of the Index is not just comparison but also visibility. It helps governments, civil society, funders, and development partners see where commitments are translating into protection, where accountability remains weak, and where collective action is most urgently needed. As AI moves from experimentation into infrastructure, that visibility is what the next phase requires to ensure it is a positive force for development rather than a driver of inequality.

**Robert Opp**
Chief Digital Officer, UNDP

# Foreword: Vilas Dhar

Someone you will never meet is teaching an AI to tell a tumor from healthy tissue, one labeled image at a time. They are paid per task, monitored by software, and ranked by a system they cannot see. The model they train will be called one of the most advanced technologies in human history, but the rules of the system where they live offer them almost nothing: no floor under their pay, no limit on the monitoring, and no way to know whether there will be dignity in their work tomorrow.

They are who this Index is built to see.

The Index assesses 135 countries against the one question that should organise this field: whether the protections governments promise reach the people most exposed to AI's harms. It is the most complete picture yet of how the world is governing AI responsibly, built by researchers in every region who gathered more than 68,000 data points in their own languages and legal systems. Its most valuable choice is also the hardest to quantify: whether frameworks reach people, not simply whether they exist on paper.

Judged that way, responsible AI has reached the end of its first phase. That phase was about writing things down: a decade of principles, declarations, and voluntary commitments that established AI as a public concern. It was necessary work, and it is no longer enough. By now, some 126 of 135 governments show a commitment. The Index then scores each country out of 100 on how far that commitment reaches in practice across rights, redress, and oversight, with a global average of just 35. Even where frameworks exist, evidence of implementation appears in just over half of cases, lower still across the Global South.

On its face, the Index assesses governments, but read more deeply, it also creates a scorecard for the rest of us: the funders, advocates, and institutions who have spent a decade building this field. Consider where the scores cluster – and what they show about our priorities: AI literacy is the strongest area, active in 106 countries. At the other end, fewer than one in five governments disclose their own use of AI, only 39 protect workers, and just 28 have built independent oversight bodies.

The data holds up a mirror to the field we built. Too often, we continue to advocate for the curricula, the convenings, the literacy campaigns: the work that is measurable, fundable, and less contested. We are slower to fund the work that binds power to responsibility: public-sector disclosure, redress, independent oversight, and institutions willing to challenge governments, companies, and funders alike.

The implementation gap this report measures is, in part, a gap in what our field has chosen to make possible.

The Index describes a two-tier world. Of all cases in wich an active framework covered a reviewed policy area, 22% in the Global South were addressed by binding instruments, compared with 58% in the North, and the EU AI Act alone accounts for roughly two-thirds of all enforceable protections gained since the 1st Edition. The reflex is to explain this as a capacity deficit: help the South catch up, harmonize towards a global standard – but such an explanation fails us all.

My experience suggests frameworks stay non-binding in the Global South for reasons that have little to do with imagination or legal skill. Peru has written one of the most comprehensive binding AI laws in the Global South; Uruguay has given platform workers the right to organize against the systems that manage them. What they lack is bargaining power: they must regulate technologies they depend on and cannot yet replace, and the right to regulate means little when infrastructure, standards, models, and capital sit elsewhere. A global system that concentrates standard-setting in a handful of capitals while exporting the burden of compliance cannot be the foundation for a just global AI future.

The Index also points to a more subtle danger – that we may rely on the promulgation of frameworks as a marker of satisfactory progress. When nearly every government can point to a commitment, the pressure to turn that commitment into protection can begin to fade. Recognition begins to stand in for remedy. An index that counted frameworks would have ratified that illusion – this one shows us we need to dig deeper.

The data leaves us with a concrete obligation: champion the half of the work we have avoided. Resource the oversight bodies, audits, redress mechanisms, and public-interest groups willing to challenge power directly. Back South-South coordination, so that the countries most exposed to AI negotiate global governance rather than inherit it, and insist that participation means more than being consulted after the design is done.

Brazil, Nigeria, and others are already building the next phase of responsible AI outside the usual centers of power, working closer to the harms, with fewer resources, under sharper constraints, and a clearer sense of what protection must mean.

That person training that model will not be protected simply by adding another framework to the count. They will be protected when the rules where they live can reach them, when workers can contest the systems that manage them, and when governments make their own use of AI accountable and visible. That work does not rest with governments alone, but with all of us. This Index is the evidence that it has already begun, and the work ahead is to close the distance between promise and protection.

**Vilas Dhar**

President, Patrick J. McGovern Foundation

# Contents

# Executive Summary

The 2nd Edition of the Global Index on Responsible AI (GIRAI, or Index) shows that responsible AI governance is expanding, but it is not expanding fast enough, deeply enough, or equitably enough to keep pace with the development and deployment of AI around the world.

More countries are adopting responsible AI frameworks; however, many remain unenforceable, weakly implemented, and disconnected from the institutions needed to protect people in practice. The central finding is clear: responsible AI governance evolution has shown the limits of the framework-building phase. The next phase must be judged by whether governments are building enforceable protections and accountable institutions, whether they are disclosing their own use of AI, and whether they are establishing clear routes for redress.

As AI becomes embedded in everyday life, responsible AI governance will help determine whether AI supports just, fair, equal, and sustainable futures or whether it will deepen existing harms and inequalities.

## What the Index Measures

The GIRAI asks whether governments are not only adopting responsible AI but governing AI in ways that protect rights, advance the public interest, and serve those most exposed to harm. The 2nd Edition assesses responsible AI governance across 135 countries and jurisdictions[1]. It is grounded in a human rights-based approach and informed by the UNESCO Recommendation on the Ethics of Artificial Intelligence.

The Index measures whether countries are building the frameworks, institutions, practices, and enabling conditions needed to govern AI responsibly. It is organised around five dimensions, which serve as broad organising principles for responsible AI:

### 1. Inclusion and Diversity

AI must be governed so that it does not reproduce existing hierarchies, deepen exclusion, or concentrate benefits among those already advantaged. This dimension assesses whether countries are protecting groups and communities most likely to be marginalised or harmed by AI systems.

1. Throughout this report, we use the term "countries" or "jurisdictions" interchangeably, to simplify communication. We recognise that some of the jurisdictions included in the GIRAI are currently disputed in nature. The naming of these jurisdictions in this report, as well as their geographical representation on the GIRAI website, does not reflect any position on such disputes or the views of the organisation.

## 2. Ethics and Sustainability

AI must be developed and deployed in ways that are fair, accountable, rights-respecting, and environmentally sustainable. This dimension assesses whether countries are addressing the ethical and ecological conditions under which AI is built and used.

## 3. Labour and Skills

AI is reshaping work, skills, livelihoods, and labour rights. This dimension assesses whether countries are preparing people to participate in AI-enabled economies while also protecting workers from displacement, exploitation, surveillance, discrimination, and weakened collective rights.

## 4. Trust and Safety

AI systems must be safe, secure, reliable, and accountable if they are to serve the public good. This dimension assesses whether countries are putting in place safeguards to prevent harm, respond to misuse, and maintain public trust.

## 5. AI Use in Public Service

When governments use AI, they exercise public power through technical systems. This dimension assesses whether countries are governing public-sector AI in ways that uphold rights, democratic values, transparency, accountability, and access to essential services.

Within these five dimensions, the Index develops 17 indicators assessing AI Policies from transversal categories of data, such as type of document, enforceability (binding/non-binding), regulatory scope (horizontal or vertical), stakeholder consultation, operationalisation (implementation body and plan, budget provisions, and monitoring and evaluation), thematic coverage, and implementation initiatives (government-led initiatives in absence of frameworks), in addition to five indicators on civil society organisation (CSO) engagement (one per dimension), 15 measuring contextual conditions, and one assessing credible evidence of government deployments of unacceptable risk AI systems (URAI systems). The URAI indicator is used to compute a penalty (URAI penalty) that is applied as an accountability adjustment to the final country score (Figure 1).

The Index does not directly assess individual AI systems or the conduct of AI companies. Its focus is on national-level governance conditions.

# Key Findings

## 1. AI is accelerating faster than governments can govern it in the public interest

Diffusion of AI is expanding, with 53% of the global population having used generative AI tools. Yet average GIRAI scores remain low, at roughly 35 out of 100, and evidence of implementation exists in only 55% of cases where frameworks are active, falling to 45% in Global South countries[2].

## 2. Responsible AI governance is expanding in Global South countries, but binding protections remain scarce

Since the 1st Edition, Global South countries substantially broadened the responsible AI content of their national frameworks. On average, the number of GIRAI topics covered rose from 2.5 to 4.7, an 83% increase[3]. In Global North countries, the number rose from 8.2 to 11.1, a 35% increase. Global South countries account for 203 of the 306 new country cases of indicators covered by frameworks identified since the 1st Edition. Despite this progress, most of the growth is in soft law: 78% of responsible AI framework cases in these countries are non-binding, compared with 42% in Global North countries.

## 3. AI safety is being governed as a technical problem, while human harms remain under-addressed

AI safety and security is one of the fastest-growing areas of governance, but much of it focuses on technical safeguards. Meanwhile, the Index found credible evidence of government use of unacceptable risk AI systems in 35 of 135 countries, and only 49 countries (36%) have frameworks addressing AI-facilitated misinformation and violence.

2. In this report, the term "Global South" refers to developing economies, and "Global North" refers to developed economies, as classified in the UN Standard country or area codes for statistical use (M49).
3. The increase of 83% is based on the unrounded averages (from 2.547 to 4.653). Using the rounded values reported in the text (2.5 and 4.7) would yield a slightly different percentage due to rounding.

## 4. Governments are regulating AI transparency but not disclosing their own use of AI

Transparency and Explainability is the strongest-performing indicator, with 58% of countries having some form of framework. Yet implementation lags behind the existence of frameworks. For AI Use in Public Service, Public Disclosure of Government Algorithmic Systems is the weakest-performing indicator, with only 18% of countries requiring disclosure of government AI systems.

## 5. Gender is increasingly recognised in AI governance, but protection from gendered harms remains weak

Gender equality is gaining visibility, with 29 new countries addressing gender and AI since the 1st Edition, but only 24 of 55 countries with gender-related frameworks show evidence of implementation. Protection from gendered AI harms remains limited.

## 6. Future generations are being prepared for the AI economy but not protected from AI-related harms

AI Literacy is one of the strongest-performing indicators, with 71 countries (53%) having some framework in place and 106 countries showing evidence of some activity in this area. By contrast, only 55 countries (41%) have frameworks addressing children's rights in AI, and only 27 of them show evidence of implementation.

## 7. AI's environmental footprint remains a blind spot in responsible AI governance

Only 27% of countries have frameworks addressing AI's environmental effects, and 83% of those frameworks are non-binding. Very few governments require disclosure of AI's energy use, water use, or environmental impact, contributing to making the environmental impact of AI a global blind spot.

## 8. Governments recognise the need for local-language AI but do not require developers to deliver it

Governments are investing in local-language technologies and cultural inclusion, with 52 countries (39%) showing government-led initiatives. Only 47 countries (35%) have frameworks addressingcultural and linguistic diversity, and few require developers to use diverse datasets or adapt systems to local contexts.

## 9. Governments are investing in AI skills but neglecting workers' rights

Labour protection frameworks exist in only 39 countries (29%), compared with 72 countries (53%) with frameworks on reskilling and upskilling. Few countries address workers' rights to organise and collectively bargain in response to AI-driven workplace change.

## 10. Global AI governance is fragmenting before a shared floor of protection has been established

Average GIRAI scores range from 55 in Global North countries to 27 in Global South countries. Available evidence shows that 164 of 215 recent AI-related frameworks are non-binding, and multi-stakeholder consultations appear only 31 times in the global implementation record. Only 73 of 135 countries (54%) have adopted a national AI policy or equivalent framework, and just 36 countries (27%) have operational mechanisms for participation of CSOs in AI governance. Without a shared rights-based floor, interoperability risks serving markets before it protects people.

# Priorities for the Next Phase of Responsible AI

**The next phase of responsible AI governance must move from commitments to protections. The report identifies 10 priorities:**

1. **Hold public-sector AI to a higher standard** through disclosure, public registers, procurement transparency, and routes for challenge.

2. **Build institutions as well as frameworks,** including oversight bodies, monitoring systems, audit capacity, and enforcement mechanisms, to complement binding frameworks that ensure sustained protection of rights over time.

3. **Establish binding floors in high-risk areas,** including public-sector AI, children's rights, gendered harms, misinformation, labour, environment, transparency, and redress.

4. **Make redress central to responsible AI** so that people can challenge AI-related decisions and obtain remedy.

5. **Bring labour and environmental harms into AI governance,** including data work, platform work, algorithmic management, energy use, water demand, and data centre impacts.

6. **Fund independent public-interest capacity and knowledge production,** including regulators, universities, civil society, labour groups, digital rights organisations, and independent researchers.

7. **Create the conditions for making participation meaningful,** especially for workers, children's rights advocates, women's rights organisations, disability groups, linguistic minorities, and affected communities.

8. **Build global convergence around rights, not only systems,** so that data, models, and standards do not move faster than protections.

9. **Treat monitoring and evidence as governance infrastructure,** including public registers, incident reporting, environmental reporting, and implementation reviews.

10. **Advance South–South cooperation on AI governance** so that Global South countries can negotiate the interoperability of frameworks and global standards rather than importing them.

## Conclusion

Responsible AI governance is growing, but growth alone is not enough. The next phase must be judged by whether responsible AI can protect people in the places where AI is already shaping rights, opportunities, dignity, environments, and everyday life.

Responsible AI will only fulfil its purpose if it moves from promise to practice.

# How to Read This Report

This report is structured into four parts. If you want to learn more about the rationale for the Index, how we undertook our research, and what we sought to measure (and what we didn't), you will find this set out in Part 1: Introduction and Background to the Global Index on Responsible AI.

An overview of the GIRAI scores distribution by region can be found in Part 2. If you wish to learn more about the data underpinning the top 10 takeaways of the GIRAI highlighted in the Executive Summary, you will find them in Part 3: Key Findings and Global Trends in Responsible AI. In each section, you will find Bright Spots that showcase the efforts of different countries in advancing responsible AI. Next steps for the GIRAI are described in Part 4: From Commitments to Protections: The Next Phase of Responsible AI Governance.

All information about the Index, including a full description of the methodology, approach to scoring, definitions, and scope of each indicator is available online at global-index.ai. You will also be able to access freely and openly all the country-level data from the 1st and 2nd Editions there.

PART 1

# Introduction and Background to the Global Index on Responsible AI

# From Promises to Protections: The Urgency of Tracking Responsible AI

AI is no longer confined to technology companies or specialised sectors of the economy. It now reaches into the ordinary places where people live their lives: classrooms, workplaces, clinics, welfare offices, borders, courtrooms, banks, and digital devices. Although AI technologies can help people access services, translate information, identify risks, improve diagnoses, and support all kinds of decision-making, it can also deny people opportunities, misclassify them as risks, expose them to surveillance, reproduce discrimination, intensify harassment, and make important decisions less visible and harder to contest.

Since the publication of the 1st Edition of the Index in 2024, AI development, diffusion, and usage has increased significantly. Generative AI technologies, like OpenAI's ChatGPT, have reached 53% population adoption, with varying degrees of usage in different countries[4]. In early 2026, ChatGPT had an estimated 1 billion users per week, all around the world. Across the Global South, large tech companies, grassroots organisations, and local initiatives have expanded their presence by establishing AI labs and development centres as well as convening conferences focusing on addressing development challenges in agriculture, healthcare, and education[5]. This increased usage of AI has been driven by major investments; over the past two years, global corporate investment in AI has nearly tripled to a reported $581.7 billion in 2025[6].

As the increasing development, diffusion, and use of AI abounds around the world, so too do the risks that it poses to people and their communities. The AI Incident Database, run by the Responsible AI Collective, documented a 46% increase in AI incidents reported in 2025[7]. This figure was already a 70% increase from 2023[8]. Behind these figures are people who have been wrongfully arrested after flawed facial recognition matches, targeted by AI-generated scams or synthetic media, or exposed to privacy breaches. Beyond these recorded incidents, many AI-enabled harms and rights violations remain hidden due to the inaccessibility of reporting mechanisms and opacity of AI systems, which further complicate efforts to understand the scale of these technologies' impact. A quickly widening terrain of easily accessible technologies with different user affordabilities is particularly challenging to navigate for younger and older generations as well as users with limited access to critical digital literacy.

With the rapid development and diffusion of AI, the stakes are especially high as systems operate at scale, across borders, and through technical processes that are difficult for ordinary people to see or contest. In these contexts, responsible AI is not an abstract ethical aspiration. It is the key condition for protecting human dignity, agency, equality, safety, and wellbeing.

Responsible AI, therefore, is the effort to ensure that the design, development,

4. Sajadieh, S., Fattorini, L., Perrault, R., Gil, Y., Parli, V., Santarlasci, L., ... & Weld, D. (2026). *The AI Index 2026 Annual Report*. AI Index Steering Committee, Institute for Human-Centered AI, Stanford University. https://doi.org/10.48550/arXiv.2606.15708
5. Okolo, C.T. (2023, November 1). 'AI in the Global South: Opportunities and challenges towards more inclusive governance'. *Brookings*. https://www.brookings.edu/articles/ai-in-the-global-south-opportunities-and-challenges-towards-more-inclusive-governance/
6. Sajadieh, S., Fattorini, L., Perrault, R., Gil, Y., Parli, V., Santarlasci, L., ... & Weld, D. (2026). *The AI Index 2026 Annual Report*.
7. The AI Incident Database. https://incidentdatabase.ai/apps/discover/?hideDuplicates=1&is_incident_report=true Accessed July 2, 2026.
8. Ibid

deployment, implementation, and governance of AI systems respect human rights, protect people from harm, promote inclusion, and serve the public good. It is grounded in the recognition that AI systems are not neutral tools. They are developed and deployed within existing social, economic, and political conditions.

As AI becomes further embedded in everyday life, how it is governed will determine our collective human future. The extent of adoption and, crucially, meaningful implementation of responsible governance of AI will determine whether the future is more just, fair, equal, and sustainable or whether existing inequalities and harms are deepened.

# Measuring Responsible AI: The Global Index on Responsible AI

Responsible AI cannot be secured through rhetoric or principles alone. Realising responsible AI as a force that shapes positive human outcomes with AI requires translating commitments into institutions, resources, oversight mechanisms, public participation, enforcement powers, monitoring systems, and accessible routes for redress.

Without measurement, responsible AI remains abstract. Measuring responsible AI allows us to evaluate whether countries are fostering the conditions needed for AI to serve human life and wellbeing and the sustainability of the planet. Measuring gaps between aspiration and action visibilises where responsible AI governance is advancing, where it is stalled or hasn't begun, and where commitments are not yet reaching the people they are meant to protect. Measurement helps move the global debate from what governments say about responsible AI towards what they are actually doing to make it possible: in short, from aspiration to evidence.

The GIRAI was established to provide this evidence. Grounded in a human rights-based approach and informed by the UNESCO Recommendation on the Ethics of Artificial Intelligence and other relevant global agreements and commitments, the Index establishes measurable benchmarks for responsible AI governance across countries. It assesses whether 135 governments around the world are adopting relevant frameworks, whether those frameworks are being implemented, whether civil society is able to participate and hold actors accountable, whether broader institutional conditions support rights-respecting AI governance, and whether governments are using AI systems for purposes that might threaten fundamental rights.

The Index is especially important because global evidence on responsible AI remains uneven. Much of the debate on AI governance has been shaped by developments in Europe and Northern America, while data from Africa, South and Central America, the Caribbean, Asia, and the Middle East has often been fragmented or absent. This distorts the global picture of responsible AI, makes the diverse governance challenges countries face harder to understand, and obscures the bright spots where responsible AI is taking place outside of dominant global centres. GIRAI constitutes the largest effort to collect global data on the state of responsible AI.

By collecting globally representative, country-level evidence, GIRAI provides a comparative view of how responsible AI

is being advanced across regions, income groups, and governance contexts. It also shows that responsible AI is a matter not only of adopting frameworks but also of building the institutional and social conditions through which those frameworks can protect people in practice.

The purpose of the Index is to support accountability, policy learning, and public action. It helps policymakers identify where stronger laws, institutions, resources, and implementation mechanisms are needed. It helps civil society actors see where governments have made commitments and where they can be held to account. It helps international organisations, funders, and researchers understand where capacity gaps are most acute and where support is most urgently required. To do so, it sets benchmarks against which countries are assessed. Qualitative evaluations of the collected evidence are transformed into quantitative variables and scores to identify global, regional, and local trends.

Ultimately, measuring responsible AI is a way of asking whether the AI futures being built around the world are worthy of the people who will live in those futures. It asks whether AI governance is creating the conditions for societies in which people are not merely protected from harm but can participate, contest, flourish, and live with dignity.

# What the Global Index on Responsible AI Measures

The 2nd Edition of the GIRAI assesses the extent to which countries are building the frameworks, institutions, practices, and enabling conditions needed to govern AI responsibly. The Index provides a detailed picture of how responsible AI is taking form, where gaps persist, where government abuse occurs, and the enabling conditions in place that may render responsible AI a lived condition for as many people as possible.

This global dataset was developed by a network of country-level experts through assessment of publicly available government frameworks, implementation activity, civil society engagement, and documented cases of government use of AI with unacceptable risks across 135 countries. Our global team of researchers identified and assessed 376 frameworks for responsible AI, 828 government initiatives advancing responsible AI in practice, and 615 responsible AI initiatives driven by CSOs. In total, 68,138 data points on responsible AI were collected and assessed, alongside secondary data sources capturing broader enabling conditions, including rule of law, equality, cybersecurity, low-carbon energy use, labour rights compliance, digital skills, and access to public information, among others.

**The Index is structured around five dimensions of responsible AI governance:**

1. Inclusion and Diversity
2. Ethics and Sustainability
3. Labour and Skills
4. Trust and Safety
5. AI Use in Public Service

Across these dimensions, the Index is organised into three pillars: AI Policy, which measures the existence, quality, and implementation of government frameworks and initiatives; CSO Engagement, which recognises the role of CSOs in shaping, scrutinising, and contesting AI governance by measuring civil society activity and mechanisms for CSO inclusion; and Enabling Conditions, which complements the other two pillars by capturing broader structural conditions that may support or constrain responsible AI governance within a country.

Each dimension contains indicators that capture core areas of responsible AI, including gender equality, children's rights, fairness and non-discrimination, transparency and explainability, human oversight, environmental impact, rule of law, labour protections, AI literacy, safety and security, cybersecurity, redress and remedy, and responsible government use of AI.

For AI Policy indicators, the Index assesses the existence and quality of government frameworks as well as evidence of government action, including both the implementation of existing frameworks and standalone government-led initiatives where no formal framework is yet in place. Assessing implementation of government frameworks on AI allows a better understanding of the distance between a government's promise to govern AI in a human-centred way and the lived reality of people affected by AI systems. Without implementation, responsible AI fails at its most basic task and loses its practical meaning to preserve human wellbeing and autonomy and protect people from harm when AI enters their lives.

A separate indicator records credible evidence of government AI deployments that pose serious risks to human rights, rule of law, or democratic governance. The Index includes this indicator because the misuse of AI by state actors represents one of the most pressing concerns in the global AI governance landscape, with documented cases ranging from mass surveillance and social scoring systems to the use of AI-powered tools in the context of armed conflict and crimes against humanity. We acknowledge that the way this indicator is incorporated into the Index may not fully capture the scale or severity of such misuses — a country engaged in serious violations may still score relatively well on other governance dimensions. However, the intent of the Index is to track governmental commitments and progress towards responsible AI governance, identify gaps, and promote accountability. We recognise the limitations of this approach and remain committed to refining our methodology in dialogue with the broader research and human rights communities.

**The 2nd Edition therefore provides insights into the following questions:**

What is the global state of responsible AI?

What actions have countries taken to advance their commitment to practicing responsible governance, use, and development of AI?

What is the gap between commitments to responsible AI in national frameworks and their implementation?

To what degree are responsible AI frameworks providing enforceable protections for individuals and groups?

What role are CSOs around the world playing in advancing responsible AI?

What are the evident regional and global trends emerging in relation to the implementation – or lack thereof – of responsible AI standards?

What are the major capacity gaps in advancing responsible AI governance and practice around the world?

## The GIRAI does not directly assess:

Whether individual AI systems are designed, developed, or deployed in line with responsible AI standards but rather focuses on government and national-level commitments and activities.

The conduct, investments, or impacts of major technology companies and AI developers across different countries.

# Scope of the 2nd Edition

Any index is fundamentally constrained by the availability and accessibility of data. It is also subject to cultural and political biases that may result in an incomplete representation of specific dimensions and thematic areas. Carrying out a study of this scale required making decisions to ensure that the measurement framework was inclusive enough to accommodate the complexities and realities of as many countries as possible yet specific enough to gauge the measurability of a new conceptual framework for responsible AI and to allow a fair comparison between countries.

The data collected in the 2nd Edition relates to responsible AI activities from 1 November 2023 to 30 September 2025 and therefore does not reflect developments in responsible AI that occurred in late 2025 or early 2026. Advancements taking place between 1 October 2025 and 30 September 2027 will be captured in the 3rd Edition of the Index.

While the methodology and scoring system of the GIRAI has been updated to better capture nuances and details on existing frameworks and government-led initiatives, this report was primarily built using basic descriptive statistics based on the existence and type of different kinds of evidence. This decision was made for two reasons. Firstly, we wanted to use clear language that easily depicts AI governance ecosystems: saying "The average score of the indicator on Children's Rights is 28 points" is not the same as saying "Only 55 countries around the world have some framework in place while 37 have government-led initiatives." We are aware that the scoring system reflects this, and many more variables were used to assign scores and positions to countries; however, for this report we leaned towards the latter example to graphically showcase the status of responsible AI in clear and direct language. Although less detailed, these kinds of insights are more easily actionable.

Secondly, transforming all variables of an indicator into scores involves multiple methodological decisions that reflect conceptual choices that may not always be clear and transparent to all stakeholders. To keep things simple, we avoided complex score calculations to make the report more transparent and accessible.

# Methodology and Scoring

First-hand data was collected by the 135 country-level researchers through a comprehensive expert Global Survey using different questionnaire templates depending on the type of indicator assessed. These included questionnaires on government AI policy frameworks and initiatives, civil society engagement, government mechanisms for CSO inclusion, and documented deployments of AI systems with unacceptable risks. Together, the four questionnaire templates covered up to 382 questions, many conditionally activated depending on the type and availability of evidence. The survey was designed to ascertain what actions were taken to advance responsible AI in each country surveyed between November 2023 and October 2025. A global team of quality assessors conducted an exhaustive review of all data collected.

GIRAI scores are derived from 38 indicators organised across the five equally weighted dimensions and three pillars, in a matrix. Each dimension combines those three pillars: AI Policy, weighted at 60%; CSO Engagement, weighted at 10%; and Enabling Conditions, weighted at 30%. The number of indicators varies by dimension and pillar: AI Policy includes 17 indicators across the five dimensions, CSO Engagement includes one indicator per dimension, and Enabling Conditions includes 15 indicators based on external datasets. AI Policy scores assess government frameworks and government-led initiatives; CSO Engagement scores assess civil society activity and mechanisms for CSO inclusion; and Enabling Conditions scores capture broader structural conditions for responsible AI governance. Indicator values are normalised to a common range (0-100), averaged into dimension/pillar scores (cells in Table 1), combined into dimension scores (weighted average, rows in Figure 1), and then averaged across the five dimensions to produce the final GIRAI score.

A separate indicator on Unacceptable Risks AI Systems (URAI penalty), which captures publicly documented cases or credible reports of a government's use of AI systems that pose unacceptable risk to human rights, rule of law, or democratic governance, is applied as an accountability penalty to the final country score. The categorisation includes: biometric surveillance, criminal justice and law enforcement misuse, social scoring systems, AI-driven cyberattacks, disinformation or influence operations, AI weapons or AI use in armed conflicts, and discriminatory public service systems. Researchers could also report additional cases not captured by these categories under "Others." For each indicator, country-level researchers received comprehensive two-page documents with definitions, justifications, and references to international frameworks. In this case, the EU AI Act's categorization of unacceptable risk uses was provided as a reference, given the close alignment with the indicator and GIRAI measurement framework and its human rights approach. The focus was on whether governments deployed systems that fell into any of these categories rather than a qualitative assessment of the negative impacts of government AI systems. All documented cases were validated by a review team. Although the research process aimed to document all verifiable instances of government misuse, we acknowledge that the findings could have been impacted by the availability of credible information on the subject. Governments may be generally less inclined to declare information about this kind of AI systems use, unlike other indicators where governments and other actors actively publicise their actions. While jurisdictions that lack robust civil society engagement or face limited access to information may have unreported incidents due to limited independent oversight and countries with greater transparency in AI systems disclosure may be victims of detection bias, the diversity of countries[9] in which this kind of evidence was found indicates that collecting such data remains feasible. Moreover, the indicator remains essential to measuring responsible AI practices by governments, whose responsibilities extend beyond legislation and implementation and include refraining from using AI in a manner that inflicts harm on civilians. It is also intended to encourage further investigation and reporting of government misuse of AI systems and to serve as a resource for analysing the correlation between documenting AI misuse and the presence of democratic freedoms to advocate for spaces where such acts can be recorded longitudinally.

The countries and jurisdictions included in the 2nd Edition were chosen based on where suitable country-level researchers were recruited and the existing research network of institutional partners. Governments were not involved in determining whether their country was included in the Index.

All data underwent a structured quality assurance process to assess consistency, relevance, and evidentiary strength. The Index is designed to provide a systematic, comparable, and globally representative assessment of responsible AI governance ecosystems. It should be read as a measure of publicly verifiable governance conditions, not as a complete account of all AI activity within a country.

9. Australia, Bangladesh, Belarus, Bolivia (Plurinational State of), Brazil, Canada, Chile, China, France, Germany, Ghana, Greece, India, Israel, Japan, Kenya, Kuwait, Kyrgyz Republic, Malaysia, Mexico, Montenegro, Myanmar, Nepal, Netherlands, Pakistan, Portugal, Saudi Arabia, South Korea, Spain, Uganda, Ukraine, United Arab Emirates, United Kingdom of Great Britain and Northern Ireland, United States of America, Uruguay.

Figure 1: GIRAI's Conceptual Framework (Dimensions, Pillars, and Indicators)

| **GIRAI Matrix —** Dimensions (rows), Pillars (columns), and Indicators (cells) | **A. AI Policy** (For each indicator, AI Policy = Frameworks + Initiatives) | **B. CSO Engagement** | **C. Enabling Conditions** |
|---|---|---|---|
| **1. Inclusion and Diversity** | 1. Gender Equality<br>2. Children's Rights<br>3. Cultural and Linguistic Diversity | 1. Civil Society Engagement in Inclusion and Diversity | 1. Egalitarian Democracy<br>2. Device Affordability<br>3. Gender Gap in Mobile Internet |
| **2. Ethics and Sustainability** | 4. Fairness and Non-Discrimination<br>5. Transparency and Explainability<br>6. Human Oversight and Determination<br>7. Environmental Impact | 2. Civil Society Engagement in Ethics and Sustainability | 4. Rule of Law<br>5. Low-Carbon Energy Share |
| **3. Labour and Skills** | 8. Labour Protections<br>9. Reskilling and Upskilling Initiatives<br>10. AI Literacy | 3. Civil Society Engagement in Labour and Skills | 6. Skills and Literacy<br>7. Labour Rights Compliance |
| **4. Trust and Safety** | 11. Safety and Security<br>12. Access to Redress and Remedy<br>13. Impact Assessments<br>14. AI-facilitated Misinformation and Violence | 4. Civil Society Engagement in Trust and Safety | 8. Cybersecurity<br>9. Data Protection and Privacy<br>10. Data Sharing and Access<br>11. Consumer Protection<br>12. Global Peace |
| **5. AI Use in Public Service** | 15. Public Sector Skills Development<br>16. Public Disclosure of Government Algorithmic Systems<br>17. Public Procurement | 5. Government Mechanisms for CSO Inclusion in AI Policy and Governance | 13. Civil Society Accountability<br>14. Public Service Delivery<br>15. Access to Public Information |
| **Unacceptable Risks AI Systems (URAI penalty) — transversal.** | | | |

*Note: AI Policy, CSO Engagement, and the URAI penalty indicators come from the Global Survey; Enabling Conditions indicators come from secondary sources.*

PART 2

# Scores Distribution per Region

## Figure 2: Scores Distribution per Region

Each bubble represents a country. The horizontal axis groups countries by region; the vertical axis shows GIRAI scores (0–100). Bubble colour reflects World Bank income group; bubble size reflects population.

Score
80
70
60
50
40
30
20
10
0

Northern America
Caribbean
South and Central America
Europe
Africa
Middle East
Asia and Oceania

Norway
ITA
FRA
DEU
GBR
ESP
Brazil
Canada
United States of America
COL
Australia
JPN
KOR
China
Dominican Republic
Nigeria
Israel
EGY
VNM
India
KEN
Indonesia
ARG
Pakistan
THA
ETH
PHL
ZAF
MEX
TZA
BGD
UGA
DZA
MMR
Honduras
Belarus
Lebanon
COD
Haiti
AFG
South Sudan

Country income level: Low … High
Population: 1M … 100M

*Source: Global Index on Responsible AI (GIRAI) Global Survey, 2026.*

PART 3

# Key Findings and Global Trends in Responsible AI

# 1 AI is accelerating faster than governments can govern it in the public interest

Public capacity to govern AI is not keeping pace with the rapid acceleration of AI investment, development, and adoption. Tracked global corporate investment in AI nearly tripled in two years, rising from around $200 billion in 2023 to $581.7 billion in 2025[10]. At the same time, AI systems are becoming embedded in workplaces, public services, education, healthcare, policing, finance, information systems, and everyday life, with 53% population adoption of generative AI alone[11].

Despite 128 out of 135 governments showing some form of commitment to responsible AI, average GIRAI scores remain low, at roughly 35 out of 100 across all dimensions. While most countries are taking some action, the depth of governance capacity remains far below what is needed to match the scale and speed of AI deployment and diffusion.

The implementation gap is central to this finding[12]. Across all indicators with an active framework for responsible AI, evidence of implementation exists in only 55% of cases globally, falling to 45% in the Global South. Many governments have not yet built the oversight, enforcement, monitoring, or redress systems needed to make responsible AI meaningful in practice.

The picture becomes clearer when we look at what implementation entails. The most common activities are training courses, workshops, and operational programmes. These may be useful, but they are short-term, reactive, and not substitutes for durable governance institutions. Across the 691 implementation activities for the government frameworks we assessed across 104 countries with some active framework, the Index identified only 28 countries with independent oversight bodies, 22 with monitoring and evaluation systems, 23 with standards or certification mechanisms, and 16 with multistakeholder advisory processes globally. In contrast, 70 countries showed evidence of implementation activities consisting of training courses and workshops.

10. Sajadieh, S., Fattorini, L., Perrault, R., Gil, Y., Parli, V., Santarlasci, L., ... & Weld, D. (2026). *The AI Index 2026 Annual Report.*
11. Ibid.
12. The AI Policy pillar indicators of the GIRAI comprise two components: Government Frameworks and Government-led Initiatives. Country-level experts first assess existing frameworks and then identify government-led initiatives that operationalise those frameworks. An implementation gap occurs when a framework is identified for a given indicator but no corresponding government-led action is found.

Figure 3: **Top 10 Implementation Activities for Active Frameworks**

Bars show the number of countries with at least one active framework accompanied by at least one of each type of implementation activity.

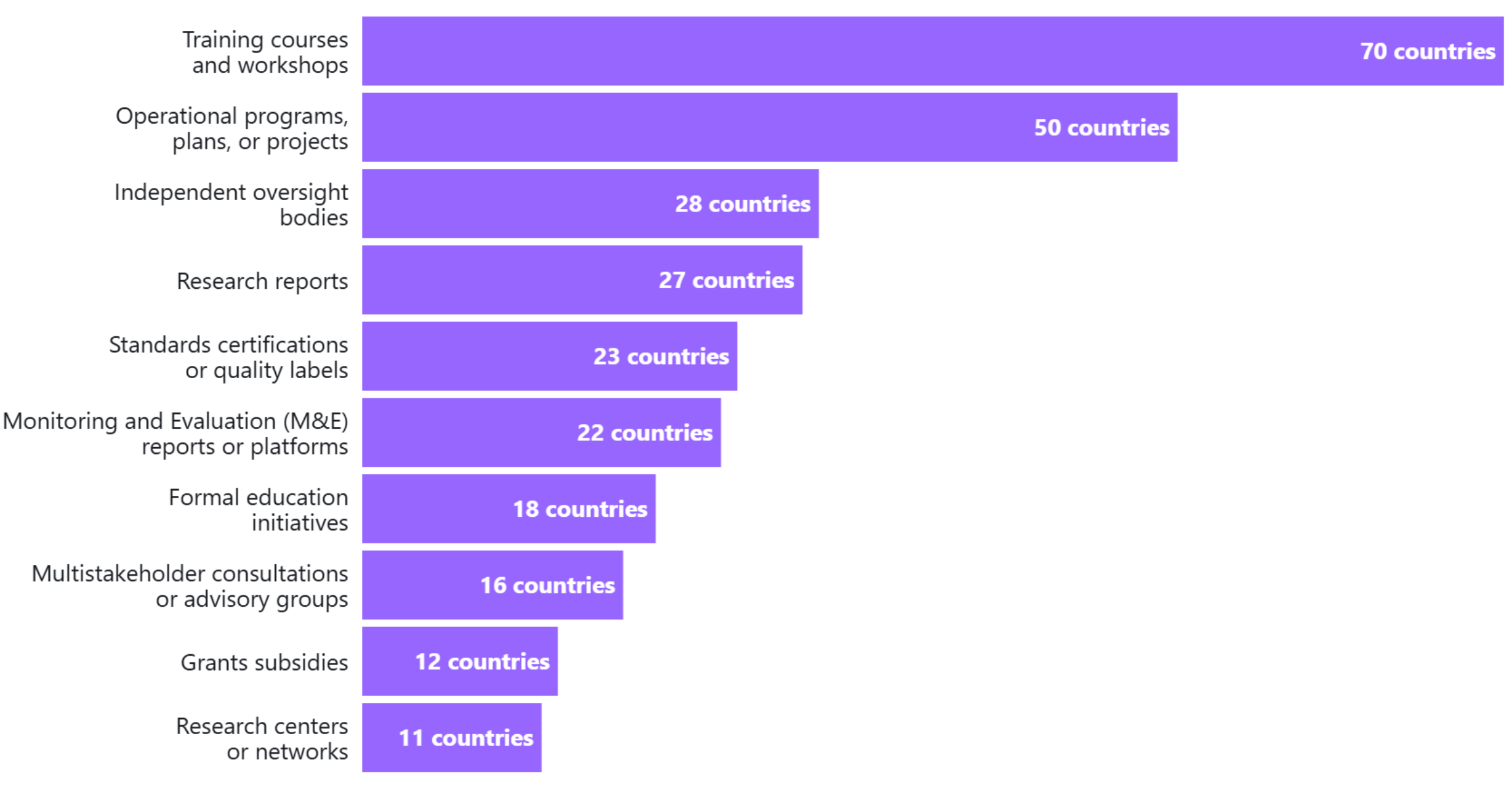

*Source: Global Index on Responsible AI (GIRAI) Global Survey, 2026.*

In addition, the weakest-performing dimension of the Index was AI Use in Public Service. Countries had frameworks addressing indicators in this dimension in 31% of cases, compared to a global average of 42% across all dimensions. The two worst-performing indicators of the Index were also within this dimension: only 26% of countries had frameworks ensuring fair and accountable Public Procurement of AI systems, and only 18% had frameworks requiring Public Disclosure of Government Algorithmic Systems.

Brazil, Chile, and Australia are some of the countries with notable public procurement mandates in place. *Brazil's AI Strategy*[13] requires that considerations for public procurement of AI go beyond technical efficiency to include ethical considerations on the transparency and data protection compliance obligations of such systems – see Bright Spot: Brazil. Chile's *Public Procurement Directive No. 44 on Recommendations for the Acquisition of Projects that Include Data Science and Artificial Intelligence (AI)*[14] establishes a well-structured process for ensuring ethical, transparent, and secure acquisitions of AI systems by government. It details procurement phases from problem identification to solution proposals to contract management and includes measures for vendor evaluation and contract management (Sections 4–5), as well as transparency, data protection, bias mitigation, and algorithmic auditing.

13. The Brazilian Ministry of Science, Technology and Innovation Ordinance No. 4,617, of April 6, 2021. (2021). *Brazilian Artificial Intelligence Strategy*. https://web.archive.org/web/20251001212006/https:/antigo.mctic.gov.br/mctic/opencms/legislacao/portarias/Portaria_MCTI_n_4617_de_06042021.html
14. Directorate of Purchases and Public Contracting, Republic of Chile. (2023, December 7). 'Public Procurement Directive No. 44 on Recommendations for the Acquisition of Projects That Include Data Science and Artificial Intelligence (AI)'. https://web.archive.org/web/20251026193125/https:/www.chilecompra.cl/wp-content/uploads/2023/12/Directiva-Recomendaciones-proyectos-Ciencia-Datos-IA.pdf

Australia's *National Framework for the Assurance of Artificial Intelligence in Government*[15] contains a dedicated "Procurement" section (p. 10) that highlights several procurement requirements, including clearly established accountability mechanisms, proof of performance testing, an assessment of government staff capacity and understanding of the operations of the proposed system, as well as knowledge transfer arrangements between the supplier and government to prevent vendor lock-in.

The governance gap in the use of AI in public service reveals an asymmetry in how responsible AI is being governed. AI is increasingly treated as something the state must regulate in the market but not as something the state must account for in its own exercise of power. Public-sector AI directly affects people's rights, freedoms, entitlements, and access to basic services, including welfare, healthcare, and education. For this reason, governments should not exempt their own deployments from the standards they set for others. To the contrary, the public sector should be held to a higher standard of responsible AI because it exercises public power.

Overall, the findings suggest that responsible AI governance is expanding, but the institutions, systems, and infrastructure needed to govern AI in the public interest remain underdeveloped and are being outpaced by the power and rapid diffusion of AI systems.

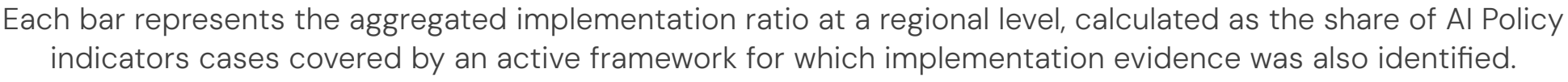

Figure 4: **AI Policy Frameworks Implementation Ratio by Region**

Each bar represents the aggregated implementation ratio at a regional level, calculated as the share of AI Policy indicators cases covered by an active framework for which implementation evidence was also identified.

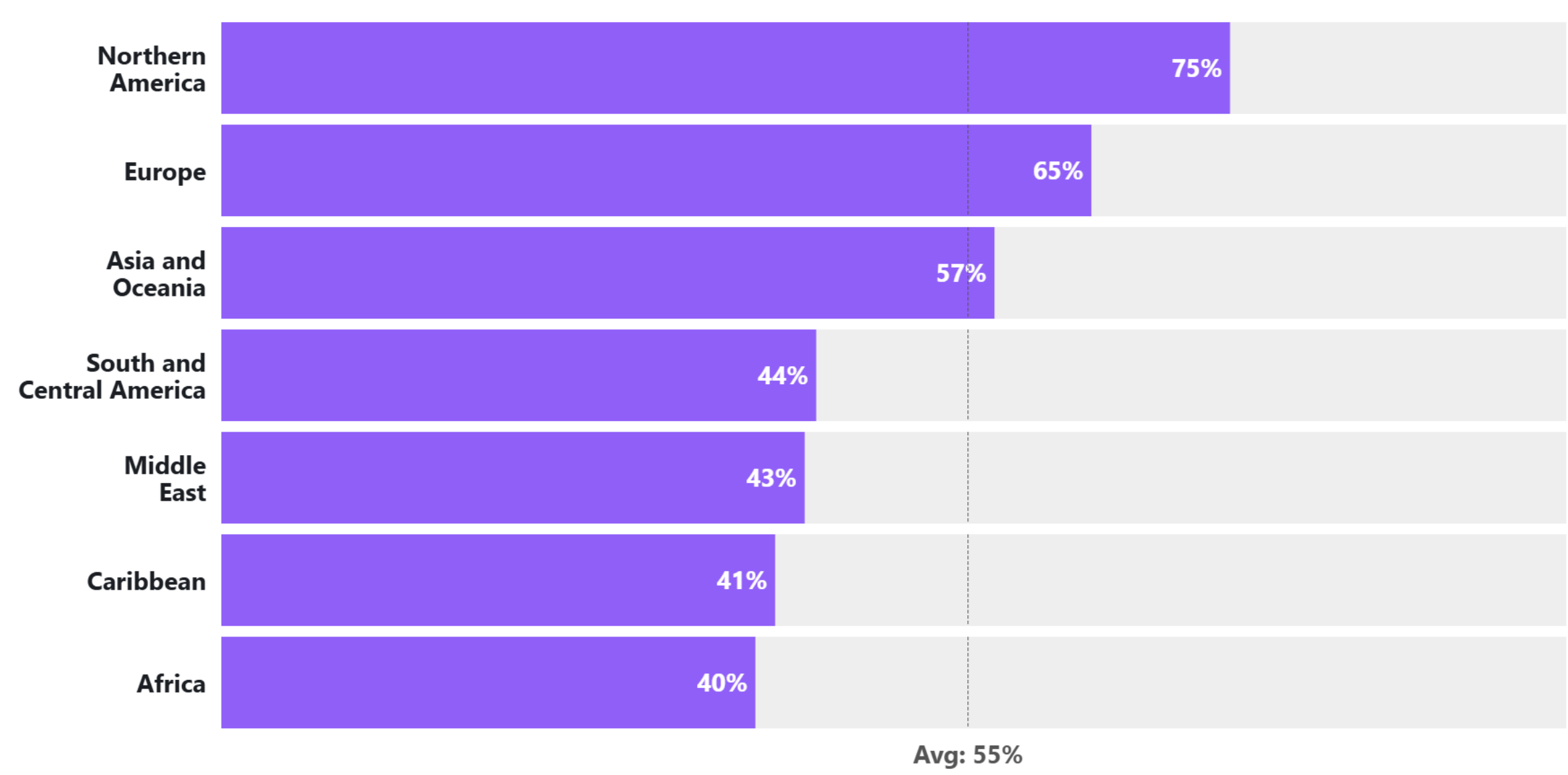

*Source: Global Index on Responsible AI (GIRAI) Global Survey, 2026.*

15. Australian Government. (2024, June 21). *National framework for the assurance of artificial intelligence in government.* https://web.archive.org/web/20260207085535/https:/www.finance.gov.au/sites/default/files/2024-06/National-framework-for-the-assurance-of-AI-in-government.pdf

Figure 5: **Framework Coverage and Implementation Ratio by AI Policy Indicator**

Grey dots show the number of countries with adopted frameworks per indicator; orange bars show the number of countries with at least one government-led implementation initiative. Percentage labels show the share of framework-adopting countries with accompanying implementation initiatives.

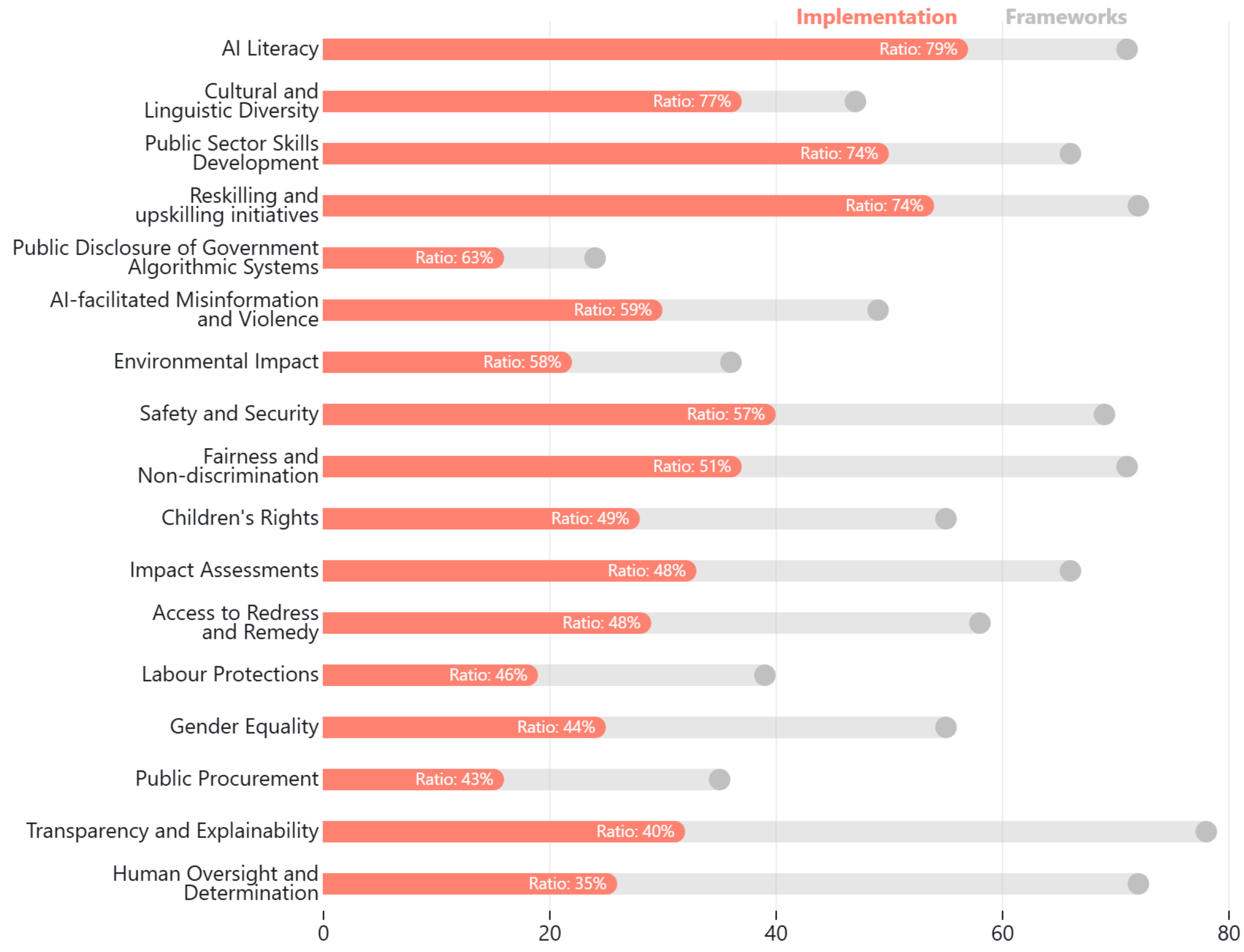

*Source: Global Index on Responsible AI (GIRAI) Global Survey, 2026.*

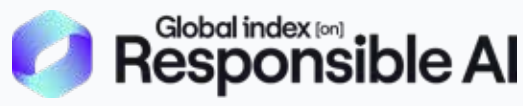

# Bright Spot: Brazil

Beyond framework adoption, government implementation is crucial to setting an example in responsible AI governance for corporate entities and individuals. Brazil stands out as one of the few Global South countries with both clear frameworks and implementation evidence in governance areas where many countries have neither. For instance, only 15[13] of all 135 countries in the Index have both a policy and a government-led initiative on public procurement of AI; Brazil is one of them. The Brazilian AI Strategy establishes strategic actions for AI procurement in public bidding processes, complemented by InovaCPIN, an online platform developed by five government agencies to support public procurement of AI solutions. Similarly, Brazil is one of the seven[14] Global South countries that have AI environmental impact frameworks and implementation of activities to strengthen them. Its AI for the Good of All Plan embeds environmental sustainability as a structuring premise, backed by the Redata regulatory act that establishes a national data centre policy with energy efficiency and sustainability requirements. On labour protections, where only 12[15] of 98 Global South countries show any sort of government implementation activities, Brazil's Ministry of Labour and Employment established a dedicated Working Group on Artificial Intelligence to study AI's impact on employment, and the Brazilian Institute of Information in Science and Technology convened a public hearing on AI and the job market. Brazil has also published a Generative AI Handbook for Public Service addressing human oversight and risk classification for AI use and mandating public disclosure of government AI systems — an indicator where only 13 Global South countries[16] have some framework or initiatives in place.

16. Brazil, Canada, Chile, Dominican Republic, France, Germany, Israel, Japan, Malaysia, New Zealand, Norway, Oman, Slovenia, United Kingdom, United States of America.
17. Brazil, China, Dominican Republic, Kenya, Malaysia, Saudi Arabia, Singapore.
18. Benin, Brazil, Chile, China, Colombia, Costa Rica, Dominican Republic, Mexico, Philippines, Trinidad and Tobago, Uruguay, Viet Nam.
19. Argentina, Bolivia, Brazil, Chile, China, China Hong Kong Special Administrative Region, Colombia, Costa Rica, Ecuador, Mongolia, Pakistan, Uruguay, Viet Nam.

# 2 Responsible AI governance is expanding in Global South countries, but binding protections remain scarce

Global South countries are no longer peripheral to global responsible AI governance. Since the 1st Edition, they have driven much of the expansion in framework coverage. Between editions, framework coverage among Global South countries increased by 83%, from an average of 2.5 key areas covered to 4.7, while Global North countries saw a 35% increase, from 8.2 to 11.1[20]. This expansion has mostly taken the form of soft law. In the Global South, from all the cases where an active framework covered a reviewed policy area, 78% were non-binding. In Global North countries, 42% of responsible AI frameworks were non-binding.

Figure 6: **Average Number of AI Policy Indicators Covered by Frameworks, by Enforceability and Global Region, 2024–2026**

Stacked columns show the average number of AI Policy indicators covered per country in the Global North and Global South, disaggregated by enforceability, for 2024 and 2026. Arrows indicate changes between editions across the 130 countries assessed in both.

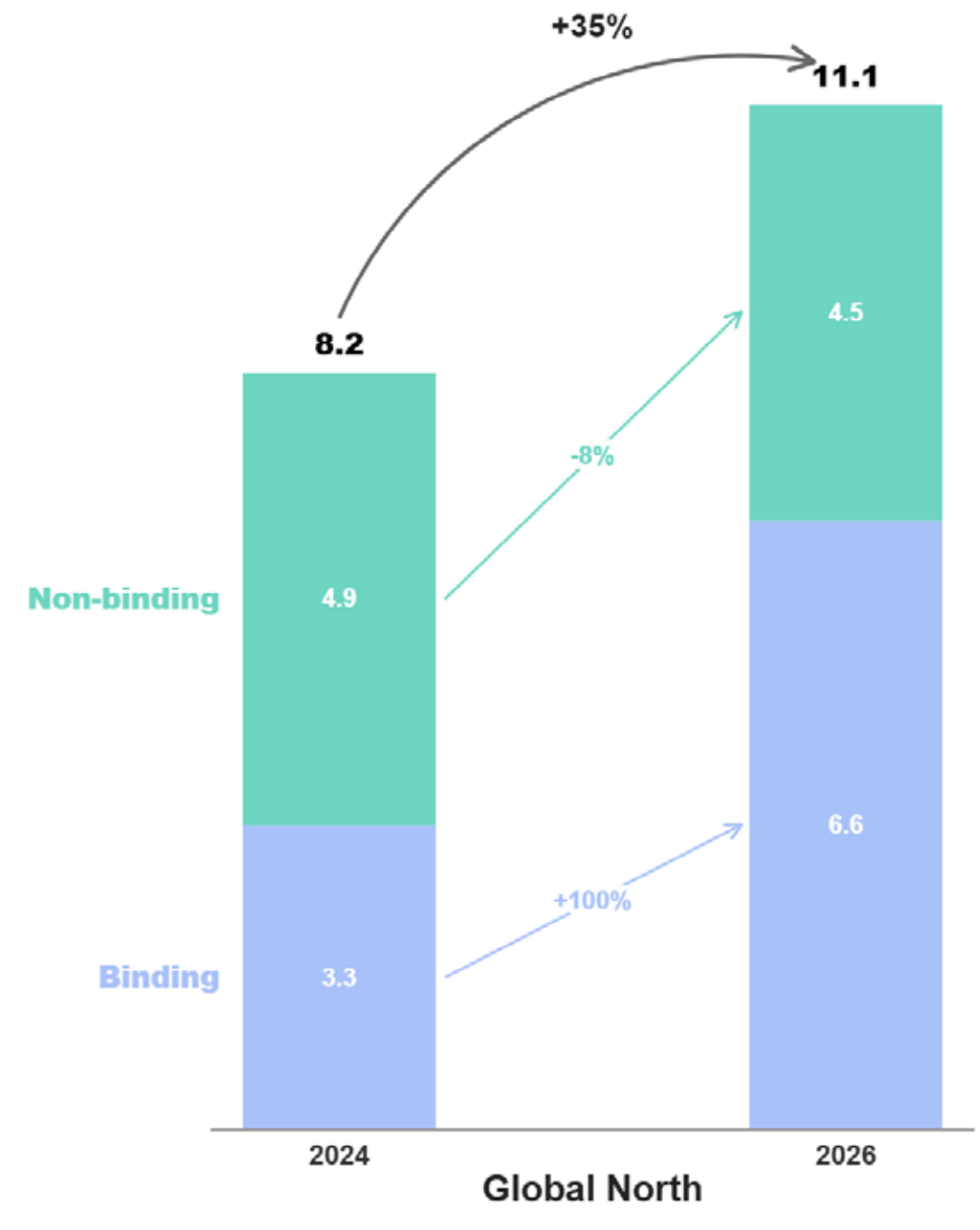

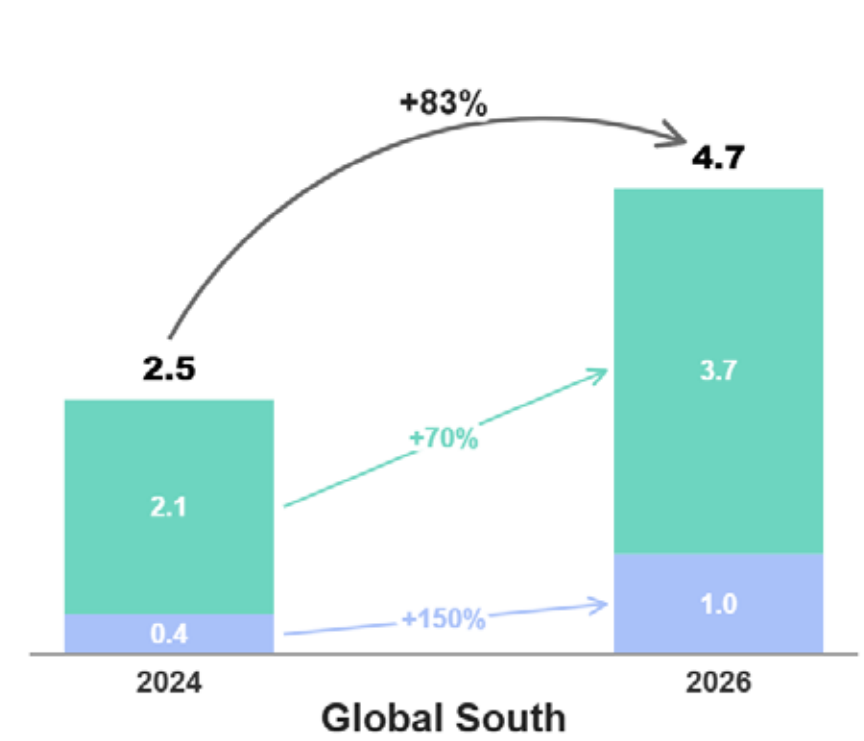

*Source: Global Index on Responsible AI (GIRAI) Global Survey, 2026.*

20. Brazil, Canada, Chile, Dominican Republic, France, Germany, Israel, Japan, Malaysia, New Zealand, Norway, Oman, Slovenia, United Kingdom, United States of America.

This means that the global diffusion of responsible AI governance is not the same as the global diffusion of enforceable protection. Global North countries account for 67 of the 76 (88%) enforceability gains recorded in the Index between Editions, while Global South countries account for only nine (12%). The *EU AI Act*[21] alone drives 51 of these gains throughout the 21 EU countries assessed in the 2nd Edition and across 10 indicators covered by this new framework, along with other EU regional regulations[22, 23]. Because the EU has advanced through a single, largely homogenous framework, it might be difficult to compare on equal terms with states pursuing divergent national approaches. In reality, the EU AI Act's influence as a centralised governance framework could have masked even further how limited progress on enforceable frameworks remains globally and how unevenly and differently that progress is being produced.

There are, however, important exceptions. In Peru, *Supreme Decree 115-2025-PCM*[24] provides binding coverage across several areas of responsible AI governance, including prohibitions on manipulative or deceptive AI uses and specific protections for vulnerable groups, including children, adolescents, and women - see Bright Spot: Peru. The *Digital Code of the Kyrgyz Republic 2025*[25] grants individuals the right to object to automated decision-making and appeal such processing to a regulator or court, while also mandating risk assessments for AI systems, including environmental impacts. Ethiopia's *Proclamation No. 1321/2024*[26] on personal data protection requires employers to inform staff about automated decision-making and profiling and gives individuals the right to request human intervention, express their views, and contest solely automated decisions. However, the majority of Global South countries have only soft-law frameworks, which can set priorities, build consensus, and create pathways towards stronger regulation but cannot on their own provide binding rights, impose legal obligations, or hold public and private actors accountable for harmful AI deployment.

The result is a growing governance asymmetry. Global South countries are adopting the language of responsible AI but often without the legal instruments needed to make it enforceable. This reflects a global political economy in which many governments are trying to attract investment, build local AI sectors, and secure access to infrastructure dominated by powerful technology companies.

Without enforceable foundations, responsible AI risks becoming a field of soft commitments in precisely the countries where governance capacity is most constrained and external technological dependence is often greatest.

21. European Union. (2024). Regulation (EU) 2024/1689 of the European Parliament and of the Council of 13 June 2024 laying down harmonised rules on artificial intelligence (Artificial Intelligence Act). *Official Journal of the European Union*, L 2024/1689. https://eur-lex.europa.eu/eli/reg/2024/1689/oj/eng
22. The assessment of EU regulations under the GIRAI methodology was addressed through a collaborative process involving all EU country-level researchers. The assessment was led by one of those researchers — with extensive experience in Brussels' debates on AI regulation — guided by the GIRAI core team and subsequently discussed with the broader research team. As a result of those deliberations, the EU AI Act, the EU Digital Services Act, and the EU General Data Protection Regulation were accepted as valid evidence of binding frameworks across 10 indicators: Gender Equality, Children's Rights, Fairness and Non-Discrimination, Transparency and Explainability, Human Oversight and Determination, Labour Protections, Safety and Security, Access to Redress and Remedy, Impact Assessments, and AI-facilitated Misinformation and Violence. These regional regulations account for much of the strong performance of EU member states in this Edition.
23. The GIRAI captures national-level frameworks, including laws, policies, and strategies. To avoid counting international commitments not translated into real protections of rights within national boundaries, international frameworks are only accepted as valid for a country or jurisdiction if they are binding at the national level. For indicators where there are no EU-level regulations, researchers assessed the presence of national frameworks in those countries. The implementation of initiatives relevant to the indicators were assessed only at the national level, regardless of whether the framework applied to that indicator is an international-level or national regulation.
24. Diario Oficial El Peruano. (2025, September 9). 'Supreme Decree approving the Regulations of Law No. 31814, Law that promotes the use of artificial intelligence for the economic and social development of the country: Supreme Decree 115-2025 PCM'. https://web.archive.org/web/20251013161211/https:/busquedas.elperuano.pe/dispositivo/NL/2436426-1
25. Ministry of Justice of the Kyrgyz Republic. (2025, July 31). *Digital Code of the Kyrgz Republic*. https://cbd.minjust.gov.kg/3-48/edition/35412/kg
26. Federal Democratic Republic of Ethiopia (2024). *Proclamation to provide for personal data protection*. Federal Negarit Gazette, Proclamation No.1321/2024. https://web.archive.org/web/20260104184121/https:/eca.et/wp-content/uploads/2024/10/personal_data_protection_proclamation_No_1321_2024.pdf

# Bright Spot: Peru

Binding AI laws offer the strongest protections for vulnerable populations. Peru stands out as one of the Global South countries with the strongest examples of adopted binding AI legislation. The country's Supreme Decree 115-2025-PCM is a comprehensive, binding decree that establishes enforceable obligations across 13 indicators in the GIRAI — the widest adopted coverage of any Global South country. In areas where AI policies are non-binding or at a draft stage in other countries, Peru's frameworks are already in force. Supreme Decree 115-2025-PCM defines guiding principles including non-discrimination, human oversight, sustainability, and transparency (Art. 7) and requires that AI-based systems be subject to human review, ensuring decisions are ethical, fair, and respectful of fundamental rights. It explicitly prohibits misuses of AI, including systems that deceptively or manipulatively influence decision-making through subliminal or deliberately misleading techniques. On vulnerable populations, Article 12 mandates that all AI-supported digital services fully consider the differentiated needs of children, adolescents, women, and girls. The law also categorises certain applications as high-risk AI (Arts. 22 and 24), triggering mandatory prior human rights impact assessments for systems such as AI used in employment, education, and access to essential services. On labour protections, high-risk obligations apply to AI systems that affect working conditions, recruitment, or task allocation. This single comprehensive framework covering several indicators is further complemented by a guideline for the ethical use of AI within the competition authority (INDECOPI).

# 3 AI safety is being governed as a technical problem, while human harms remain under-addressed

AI safety and security is one of the fastest-growing areas of responsible AI governance. Since the 1st Edition, researchers identified approximately 30 countries that did not previously have a framework covering this area but now have one, suggesting that governments are increasingly recognising the need to manage AI-related risks. The increase is geographically broad: 10 of these countries are in Europe, six in Asia and Oceania, six in South and Central America, five in Africa, and three in the Middle East. In Europe, nine of the 10 newly covered countries address the area through the EU AI Act[27].

Much of this governance remains focused on technical safeguards: robustness, reliability, security, resilience, and protection against system failure or misuse. These are essential, but they do not fully address the material harms that AI can create when deployed in public life.

The gap is visible in the Index's evidence on government use of AI with unacceptable risks. Researchers found credible evidence of misuse in 35 of 135 countries, including cases that raise serious concerns for human rights, rule of law, and democratic accountability. More than half involved mass biometric surveillance.

It is also visible in responses to AI-facilitated misinformation and violence. Frameworks in this area exist in only 36% of countries globally. Of these, only 59% show evidence of implementation, meaning that just 29 countries with active frameworks in this field have implemented measures addressing AI-related misinformation and violence.

The findings demonstrate that AI safety is gaining policy attention as a technical measure, but material harms remain under-governed. Governments are beginning to address the technical safety of AI systems, but they are not yet adequately governing the harms people experience through surveillance, misinformation, violence, intimidation, discrimination, and other forms of misuse.

Bangladesh's *Cyber Security Ordinance 2025*[28] is a good reference point for how states can address human harms beyond technical safety governance. The law criminalises the use of digital tools (including an explicit reference to AI in Chapter 6, Arts. 17, 18, 21, and 25) for creating or broadcasting harmful content, including misinformation, sexual harrasment, and child sexual abuse material, among others. Australia's *Criminal Code Amendment (Deepfake Sexual*

27. Europe: Belgium, Bulgaria, Croatia, Estonia, Greece, Hungary, Latvia, Republic of Moldova, Romania, Slovakia.
Asia and Oceania: Bhutan, China, Kyrgyz Republic, Nepal, New Zealand, Pakistan.
South and Central America: Chile, Colombia, Costa Rica, El Salvador, Peru, Uruguay.
Africa: Cote d'Ivoire, Ethiopia, Libya, Morocco, Nigeria.
Middle East: Bahrain, Oman, Qatar.
28. Government of the People's Republic of Bangladesh, Ministry of Law, Justice and Parliamentary Affairs. (2025, May 21). The Cyber Security Ordinance 2025. https://web.archive.org/web/20251008181125/https:/www.dpp.gov.bd/upload_file/gazettes/57587_41613.pdf

*Material) Act*[29] of 2024 similarly creates offences for using a carriage service to transmit sexual material of an adult without consent, explicitly extending liability to material "created or altered in any way using technology" — including AI. Meanwhile, The Gambia provides special protection for children and other vulnerable groups through a distinct policy instrument, *The National Child and Vulnerable Groups Online Protection Policy*[30], recognising the increasing role of AI in exacerbating different forms of risks to these groups and assigning obligations to service providers and law enforcements to mitigate them – see Bright Spot: The Gambia.

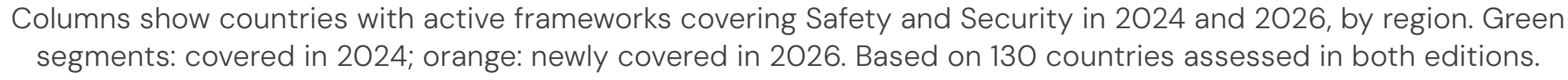
Figure 7: **Growth in Safety and Security Framework Coverage by Region, 2024–2026**

Columns show countries with active frameworks covering Safety and Security in 2024 and 2026, by region. Green segments: covered in 2024; orange: newly covered in 2026. Based on 130 countries assessed in both editions.

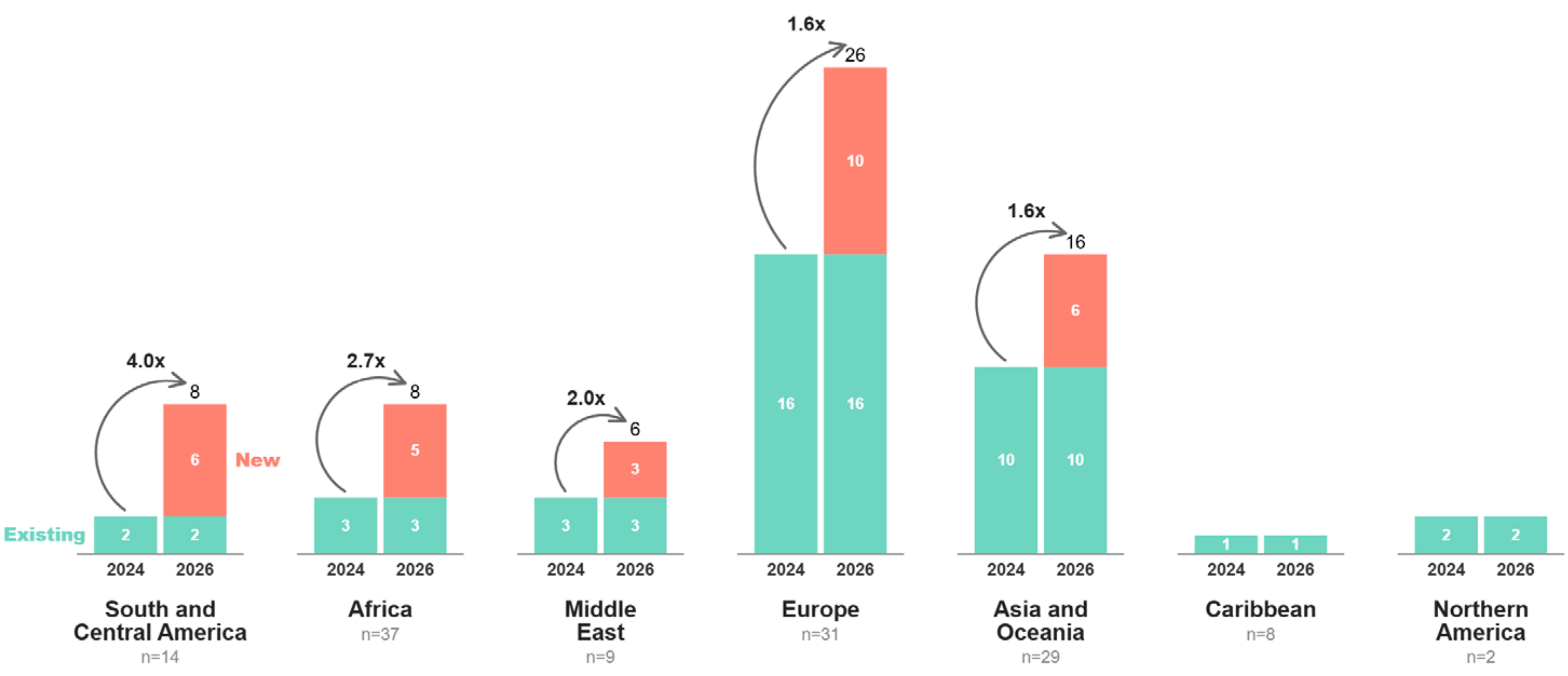

*Source: Global Index on Responsible AI (GIRAI) Global Survey, 2026.*

29. Parliament of Australia. (2024, September 2). *Criminal Code Amendment (Deepfake Sexual Material) Act 2024*. No.78, 2024. https://web.archive.org/web/20251204061605/https:/www.ato.gov.au/law/view/pdf/acts/20240078.pdf
30. Ministry of Communications and Digital Economy (MOCDE), The Gambia. (2024, October). *The Gambia's National Child and Vulnerable Groups Online Protection Policy*. https://mocde.gov.gm/wp-content/uploads/2024/11/Gambia-CVGOP-Policy.docx

Figure 8: **Countries with Recorded Cases of Government Deployment of Unacceptable Risk AI Systems (URAI)**

Each bubble represents a country. Bubble size indicates the number of recorded URAI cases (1–4). Based on 135 countries assessed in the 2026 edition

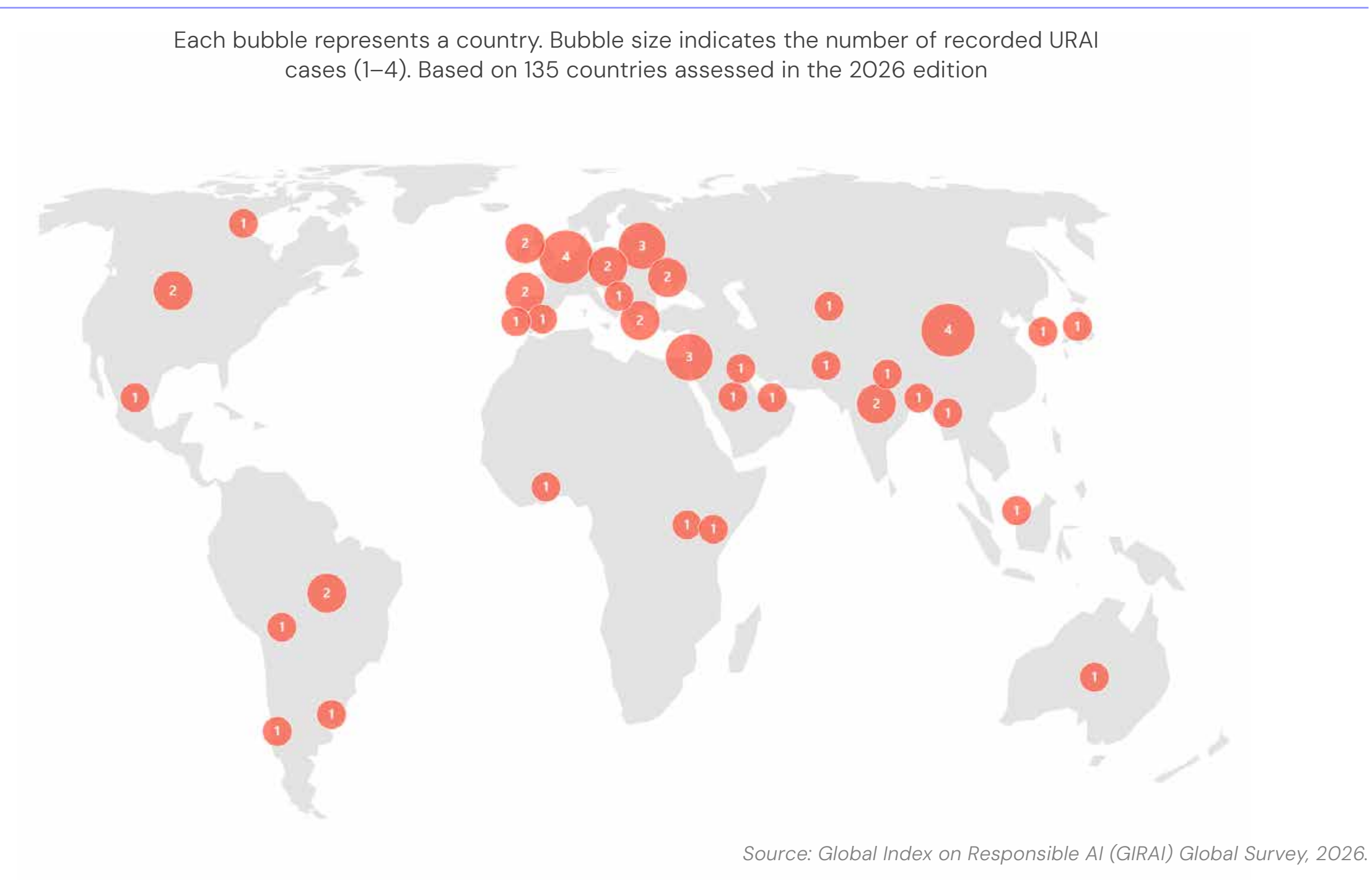

*Source: Global Index on Responsible AI (GIRAI) Global Survey, 2026.*

Figure 9: **Framework Coverage of AI-Facilitated Misinformation and Violence, by Region**

Bars show the share of countries with active frameworks covering AI-facilitated misinformation and violence indicator, by region. The dashed line shows the regional average across all AI Policy indicators; arrows show the percentage-point gap between the regional average of all AI policy indicators and the selected one.

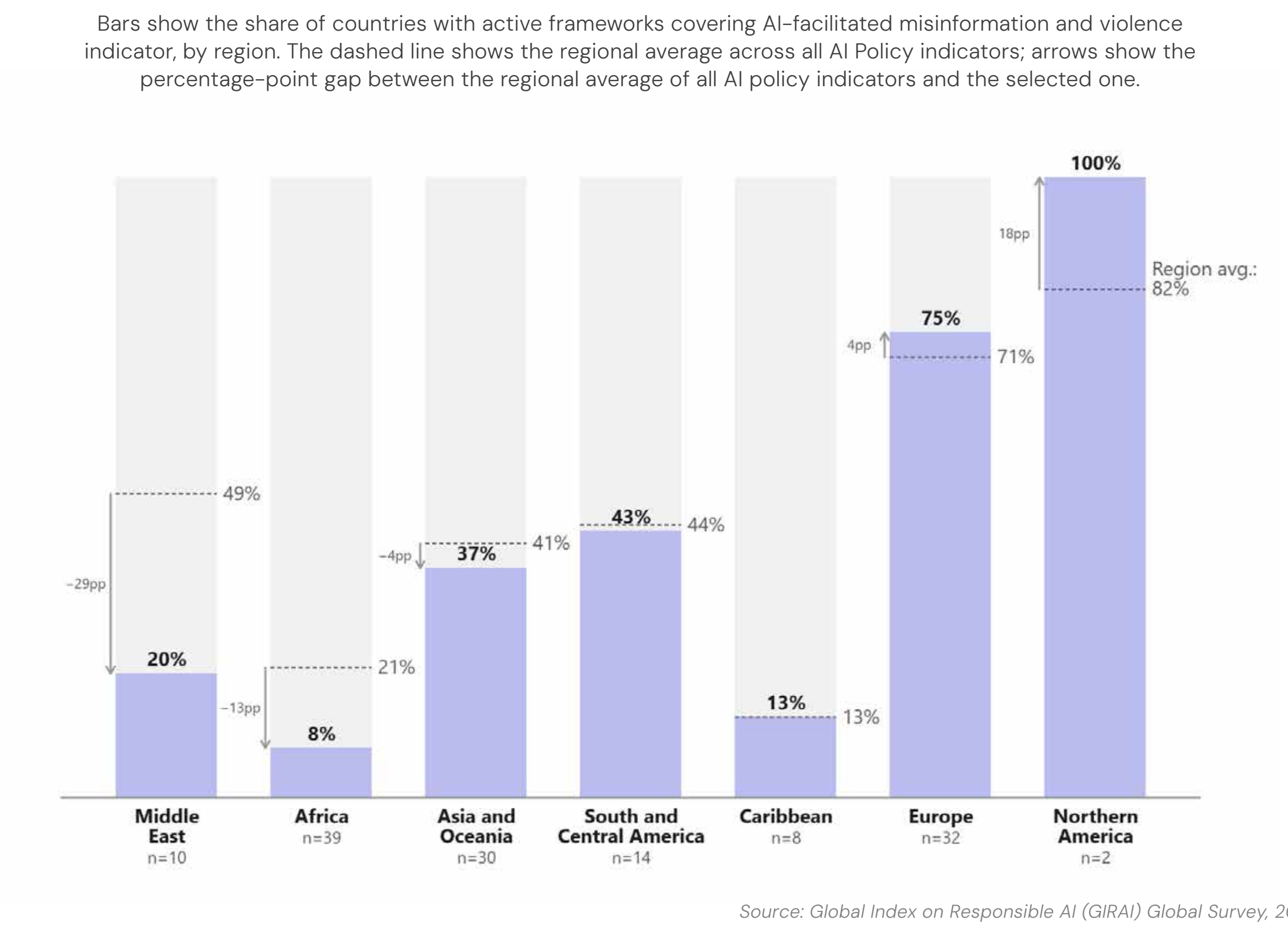

*Source: Global Index on Responsible AI (GIRAI) Global Survey, 2026.*

## Bright Spot: The Gambia

In addition to policies specifying safety and security requirements from providers of AI systems, frameworks that aim to specifically address AI-facilitated misinformation and violence offer the strongest safety guarantee for users. The Gambia stands out as one of the highest-performing Global South countries in this area. Through its National Child and Vulnerable Groups Online Protection Policy, developed by the Ministry of Communications and Digital Economy, The Gambia established the first framework in the country specifically designed to protect children and vulnerable people from online harms, including those facilitated by AI. The policy recognises that there is a wide spectrum of online risks and harms for children, which are exacerbated by AI. It assigns obligations to internet service providers, mobile network operators, product and content developers, and law enforcement. Oversight on the policy is provided by the National Cybersecurity Coordination Directorate. The policy is complemented by a nationwide campaign against misinformation and disinformation launched in January 2025, aimed at raising public awareness about the dangers of false information (including AI-generated misinformation) and promoting responsible media consumption. Although occurring outside the timeline of the 2nd Edition, in 2026 The Gambia also launched the National Misinformation and Disinformation Response Centre (NMDRC), a national mechanism for monitoring, verifying, and fact-checking information circulating on social media.

# 4 Governments are regulating AI transparency but not disclosing their own use of AI

Transparency and Explainability is the strongest-performing indicator in the Index, with 58% of countries having some form of framework in place. Yet it also has one of the largest implementation gaps: in 60% of cases, there is no evidence that these frameworks are being meaningfully operationalised.

The contrast is even sharper when it comes to government use of AI. Public Disclosure of Government Algorithmic Systems is the lowest-performing indicator in the Index, with only 18% of countries requiring disclosure of government AI systems. Three notable examples are Canada, Colombia, and The Netherlands. Canada's *Directive on Automated Decision-Making*[31] requires all federal departments to conduct algorithmic impact assessments and publish information about automated systems used in decisions affecting individuals (sections 6.1.1, 6.5.1, and 6.5.2). This is complemented by the *Government of Canada AI Register*[32], a public inventory of AI use cases across federal agencies. Colombia's *Joint Directive 007 of 2025*[33] establishes active disclosure obligations regarding government algorithmic systems (Arts. 8 and 9) requiring minimum public content such as purpose, data used, and developer - see Bright Spot: Colombia. Similarly, The Netherlands' *Open Government Act (Wet open overheid)*[34] establishes a binding legal basis for public access to government information, including algorithmic decision-making systems. The framework commitment is backed by two concrete implementation tools: The Dutch Algorithm Register[35], which provides a centralised, publicly accessible inventory of all government algorithms, and the *Fundamental Rights and Algorithms Impact Assessment (FRAIA)*[36], which requires government organisations to conduct structured, interdisciplinary assessments of how their algorithmic systems affect fundamental rights before deployment.

Barring these few examples, governments' lack of strong commitment to transparency in their own AI systems points to a clear asymmetry in AI governance. Governments are increasingly adopting transparency requirements for AI systems but not applying the same standards to their own use of AI in public services, policing, migration, welfare, and other areas of public administration. This creates a transparency gap in which government AI deployments are largely

31. Government of Canada (2019, April 1). 'Directive on Automated Decision-Making'. https://web.archive.org/web/20250929225202/https:/www.tbs-sct.canada.ca/pol/doc-eng.aspx?id=32592
32. Government of Canada (n.d.) 'Government of Canada AI Register (Minimum Viable Product) - GC AI Register' (MVP). https://web.archive.org/web/20260409111123/https:/open.canada.ca/data/en/dataset/fcbc0200-79ba-4fa4-94a6-00e32facea6b/resource/369f6f34-148a-42ed-b581-8c164e941a89. Accessed April 9, 2026.
33. Office of the Attorney General and Ombudsman, Republic of Colombia. (2025, September 30). 'Joint Directive 007 of 2025'. https://web.archive.org/web/20251023190531/https:/www.defensoria.gov.co/documents/20123/3407303/300925DirectivaConjunta007.pdf/c47e1175-6f60-058a-3e0b-3dfaf82d5f23?t=1759261267112
34. Ministry of the Interior and Kingdom Relations, Government of The Netherlands. (2021, October 27). *Act of 25 October 2021, laying down rules on the accessibility of information of public interest (Open Government Act).* https://web.archive.org/web/20251215155023/https:/zoek.officielebekendmakingen.nl/stb-2021-499.html
35. See https://algoritmes.overheid.nl/en
36. Ministry of the Interior and Kingdom Relations, Government of The Netherlands. (2022, March). *Impact Assessment: Fundamental rights and algorithms.* https://www.government.nl/documents/2021/07/31/impact-assessment-fundamental-rights-and-algorithms

hidden from public view. Without public disclosure, people may not know when AI is being used, how it affects decisions about their lives, or how to challenge harmful outcomes. This is particularly concerning given the extent of countries where evidence of government misuse of AI was found.

Figure 10: **Framework Coverage Across AI Policy Indicators**

Bars show the percentage of countries with active frameworks covering each AI Policy indicator. The dashed line indicates the average framework coverage across all AI Policy indicators.

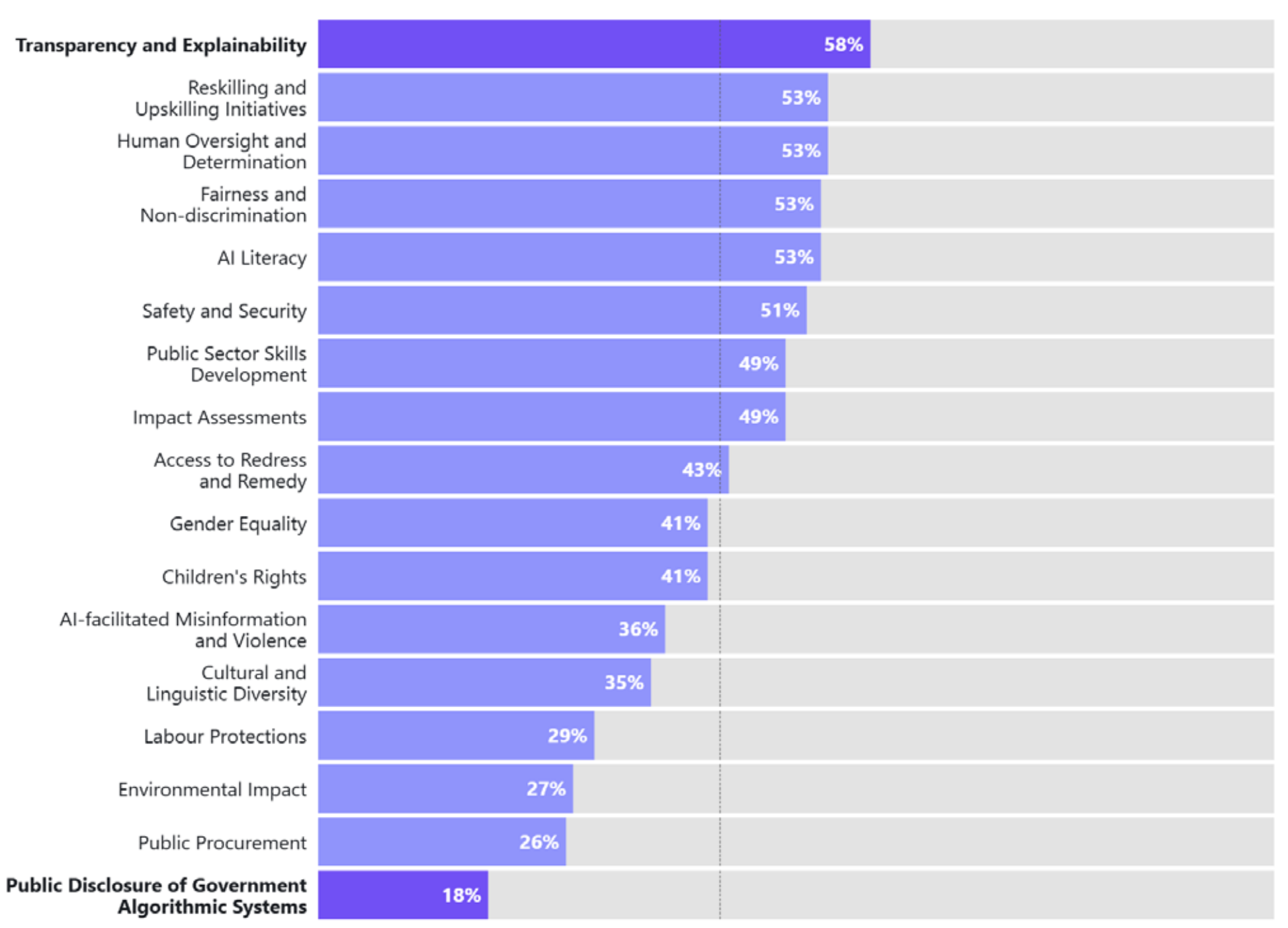

*Source: Global Index on Responsible AI (GIRAI) Global Survey, 2026.*

## Bright Spot: Colombia

Colombia's binding framework "Standards on Algorithmic Transparency for the Algorithmic Systems used by the state (Office of the Attorney General of Colombia and Office of the Ombudsman of Colombia, Joint Directive 007 of 2025)" establishes active disclosure obligations regarding algorithmic systems (Arts. 8 and 9) requiring minimum public content such as purpose, data used, and developer. It also regulates enhanced transparency for high-impact systems (Art. 13), algorithmic impact analysis (Art. 14), and access to source code or meaningful explanation (Art. 17), which consolidates public disclosure mechanisms.

# 5 Gender is increasingly recognised in AI governance, but protection from gendered harms remains weak

Gender equality is becoming more visible in responsible AI governance, with South and Central America standing out in their priorisation of gender, together with Northern America and Europe.

Since the 1st Edition, researchers identified 29 new countries addressing gender equality and AI. As such, it is one of the areas with the biggest growth, yet gender overall continues to rank low, indicating that it remains a structurally deprioritised area.

Some existing frameworks focus on preventing AI systems from reproducing gender discrimination and bias. Others demonstrate government commitments to increase gender equality and women's representation in the AI workforce – an important step not only for identifying and responding to harms but also for ensuring that women and girls actually benefit from the systems being built.

This growing recognition has, however, not translated into strong protection in practice. Gender Equality has one of the widest implementation gaps in the Index: while 55 countries have frameworks addressing gender and AI, only 24 show evidence of implementation activity.

Although some frameworks address remedy and accountability for AI-facilitated violence in general, this is not comprehensively covered for individuals affected by gender inequality or discrimination perpetuated by AI. Among the few countries that do speak to the need for these channels, approaches are not well developed. What is more, few governments are taking steps to address the use of misinformation to facilitate gender-based harassment and violence. These are serious gaps amid the rapid growth of technology-facilitated gender-based violence (TFGBV), including online harassment, stalking, image-based abuse, deepfake sexual content, and other forms of AI-enabled harm[37].

A few countries have targeted provisions. Chile's *National AI Policy (2024 update)*[38] sets out specific measures to stop AI systems from reinforcing gender bias, including the promotion of bias and non-discrimination assessments across the full AI lifecycle – see Bright Spot: Chile. Colombia, through its *Document CONPES 4144 (National Planning Department of Colombia -NDP- 2025)*[39], proposes strategies for addressing gender-based violence in digital environments, while Australia's *Criminal Code Amendment (Deepfake Sexual Material) Act of 2024*[40]

37. Many countries address these offences through traditional legal instruments addressing technology-mediated infractions. Therefore, general legislative frameworks or digital frameworks were accepted when there was demonstrable evidence of such instruments having been applied to an AI-specific context.
38. Government of Chile. (2024). Chilean National Policy on Artificial Intelligence (UPDATE 2024). https://web.archive.org/web/20260226183216/https:/cens.cl/wp-content/uploads/2025/10/Politica-Nacional-de-Inteligencia-Artificial.pdf
39. Departmento Nacional De Planeación, República De Colombia (2025, February 14). *Política Nacional De Inteligencia Artificial* [National Artificial Intelligence Policy (CONPES 4144)]. https://web.archive.org/web/20250826131733/https:/colaboracion.dnp.gov.co/CDT/Conpes/Econ%C3%B3micos/4144.pdf
40. Parliament of Australia. (2024, September 2). *Criminal Code Amendment (Deepfake Sexual Material) Act 2024.*

creates offences for using a carriage service to transmit sexual material of an adult without consent - an act for which women are disproportionately victims. The Dominican Republic offers a unique example of a government implementing initiatives to protect women from gender harms. Its SARA Automated Response and Attention System[41] is a government-created AI chatbot that serves as a first point of contact for women and girls facing gender-based violence, providing guidance on recognising abuse, connecting users to shelters and legal support, and linking them to an emergency line.

AI-facilitated gender harms are already widespread: 38% of women report having experienced online violence and 85% report having witnessed it[42], with actual figures likely being far higher. AI-powered deepfake content heavily targets women and girls with pornographic content while chatbots enable extreme abuse scenarios with almost no safeguards, effectively accelerating the scale and gravity of gendered harm[43]. What statistics are unlikely to capture, beyond the underreporting of harms and justice systems that already struggle to respond to gender-based violence in general and TFGBV specifically, is how multiple forms of marginalisation overlap and compound risks. Social understandings of gendered identities (including LGBTQIA+), race, class, migration status, disability, and other factors inform people's exposure to harm in ways that are difficult to quantify and therefore easily fall off the governance radar. Meanwhile, human rights defenders, activists, journalists, lawmakers, and others working to challenge these inequities and seek justice are increasingly targeted[44], showing how gendered abuse can be used to stifle public participation and democratic engagement.

The overall lack of targeted action by governments clearly reveals that, while gender is increasingly present in responsible AI policy language, the governance response remains partial and inadequate. Frameworks might recognise bias and representation but do far less to provide enforceable obligations, accessible redress, or accountability for gendered harms. Without these protections, gender equality in AI governance risks remaining a signal of responsibility rather than a driver of material change.

41. Ministerio De La Mujer, República Dominicana. (n.d). 'Automated Attention and Response System or SARA: an information assistant on gender and domestic violence'. https://web.archive.org/web/20251008194521/https:/mujer.gob.do/index.php/sobre-nosotros/sara
42. Carmo, A. (2025, November 20). 'AI and anonymity fuel surge in digital violence against women'. *UN News*. https://news.un.org/en/story/2025/11/1166411
43. McGlynn, C., McDermott, Y., Macdonald, S., Toparlak, R.T., Tarrant, F., & Treacy, S. (2026). *Invisible No More: How AI Chatbots Are Reshaping Violence Against Women and Girls*. Swansea University, Durham University and UKRI. https://e87dab74-be98-4bb1-83c5-05251d2bc6f4.usrfiles.com/ugd/e87dab_06a7f0801de549689c294d42e0478a3c.pdf
44. UN Women. (2025). *Global Trends to Prevent and Respond to Technology-Facilitated Violence Against Women and Girls: A Compendium of Emerging Practices*. UN Women Sexual Violence Research Initiative. https://www.unwomen.org/sites/default/files/2025-12/global-trends-to-prevent-and-respond-to-technology-facilitated-violence-against-women-and-girls-en_1.pdf

Figure 11: **Framework Coverage of Gender Equality, by Region**

Bars show the percentage of countries with active frameworks covering the Gender Equality indicator, by region. The dashed line shows the regional average; arrows indicate the percentage-point difference from overall regional framework coverage.

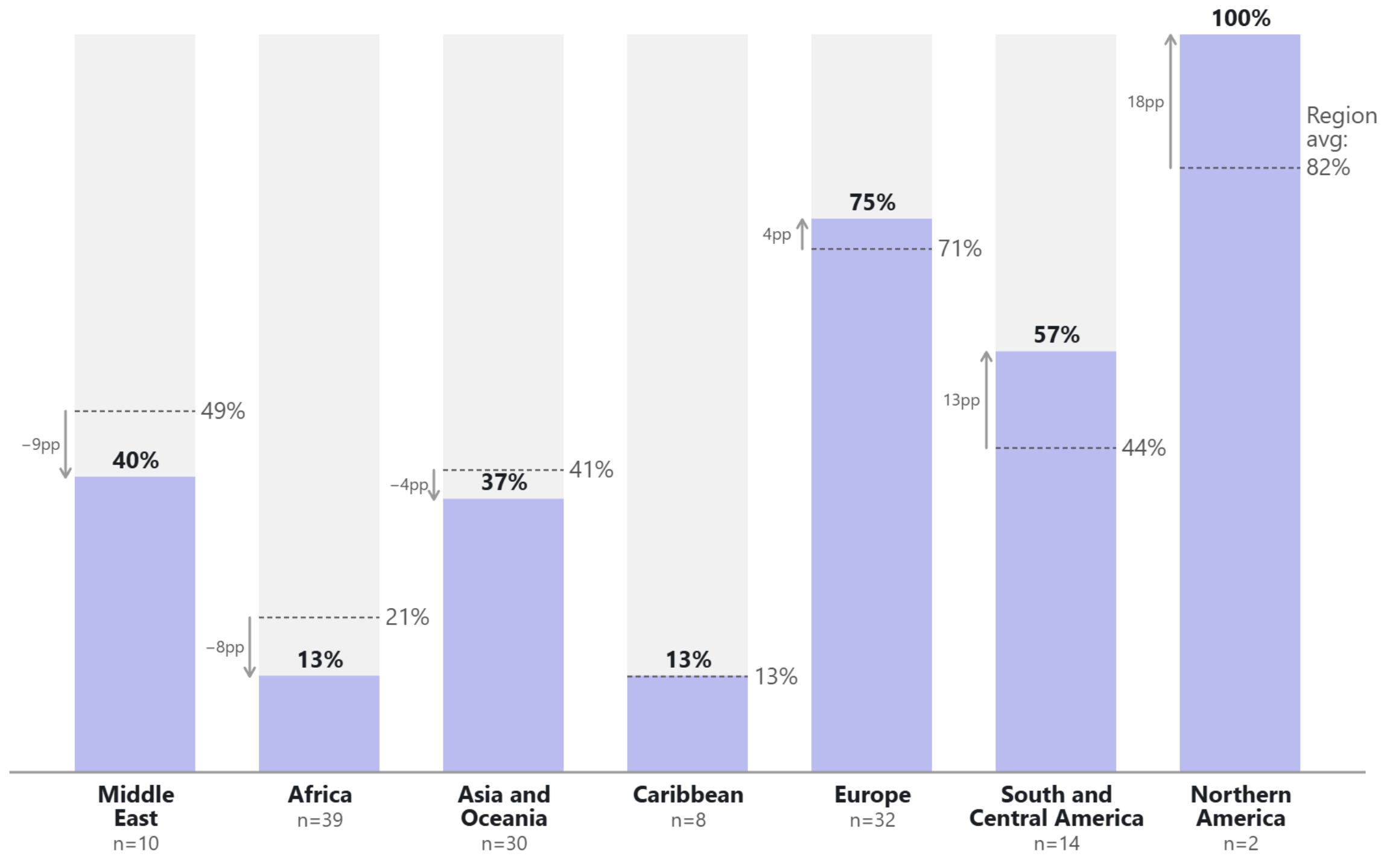

*Source: Global Index on Responsible AI (GIRAI) Global Survey, 2026.*

## Bright Spot: Chile

Chile's updated National Policy on Artificial Intelligence (2024) takes a broader approach to gender equality than simply encouraging women into AI research, development, and industry — it also tackles the risk that automation will widen gender gaps. The policy sets out specific measures to stop AI systems from reinforcing gender bias, including the use of bias and non-discrimination assessments across the full AI lifecycle, requirements that training data and research draw on a diverse range of social groups, and active support for the participation and advancement of underrepresented groups, women among them. Together, these measures aim to reduce gender bias at the data and development stages so that AI contributes to greater equity and inclusion rather than entrenching existing inequalities.

# 6 Future generations are being prepared for the AI economy but not protected from AI-related harms

Governments are increasingly investing in AI literacy as a way to prepare children and young people for the AI economy; 71 countries (53%) have some framework in place. When government frameworks and government-led initiatives are taken together, AI Literacy is the strongest-performing indicator in the Index, with 106 countries showing evidence of activity. Almost all of the frameworks address AI literacy across primary, secondary, and tertiary education.

This is an important development. AI literacy can help children and young people understand, use, question, and shape AI systems. It can support future skills, widen participation, and help prepare students for changing labour markets. But AI literacy alone does not protect children from AI-related harms.

Children's Rights remains one of the weakest and slowest-growing indicators of responsible AI governance. Frameworks addressing children's rights exist in only 55 countries, and just 27 of these show evidence of implementation. Thus, 51% of countries with children's rights frameworks show no evidence that these protections are being put into practice. Across regions, children's rights receive greater priority in Europe, Asia, and Oceania. Further, the regional disparities raise concerns highlighted in the analysis of the Human Rights and AI dimension of the 1st Edition, warning how unequal protection systems in the Global South will entrench geopolitical inequities onto future generations[45]. Considerations of children's rights predominantly focus on the use of AI to exploit, manipulate, or harm children, including via targeted advertisements and inappropriate content.

The prioritisation being afforded to AI literacy by governments around the world, together with the lack of attention many governments are affording to the specific AI-related risks children face, indicates that governments are moving faster to prepare children for AI than to protect them from AI-related harms. Children face specific and foreseeable risks from AI systems, including profiling, targeted manipulation, privacy violations, unsafe recommender systems, AI-facilitated misinformation, image-based abuse, deepfake sexual content, discriminatory educational technologies, and harmful uses of AI in welfare, migration, policing, and child protection systems. These risks are already visible. The Internet Watch Foundation reported 8,029 AI-generated images and videos assessed as showing realistic child sexual abuse in 2025[46]. Teaching children how to navigate AI systems cannot replace the responsibility of governments to ensure that those systems are safe, rights-respecting, age-appropriate, transparent, and accountable.

45. Abdella, S.E. (2025, February). *Global Landscape: Human Rights and AI - A Global Index on Responsible AI Analytical Report Series.* Global Center on AI Governance. https://globallandscape.tiiny.site/
46. Internet Watch Foundation (IWF). (2026). *Harm without limits: AI child sexual abuse material through the eyes of our analysts.* https://admin.iwf.org.uk/media/hl1nvdti/iwf-ai-csam-report-2026.pdf

A few countries are demonstrating that equal commitment to building both skills and systems that are aligned with human rights values is possible through a combination of comprehensive frameworks and credible initiatives. Kazakhstan's *Conceptual Framework for the Implementation of Artificial Intelligence in the System of Secondary Technical and Vocational and Post-Secondary Education for 2025–2029*[47], for instance, is a multipurpose policy that treats AI literacy as a foundational competence for both learners and teachers. It does this while establishing children's protection through pedagogical responsibility for human oversight, mandating privacy and personal data protection considerations, and prohibiting discrimination, particularly for learners with special educational needs.

The Krygyz Republic follows a similar approach, with its Digital Transformation of the *Kyrgyz Republic for 2024-2028*[48] establishing a National Council for AI Development that, among other duties, is tasked with involving schoolchildren and students in the process of AI learning (page 7), while its *Digital Code of the Kyrgyz Republic 2025*[49] establishes safeguards for lawful processing of children's data (Art. 81).

Nigeria's AI strategy makes clear commitment to AI literacy, backing this up through its 3 Million Technical Talent (3MTT) programme that targets skill development across demography, including youth, while establishing protection mechanisms through its *Data Protection Act*[50] and an addendum framework (the General Application and Implementation Directive [GAID] 2025[51]) that specifies safeguards for the processing of children's data - see Bright Spot: Nigeria.

Developing the critical thinking skills needed when engaging with AI - skills rooted in a situated understanding of when these tools are useful and when they are not, as well as how they work - is central to any meaningful approach to AI literacy. When existing AI literacy approaches are assessed, their focus on conveying information in a way that actually resonates with people should be evaluated, as well as who was involved in developing the approach they use. A report by the Global Center on AI Governance[52], informed by expert interviews in South Africa, warns that even very ambitious AI literacy efforts may not account for the contextualised pedagogical strategies needed to deepen people's understanding of AI, particularly children. Without this grounding, initiatives may become hollowed. The report shows that both in policy design and educational technology development, AI literacy tends to be defined in broad generic terms that leave too much room for narrow or technocratic interpretations. Without appropriate guidance, overworked and under-resourced educators, who are often new to the technologies they are meant to educate on, are unlikely to adopt the kinds of context-specific teaching practices that make AI truly intelligible in socio-economically and culturally diverse classrooms. Responsible AI governance must therefore address children in two ways: as future participants in AI-enabled societies and as children whose rights, safety, development, privacy, dignity, and wellbeing require protection in the present.

47. Ministry of Education and Ministry of Digital Development, Innovation and Aerospace Industry, Republic of Kazakhstan. (2025, September 18). 'Joint Order of the Minister of Education of the Republic of Kazakhstan dated 18 September 2025 No. 221 and the Minister of Digital Development, Innovation and Aerospace Industry of the Republic of Kazakhstan On the Approval of the Conceptual Framework for the Implementation of Artificial Intelligence in the System of Secondary, Technical and Vocational, and Post-Secondary Education for 2025–2029'. https://prg.kz/document/?doc_id=39475691&pos=4;201
48. Ministry of Digital Development and Innovative Technologies of the Kyrgyz Republic. (2024, April 5). *Concept: Digital Transformation of the Kyrgyz Republic for 2024–2028*. https://digital.gov.kg/wp-content/uploads/2024/06/konczepcziya-czifrovoj-transformaczii-kyrgyzskoj-respubliki-na-2024-2028-gody.pdf
49. Ministry of Justice of the Kyrgyz Republic. (2025, July 31). *Digital Code of the Kyrgz Republic.*
50. Federal Government of Nigeria (2023, July 1). *Nigeria Data Protection Act, 2023*. https://web.archive.org/web/20260101112811/https:/cert.gov.ng/ngcert/resources/Nigeria_Data_Protection_Act_2023.pdf
51. Nigeria Data Protection Commission. (2025). 'Nigeria Data Protection Act (Ndp Act) 2023: General Application And Implementation Directive (Gaid) 2025.' https://web.archive.org/web/20260309110501/https:/ndpc.gov.ng/wp-content/uploads/2025/07/NDP-ACT-GAID-2025-MARCH-20TH.pdf
52. Junck, L. (2026). AI literacy and digital knowledge-making in South Africa: Towards an educator-centred approach. Global Center on AI Governance.

Figure 12: **Framework and Initiative Coverage for AI Literacy and Children's Rights**

Bars show the number of countries with active frameworks and government-led initiatives addressing AI Literacy and Children's Rights, out of 135 countries assessed in 2026.

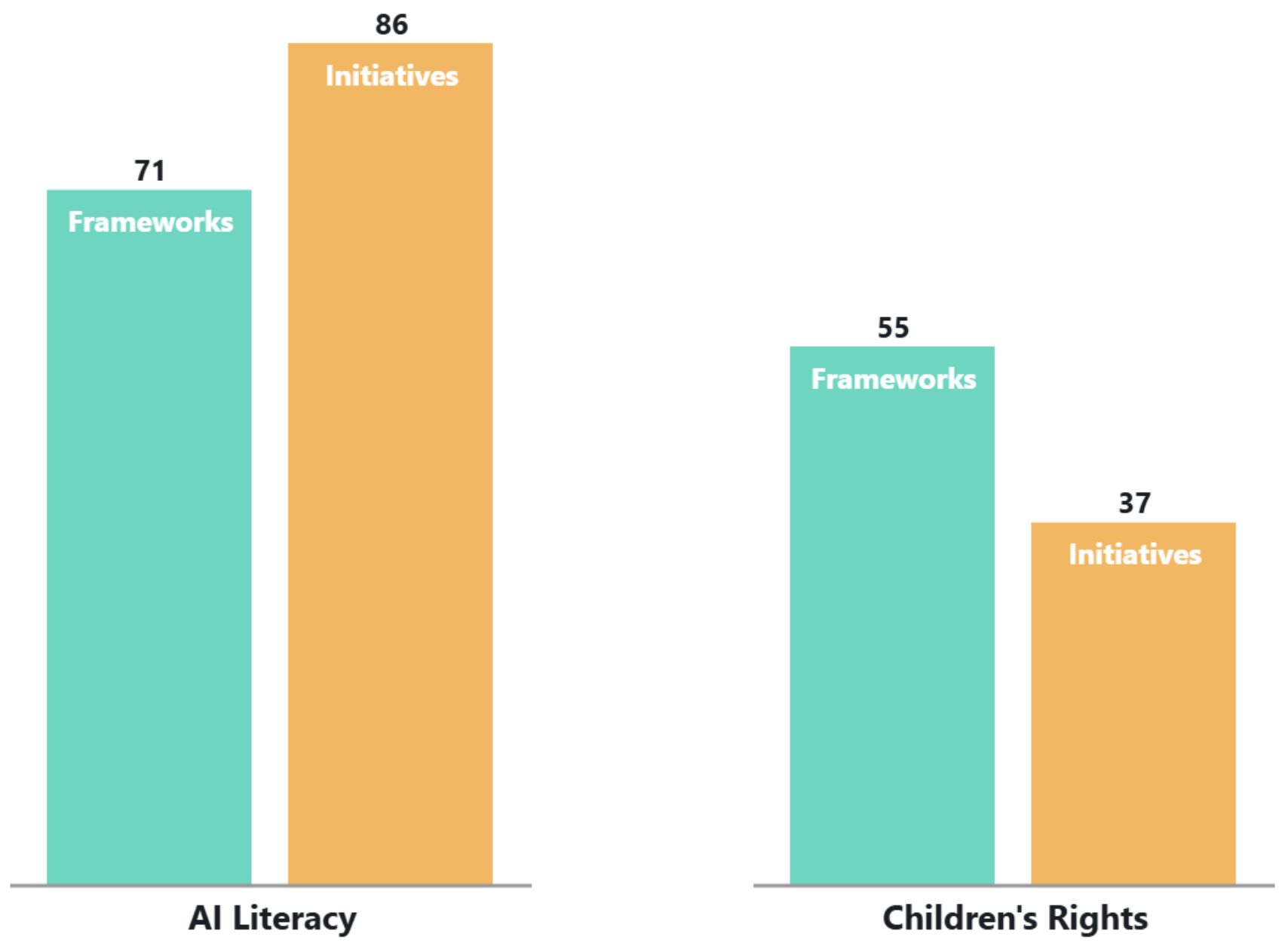

*Source: Global Index on Responsible AI (GIRAI) Global Survey, 2026.*

# Bright Spot: Nigeria

With children's rights receiving limited priority in the African region, Nigeria is one of the few African countries that has attempted to close the gap between skills development and AI harm protections for children. On AI literacy, the National Artificial Intelligence Strategy 2025 calls for increasing AI literacy among all stakeholders and mandates a comprehensive AI skills development programme including teacher training. This commitment is backed by the 3 Million Technical Talent (3MTT) programme, which delivers structured training in AI and machine learning at scale through a hybrid model designed to reach youth populations across the country. On the protection element, the Nigeria Data Protection Act is a binding law that requires parental or guardian consent for processing children's personal data and prohibits subjecting any data subject, including children, to decisions based solely on automated processing. The Act's implementation is further operationalised through the General Application and Implementation Directive (GAID) 2025, which establishes specific safeguards for children's data, requiring data controllers to implement enhanced protections when processing personal data of minors and other vulnerable groups. With adopted frameworks and active initiatives spanning both AI literacy and children's protections (and a binding legal instrument underpinning the latter), Nigeria illustrates combined efforts at both preparing young people for AI and safeguarding them from its harms.

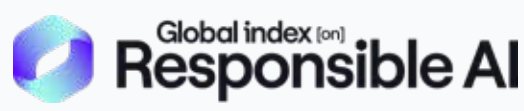

# 7 AI's environmental footprint remains a blind spot in responsible AI governance

AI's environmental impact remains one of the least regulated areas of responsible AI governance. Only 27% of countries have frameworks addressing AI's environmental effects, and 83% of those frameworks are non-binding. While some countries speak to the resource-intensive nature of AI development and deployment, including excessive energy consumption, mounting e-waste, increasing water demand, and the need for energy efficiency of AI models in data centres, there are few robust governance mechanisms. Among Global North countries, addressing the environmental impact of AI is critically unprioritised.

Figure 13: **Framework Coverage Across AI Policy Indicators in the Global North and Global South**

Dots show the percentage of countries with active frameworks per indicator in the Global North and Global South; connecting lines show the gap.

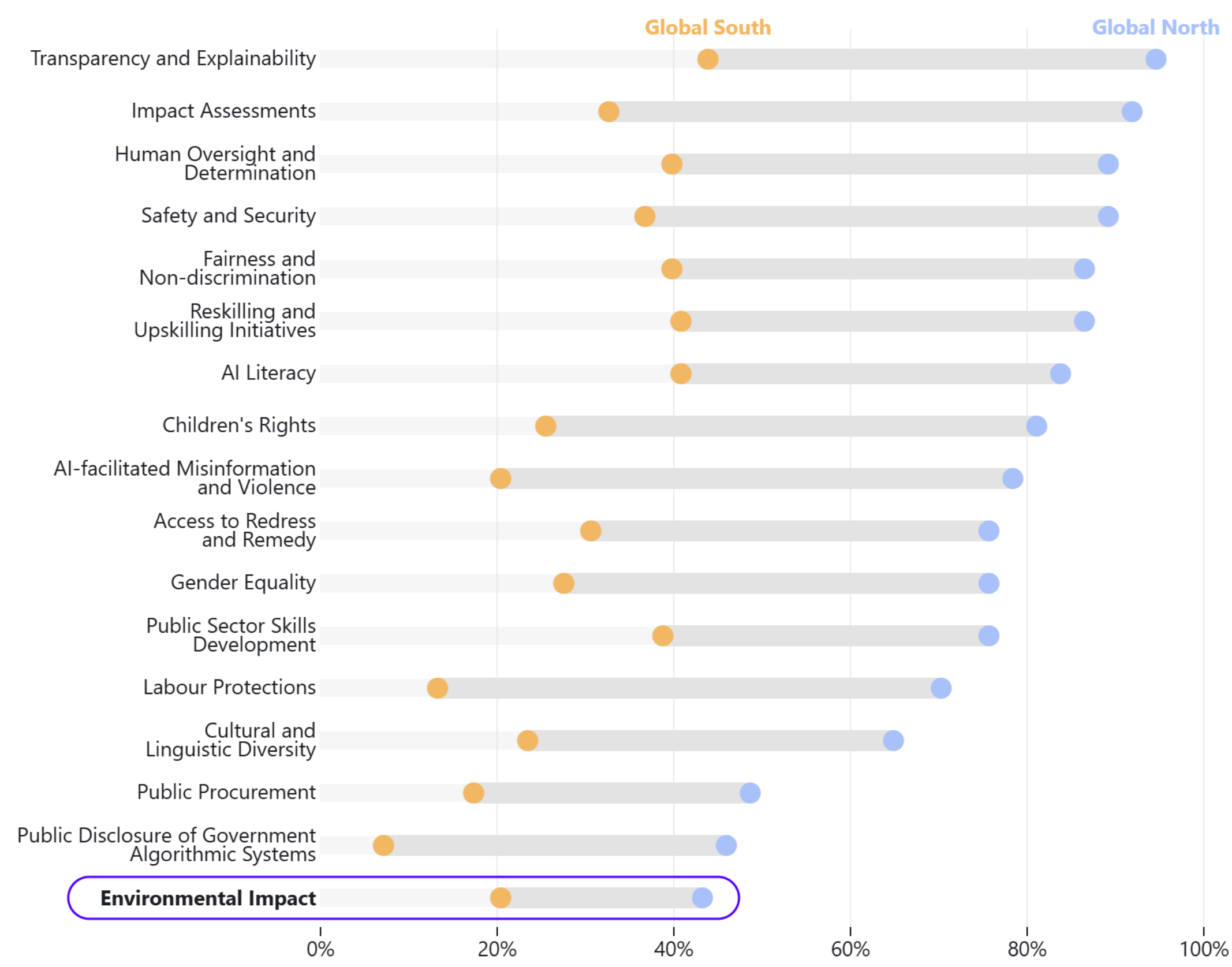

*Source: Global Index on Responsible AI (GIRAI) Global Survey, 2026.*

Global environmental governance is also overwhelmingly unenforceable; only six of the 36 existing frameworks have any legal force. With environmental safeguarding among the weakest areas of enforceability, governance actions clearly lack the leverage needed to influence industry behaviour and protect affected communities.

This is a significant governance gap given the growing environmental footprint of AI infrastructure and lifecycles[53]. A report of the UK's Government Digital Sustainability Alliance Planetary Impact Working Group[54] warns that AI-related water demand could rise from 1.1 billion to 6.6 billion cubic metres by 2027. Increased use of electricity and water for AI diverts these resources from other public and social needs, placing additional pressure on communities already facing energy insecurity, water scarcity, and climate stress. An instructive example of government action is Kenya's decision to discontinue a large-scale geothermal (AI and cloud) data centre partnership due to the facility's projected energy needs – nearly one-third of Kenya's current power generation capacity[55].

AI is known to be a major driver of soaring data centre energy demands, but its precise environmental footprint and growth remain hidden because operators are not disclosing the data needed to distinguish between AI-specific and other workloads[56]. The Index demonstrates that few countries address this crucial oversight by assigning obligations to developers to measure and report the energy use and environmental impact of their AI models. Panama is one of those countries; its proposed *Bill of Law 413*[57] (yet to be adopted as of the time of writing) establishes environmental obligations for AI developers. Articles 6.4 and 17 stipulate that public and private institutions developing AI systems must strive to reduce their carbon footprint, minimise climate change and environmental risk factors, and prevent the unsustainable exploitation and transformation of natural resources. Viet Nam provides more precise requirements through its *Decision 36/QD-TTg of the Prime Minister*, which sets out green standards for data centres and a requirement for a Power Usage Effectiveness (PUE) below 1.4 for greenfield data centres[58]. A small number of countries, including Spain, are beginning to move beyond rhetorical commitments by embedding concrete environmental obligations into their AI strategies – see Bright Spot: Spain.

While 36 countries in the GIRAI have some framework addressing AI's environmental impact, few of these stipulate enforceable obligations for developers. The lack of clear obligations leaves governments, CSOs, and the public with only a partial picture, making climate justice claims easier to deny. Additionally, the Index reveals an absence of government efforts to strengthen the rights of communities and organisations to be informed about the environmental damage they are exposed to through large-scale AI projects, particularly their right to advocate for more

53. d'Orgeval, A., Assoumou, E., Sessa, V., Colak, I., Sheehan, S., & Avenas, Q. (2024, September). Carbon footprint of AI data centers: a life cycle approach. In Proceedings of the 16th International Conference on Applied Energy (pp. 112-123). https://www.energy-proceedings.org/carbon-footprint-of-ai-data-centers-a-life-cycle-approach/
54. Kenny, R. (n.d.). 'Report: Water use in AI and Data Centres'. Government Digital Sustainability Alliance (GDSA) Planetary Impact Working Group UK. https://assets.publishing.service.gov.uk/media/688cb407dc6688ed50878367/Water_use_in_data_centre_and_AI_report.pdf
55. Adeyemi, S. (2026, May 6). 'Kenya suspends $1 billion Microsoft data centre as energy shortfall raises doubts over Africa's AI ambitions' *Business Insider Africa*. https://africa.businessinsider.com/local/markets/kenya-suspends-dollar1-billion-microsoft-data-centre-as-energy-shortfall-raises/hdtfmxz
56. de Vries-Gao, A. (2026). The carbon and water footprints of data centers and what this could mean for artificial intelligence. *Patterns*, 7(1). https://www.cell.com/patterns/fulltext/S2666-38992500278-8
57. Asamblea Nacional, Panama (2025, March 19). 'Regulating The Use Of Artificial Intelligence In The Republic Of Panama (Bill413)'. https://web.archive.org/web/20260224183425/https:/panamabankingnews.com/wp-content/uploads/2025/03/AdL-339.-Marco-regulatorio-de-la-IA-en-Panama-1.pdf
58. The Prime Minister, Republic of Vietnam. (2024, January 11). 'Decision No. 36/QD-TTg of the Prime Minister: Approving the Information and Communication Infrastructure Planning for the period 2021-2030, with a vision to 2050'. https://web.archive.org/web/20251210133232/https:/vanban.chinhphu.vn/?pageid=27160&docid=209501

sustainable practices. This vacuum of data and transparency directly prevents meaningful and potentially impactful responses as demands for protection are starved of the evidence and visibility needed to counter practices backed by massive investments.

This is reflected in the finding that few CSOs globally are working on addressing the environmental impact of AI. Civil society needs sustained backing to support communities in responding to threats to their life sources – especially in Global South countries, which are often at the losing end of extractive AI supply chains but where government actions are still comparatively low. This demand was explicitly made by civil society actors ahead of the AI Action Summit calling for concrete mechanisms, including for stakeholders in the AI discourse to dedicate all necessary means to bring AI systems in line with planetary boundaries[59].

Responsible AI cannot fulfil its purpose if it overlooks the material costs of AI to communities and ecosystems. Environmental governance must become a core part of responsible AI, with binding reporting obligations, public disclosure, enforceable standards, and meaningful community participation.

59. Association for Progressive Communications (ACP). (2025, February 5). 'Joint statement from civil society for the AI Action Summit'. https://www.apc.org/en/pubs/joint-statement-civil-society-ai-action-summit

# Bright Spot: Spain

Spain's 2024 Artificial Intelligence Strategy addresses environmental impact by integrating sustainability requirements into AI infrastructure and model development, notably through measures to reduce the energy, water, and carbon footprint of data centres and AI training and inference (Lever 2 pp. 19–21). It promotes "sustainable AI" and "Green AI" via standards, voluntary certification schemes, and assessment frameworks that explicitly include environmental impact as an evaluation criterion for AI systems. The document also highlights AI applications for environmental and climate-related goals such as energy efficiency, renewable energy optimisation, water resource management, agriculture, and biodiversity protection. In addition to this framework, Spain has the National Plan for Green Algorithms, which tackles AI's environmental impact by standardising how energy use is measured and compared across the AI lifecycle (design, development, training, and execution) and by funding tools to help developers quantify and reduce energy consumption. It also provides research funds and the creation of two academic chairs for "Green Artificial Intelligence and Algorithms" (p. 16).

# 8 Governments are recognising the need for local-language AI but not requiring developers to deliver it

Governments are increasingly recognising that cultural and linguistic diversity is central to whether AI is locally appropriate and accessible to those most in need. This is especially important in Global South countries, where many people speak languages that are poorly represented in dominant AI systems, but also in Global North nations whose languages remain under-represented in global AI development.
The Index finds some meaningful government activity in Cultural and Linguistic Diversity, although it is one of the least covered indicators in terms of framework existence and is one of the slowest growing indicators since the 1st Edition. 52 countries have government-led initiatives focused on promoting local cultural and linguistic identities in the context of AI or preventing AI tools from reinforcing harmful stereotypes or excluding populations based on those identities. Exploratory analysis shows that many of these efforts focus on local-language technologies, multilingual datasets, natural language processing (NLP) tools, and national or regional language models. A noteworthy early-stage initiative is the AI Language Proficiency Monitor[60] backed by the German government, which aims to track how well leading LLMs perform across up to 200 languages — with particular attention to low-resource ones. Rather than building a new model, it functions as a benchmarking tool that maps where language technology succeeds and where it fails, offering a foundation for more equitable development. These initiatives suggest growing recognition that, to be universally beneficial, responsible AI must be culturally grounded and linguistically accessible.

But this recognition via government initiatives is not yet translating into concrete requirements for those designing, developing, and deploying AI systems. Only 47 countries have frameworks addressing cultural and linguistic diversity, and in most cases they refer to the promotion of these values rather than to concrete mandates to the different actors involved in the AI life cycle. The impact of such a coverage gap is distinct in different contexts and geographies, as research[61] indicates that the world's poorest populations are increasingly using AI to access essential services such as healthcare and education. Global South countries' prioritisation of linguistic inclusion and cultural appropriateness therefore does not merely provide access to AI tools but contributes to improving livelihoods and addressing access barriers for those most in need.

Even where frameworks exist, few countries require entities developing or using AI to ensure their systems are trained on diverse datasets and are adaptable to different cultural and linguistic contexts.

There is also very limited attention to how AI systems may affect cultural diversity itself. Among countries with some framework coverage, one of the least prioritised issues is the need to evaluate the impact of AI

60. Pomerenke, D., Nothnagel, J., & Ostermann, S. (2025). The AI Language Proficiency Monitor--Tracking the Progress of LLMs on Multilingual Benchmarks. arXiv preprint arXiv:2507.08538. https://arxiv.org/pdf/2507.08538
61. Björkegren, D. (2026, April 21). '(How) Do the Poorest Use AI?'. Center for Global Development. https://www.cgdev.org/blog/how-do-poorest-use-ai

systems – including translation tools, voice assistants, and other NLP applications – on local cultural diversity, minority languages, and the cultural nuances of human expression. Similarly neglected are mechanisms for remedy and redress for individuals or communities harmed by AI systems that misrepresent or discriminate against their cultural or linguistic identities. Notably, New Zealand's *Responsible AI Guidance for the Public Service* places strong emphasis on cultural diversity, particularly the protection of Māori, Pacific peoples, and ethnic communities. It mandates government agencies to assess the impact of generative AI on these groups, consult with iwi Māori, and respect Māori data sovereignty while also encouraging Māori-led and community-informed AI applications[62]. Singapore, through SEA-LION (Southeast Asian Languages in One Network), supports the development of multimodal large language models that are more attuned to Southeast Asia's cultural contexts and linguistic diversity. The country has gone a step further to conduct safety testing on AI systems in across 10 Asia-Pacific languages and nine countries[63], using a specially developed bias taxonomy and red teaming methodology – see Bright Spot: Singapore. Similarly, Kenya's National AI Strategy 2025-2030 identifies cultural preservation as a guiding principle and explicitly highlights NLP as a key tool for promoting the country's linguistic diversity, with research aimed at developing chatbots capable of interacting in various local languages. The strategy frames the development of local language models as an opportunity to democratise access to AI while ensuring that AI systems are enriched with Kenyan cultural values and actively contribute to preserving the nation's cultural heritage[64].

These findings demonstrate that governments are increasingly aware of the importance of promoting culturally and linguistically responsive AI, but they are not yet requiring AI developers and deployers to build it, test for it, or remedy harm when systems fail.

62. Internal Affairs, New Zealand Government. (2025, February). *Responsible AI Guidance for the Public Service: GenAI.* https://www.digital.govt.nz/assets/Standards-guidance/Technology-and-architecture/AI/Responsible-AI-Guidance-for-the-Public-Service-GenAI-Print.pdf

63. Infocom Media Development Authority. (2025). *Singapore AI Safety Red Teaming Challenge: Evaluation Report.* https://web.archive.org/web/20251215133217/https:/www.imda.gov.sg/-/media/imda/files/about/emerging-tech-and-research/artificial-intelligence/singapore-ai-safety-red-teaming-challenge-evaluation-report.pdf

64. Government of Kenya. (2025, March). *Kenya Artificial Intelligence Strategy, 2025-2030.* http://web.archive.org/web/20250904204355/https://ict.go.ke/sites/default/files/2025-03/Kenya%20AI%20Strategy%202025%20-%202030.pdf

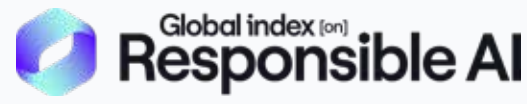

Figure 14: **Countries with Active Frameworks and Government-Led Initiatives, by AI Policy Indicators**

Each point represents an AI Policy indicator. The horizontal axis shows countries with active frameworks; the vertical axis shows countries with government-led initiatives (including where no framework exists). Cultural and Linguistic Diversity, highlighted in purple, is one of the few indicators where activities outperform frameworks.

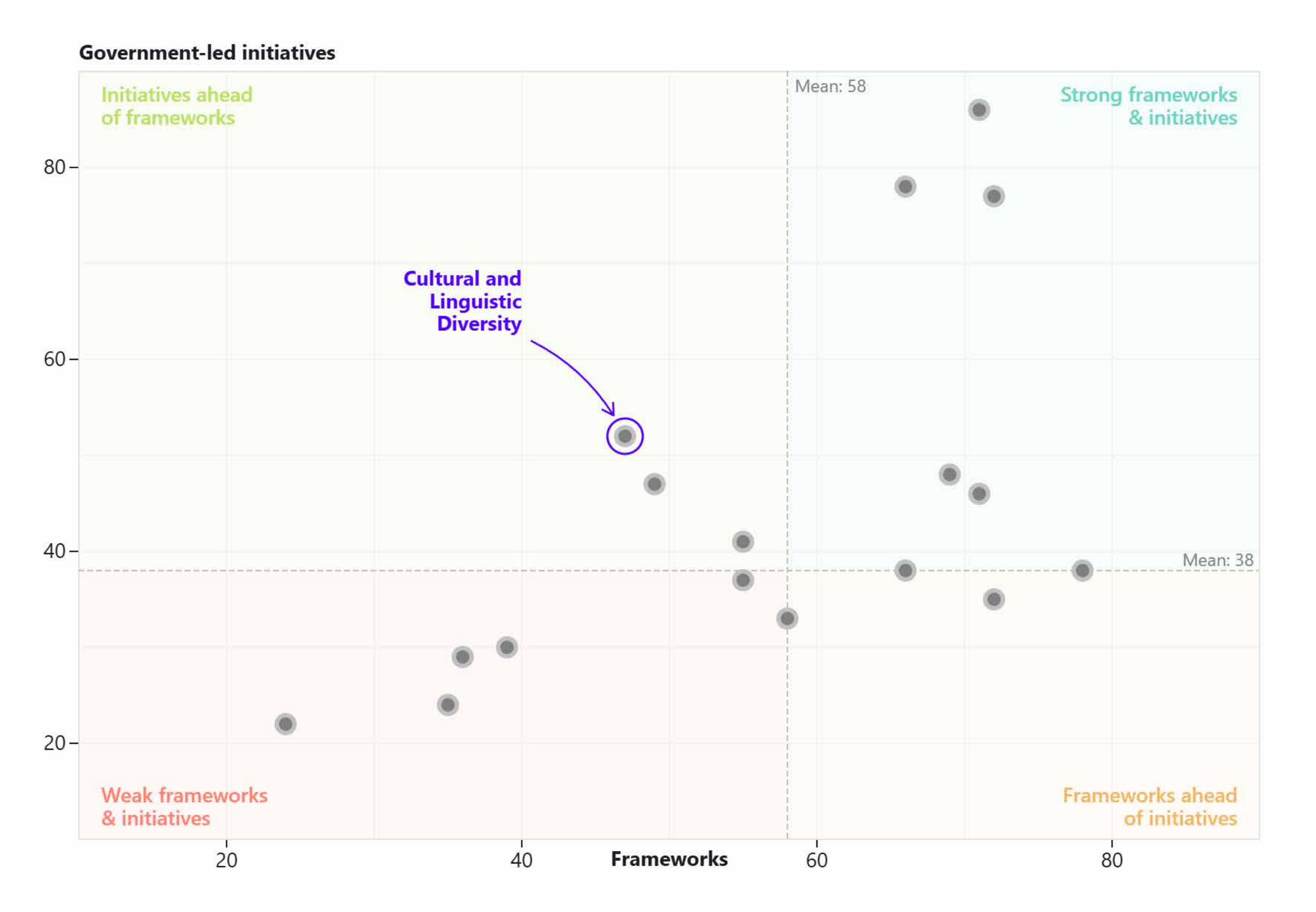

*Source: Global Index on Responsible AI (GIRAI) Global Survey, 2026.*

# Bright Spot: Singapore

A useful bright spot is Singapore, where responsible AI practice is beginning to address cultural and linguistic diversity as a matter of access, representation, and safety evaluation.

Singapore's National Multimodal Large Language Model Programme provides one example. The SGD 70 million programme supports the development of multimodal large language models that are more attuned to Southeast Asia's cultural contexts and linguistic diversity. It builds on SEA-LION (Southeast Asian Languages in One Network), which supports 13 regional languages, including Thai, Vietnamese, and Indonesian, and is designed to better handle multilingual contexts where users move between languages.

Singapore's AI Safety Red Teaming Challenge Evaluation Report is another example. The 2025 exercise tested AI systems across 10 Asia-Pacific languages and nine countries, using a cultural bias taxonomy and red-teaming methodology tailored to regional contexts. The findings showed that safety guardrails were weaker in regional languages than in English, with more successful exploits occurring in regional languages than in English.

Singapore shows comprehensive efforts combining investment in local language model development, testing systems across diverse cultural and linguistic contexts, and identifying where safety, bias, and representation failures emerge outside English-dominant settings.

# 9 Governments are investing in AI skills but neglecting workers' rights

Governments are more likely to invest in reskilling than to protect workers from the harms of AI at work. The 2nd Edition finds that while skills-related policies are relatively common, labour protections remain one of the least addressed areas of responsible AI governance.

Labour protections frameworks were found in only 39 countries (29%). By comparison, 72 (53%) countries have frameworks to support reskilling and upskilling in the context of AI.

Growth on labour protections since the 1st Edition has also been limited. Only 15[65] additional countries have added labour protection framework coverage, making this one of the slowest-growing areas of responsible AI governance and one of the worst performing indicators across both the 1st and 2nd Editions. Labour Protections also remains among the weaker indicators for government-led initiatives, with evidence of implementation for only 46% of the frameworks.

The protections that do exist are important but partial. 35 countries address AI-driven discrimination in hiring, promotion, termination, or algorithmic management, and 34 include transparency requirements for AI-enabled management. Only a few outstanding exceptions – Mexico, Singapore, Uruguay, and Chile – address workers' rights to organise and collectively bargain in response to AI-driven workplace change, specifically targeting platform workers.

These gaps are especially concerning because AI-related labour harms are already visible across both AI supply chains and platform work, particularly in Africa, South and Central America, and Asia. Data labellers and content moderators have reported low pay, exposure to traumatic material, intense productivity targets, inadequate mental health support, and limited routes to organise or challenge conditions[66]. At the same time, algorithmic management systems are increasingly used across industries to allocate tasks, monitor performance, set pay, evaluate workers, and discipline or dismiss them, often through opaque systems workers struggle to contest[67]. Research from the World Bank estimates that there are 154 million to 435 million online gig workers[68] globally, representing 4.4% to 12.5% of the global labour force, and that gig work is growing particularly quickly in Global South countries with demand rising 41% between 2016 and early 2023 alone[69].

The gap in labour protections is even more pronounced across regions. Labour

65. Austria, Bulgaria, China Hong Kong Special Administrative Region, Colombia, Ethiopia, France, Latvia, Lithuania, Mexico, Peru, Republic of Moldova, Romania, Slovakia, United Kingdom of Great Britain and Northern Ireland, Uruguay.
66. Abdelkadir, N. A., Yang, T., Kapania, S., Malgwi, K. I., Berhane Gebrekidan, F., Dinika, A. A., ... & Chancellor, S. (2026, April). Beyond Content Exposure: Systemic Factors Driving Moderators' Mental Health Crisis in Africa. In *Proceedings of the 2026 CHI Conference on Human Factors in Computing Systems* (pp. 1-18). https://dl.acm.org/doi/full/10.1145/3772318.3791639
67. Karimova, T. (2026, April). *AI Systems at Work: A Changing Psychosocial Work Environment*. [ILO Working Paper 170]. Geneva: International Labour Organization. https://www.ilo.org/sites/default/files/2026-05/9789220434086_PDF_Web_EN.pdf
68. While gig work, often in the Global South, can include different forms of labour, it is increasingly linked to AI systems, either through direct data work that trains AI systems or through AI-driven platform management determining workers' conditions and incomes.
69. Datta, N., Rong, C., Singh, S., Stinshoff, C., Iacob, N., Nigatu, N. S., ... & Klimaviciute, L. (2023). *Working without borders: The promise and peril of online gig work* (No. 32573393). The World Bank. http://hdl.handle.net/10986/40066

protection coverage is strongest in Europe, where 78% of countries have frameworks in place, but weakest in regions more exposed to outsourced AI labour. In Africa, only 5% of countries[70] have labour protection framework coverage; in the Caribbean, no countries have coverage on this topic. Coverage is also substantially below average in Asia and Oceania, where only 13% of countries have frameworks addressing labour protections, and slightly below average in South and Central America, where 43% of countries have frameworks in place. Altogether, only 13% of Global South countries have AI labour protection frameworks in operation. This notably includes Mexico, where the binding *Decree Adding Provisions to the Federal Labor Law* Regarding Digital Platforms[71] requires platform companies to make assignment algorithms transparent and develop an algorithmic work management policy explaining several components of the algorithm, including the impact of customer ratings, factors affecting the frequency and pace of work assigned, and rules affecting task assignment (Art. 291-J). Similarly, *Uruguay's Law 20.396/2025*[72] mandates minimum remuneration, working time limits, occupational accident coverage, and social security access for workers who carry out their tasks through digital platforms, including those engaged in autonomous employment relationships – see Bright Spot: Uruguay.

Figure 15: **Countries with Active Frameworks and Government-Led Initiatives in Labour Protections and Reskilling and Upskilling Initiatives**

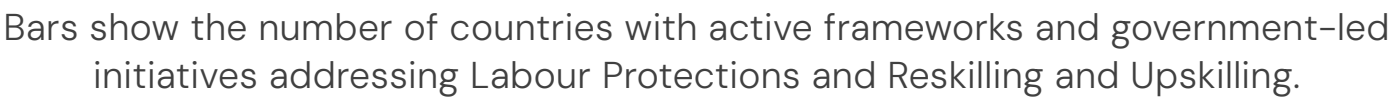

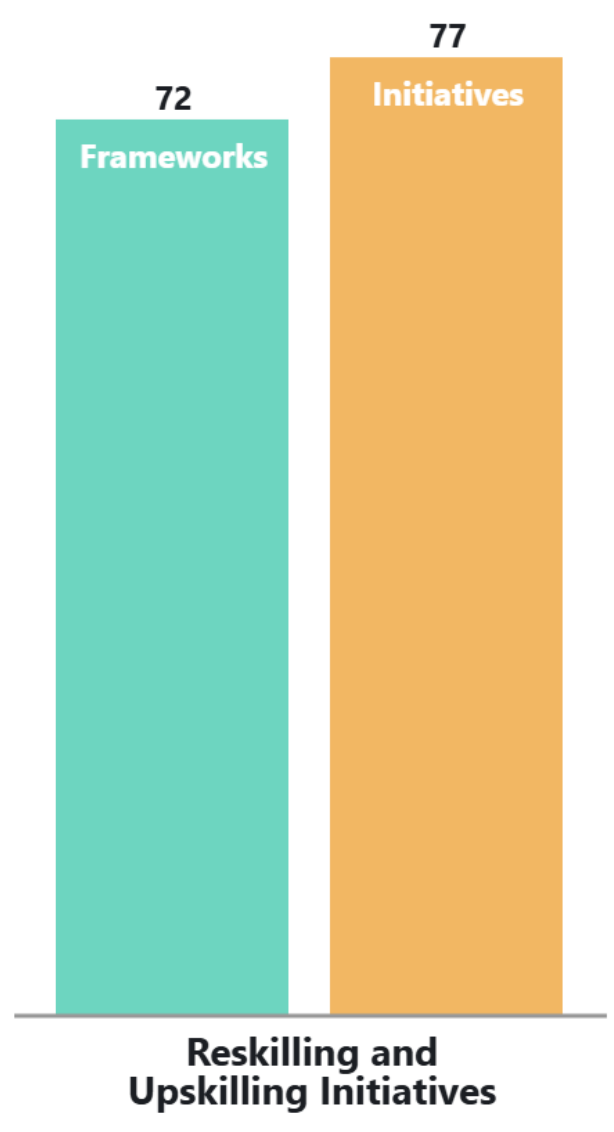

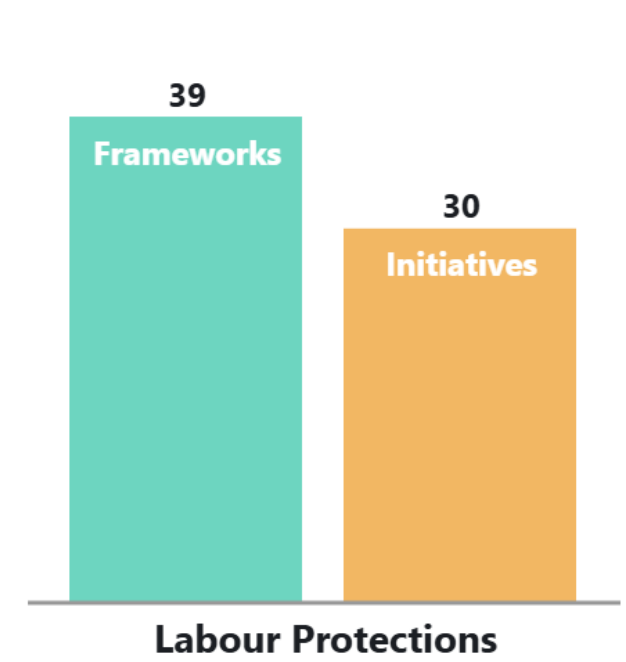

*Source: Global Index on Responsible AI (GIRAI) Global Survey, 2026.*

70. Egypt, Ethiopia.
71. Senadora De La República, Del Grupo Parlamentario Morena, Mexico. (2024, November 5). 'Initiative with Draft Decree to Reform the Articles 153-CY and 153-K of the Federal Labour Law.' https://web.archive.org/web/20250602215247/http:/sil.gobernacion.gob.mx/Archivos/Documentos/2024/11/asun_4799388_20241106_1730916446.pdf
72. Ministry of Labor and Social Security, Uruguay. (2025, February 18). 'Law 20.396 - Minimum levels of protection are established for workers who perform tasks through digital platforms, ensuring fair, decent and safe working conditions'. https://web.archive.org/web/20251111195227/https:/www.impo.com.uy/bases/leyes-originales/20396-2025

This gap in the Global South shows that, although responsible AI governance is focused on preparing workers for AI, it is not protecting the workers whose labour sustains AI systems. The people who label data, moderate content, annotate datasets, and work under algorithmic management are central to the AI economy, yet they remain almost wholly unprotected.

Across regions, governments are more likely to govern AI as a matter of innovation, skills, and high-level principles than as a material force reshaping labour conditions – a domain where the negative effects of AI are particularly tangible.

## Bright Spot: Uruguay

In Uruguay, Law 20.396/2025 on Minimum Protections for Workers Performing Tasks through Digital Platforms establishes minimum remuneration, working time limits, occupational accident coverage, and social security access for workers who carry out their tasks through digital platforms, including those engaged in autonomous employment relationships. The law requires employers to uphold equality and non-discrimination when implementing algorithms in the workplace. It mandates the disclosure of automated monitoring and decision-making systems as well as disclosure of key parameters used in automated decisions in the workplace. Workers are also entitled to written explanations of any automated decision affecting their working conditions, including suspension and termination. The law explicitly guarantees autonomous platform workers the right to freedom of association and collective bargaining, enabling them to negotiate conditions more favourable than the legal minimums.

# 10 Global AI governance is fragmenting before a shared floor of protection has been established

Global responsible AI governance is expanding, but it is not converging around a shared baseline of protection. Countries and regions are moving at different speeds, prioritising different issues, and relying on instruments with very different levels of legal force. The result is a fragmented governance landscape in which the protections people receive from AI-related harms may depend heavily on where they live, which sector is involved, and whether commitments are binding or voluntary.

This fragmentation appears in four directions.

First, there is a vertical divide between Global North and Global South countries. Average GIRAI scores range from 55 in Global North countries to 27 in the Global South. Evidence of implementation of existing frameworks is available in 66% of cases in Global North countries, compared with 45% in the Global South. This suggests that countries are not only adopting responsible AI governance at different rates but also building implementation capacity at different depths.

Second, there is horizontal fragmentation across responsible AI priorities across regions. Despite some clear global trends, as those covered in the previous findings, regions differ significantly in what they are choosing to govern and prioritise. Only two indicators remain relatively consistent as priorities across regions: Safety and Security, and Transparency and Explainability. This suggests some emerging consensus around technical and procedural safeguards but far less agreement on the broader social, economic, and rights-based dimensions of responsible AI. It also means that governments find it easier to agree on the machinery of AI governance than on its purpose. The areas where convergence occurs are procedural, whereas responses to the substantive questions of what responsible AI is for, whose rights it protects, whose labour it values, and whose environment it spares remain fragmented.

Third, there is fragmentation in the enforceability of frameworks, with a large variance between binding and voluntary frameworks. Of 197 AI-related frameworks approved between November 2023 and September 2025, 150 are non-binding instruments such as strategies, guidelines, white papers, principles, or voluntary commitments. The majority of the growth in terms of enforceability lies in Europe because of the EU AI Act, covering 21 countries measured in this Edition of the Index. This means that responsible AI principles may be translated into enforceable obligations in some countries while remaining largely aspirational in others.

Fourth, there is fragmentation in multi-stakeholder participation in the development of frameworks. This is visible both in the incomplete diffusion of whole-of-government AI governance framework adoption and in the limited institutionalisation of civil society participation. While 73 of 135 countries have adopted a national AI policy or equivalent framework articulating a national approach to AI governance, CSOs were consulted in the drafting process in just 37 cases, while only 25 policies contain provisions for ongoing CSO participation or oversight. Across the full sample of countries, 40 countries have some form of operational mechanism through which CSOs can participate in, monitor, or advise on AI governance. These mechanisms include, for example, advisory or oversight bodies with CSO seats, AI observatories, or participation in audits or assessments of public sector AI use.

The fragmentation of governance approaches can reflect legitimate differences in national and regional priorities, as countries face different risks, capacities, development needs, institutional histories, and technological dependencies. But fragmentation becomes a problem when it prevents the emergence of a shared, rights-based floor for AI governance.

Without a minimum threshold, global interoperability risks being built around technical compatibility, market access, and cross-border deployment rather than common safeguards for people. Systems may become interoperable before protections do. Data, models, services, and standards may move across borders more easily than rights, remedies, transparency obligations, labour protections, environmental safeguards, or meaningful public participation.

This risks deepening global inequality. Global North countries and powerful technology companies would be better placed to set the terms of AI governance, while Global South ones may be left to adopt systems and standards they had limited power to shape. Interoperability would then serve those already positioned to benefit from AI, while exposing less powerful countries and communities to weaker protections and greater dependency.

Global AI governance needs enough alignment around minimum protections to ensure that responsible AI does not become a fragmented patchwork of soft commitments, uneven implementation, and unequal protection.

Figure 16: **Regional Variation in Framework Coverage Across AI Policy Indicators**

Each dot represents a region's deviation, in percentage points, from the global average framework coverage for a given AI Policy indicator: positive values indicate above-average coverage, negative values below-average. The standard deviation (σ), shown on the right, measures how much regional coverages diverges from the global average for each indicator.

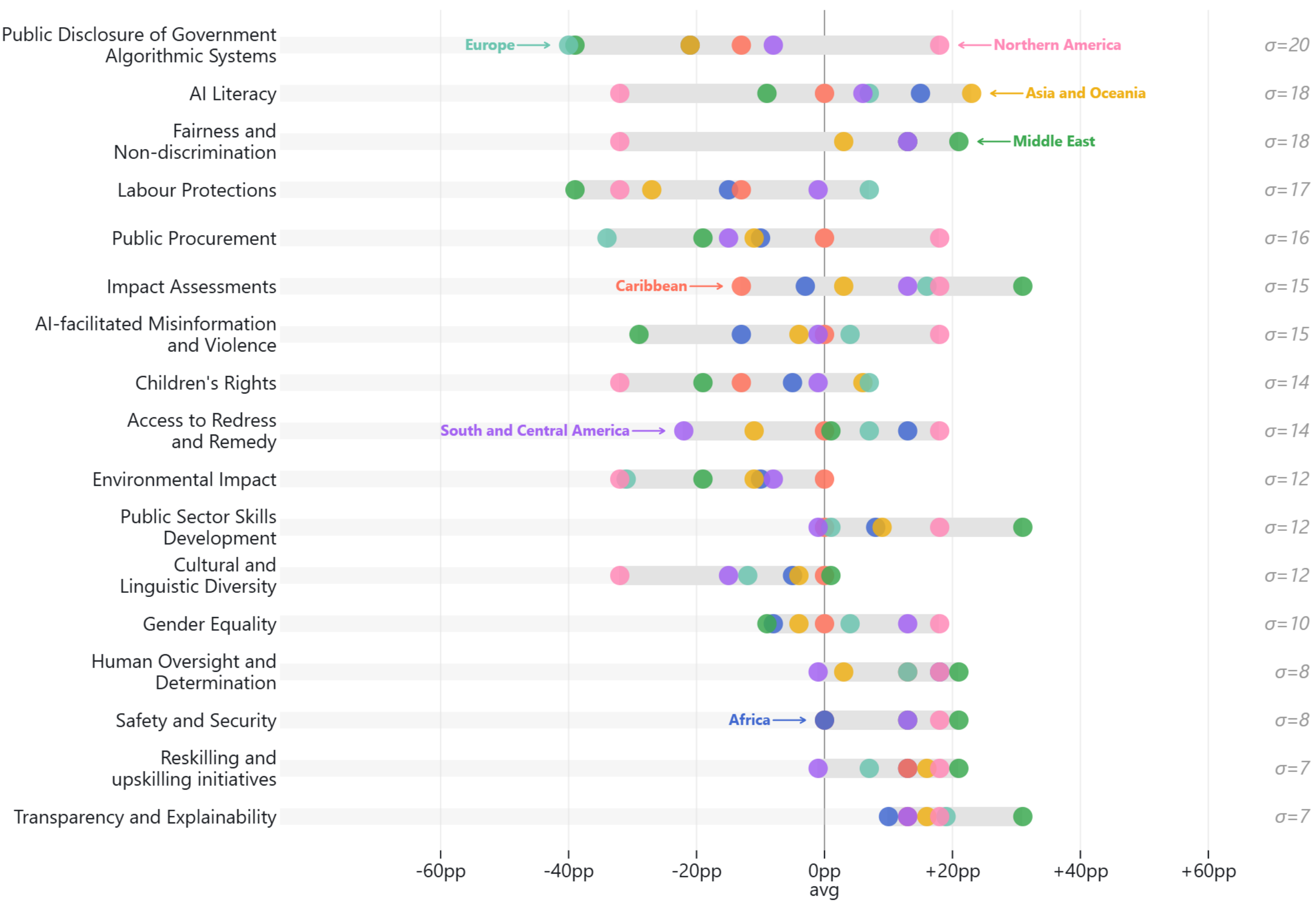

*Source: Global Index on Responsible AI (GIRAI) Global Survey, 2026.*

PART 4

# From Commitments to Protections: The Next Phase of Responsible AI Governance

The 2nd Edition shows that responsible AI governance has entered a new phase. The central question is no longer whether governments are beginning to recognise responsible AI. Across the Index, 128 of 135 governments show some form of commitment to responsible AI, and around 200 new responsible AI-related frameworks have been adopted since the 1st Edition. The question now is whether these commitments are becoming meaningful protections in people's lives.

The findings of this report show a widening gap between responsible AI as a language of commitment and responsible AI as a system of governance. Frameworks are spreading, but implementation remains uneven. Many new instruments are non-binding. Oversight bodies, monitoring systems, redress mechanisms, public disclosure requirements, environmental reporting, labour protections, and participatory processes remain non-existent or underdeveloped.

The harms that responsible AI must address are no longer abstract or hypothetical; rather, the gap between commitment and implementation is affecting lives. AI is increasingly embedded in the systems through which people all over the world access work, education, healthcare, welfare, public services, information, security, and political participation. As AI becomes a structural force in everyday life, its governance will help determine whether the societies of tomorrow are more just, free, equal, fair, and sustainable.

In this section of the report, we draw together the main implications of the 2nd Edition and set out the priorities for the next phase of responsible AI governance. We show why the task ahead is not only to adopt more frameworks but to ensure that responsible AI becomes enforceable, implemented, participatory, and capable of protecting people in practice.

## Responsible AI must become public infrastructure

The next phase of responsible AI requires a shift from framework adoption to institution-building. The 2nd Edition shows that responsible AI governance cannot remain a set of policy commitments. It must be built into the institutions and systems that determine how AI is procured, deployed, monitored, challenged, and corrected.

Governance needs to be sturdier and more durable than a strategy document will allow. It needs the practical capacity to trace where AI is used across areas, including public services, workplaces, schools, welfare systems, courts, borders, information environments, and local infrastructures; to evaluate what AI does; to intervene when harm occurs; and to ensure that people can challenge decisions that affect their lives.

Without that shift, responsible AI will remain visible in policy language but weak in the places where people need it most.

## Governments must govern AI as public power, not only as market activity

A central lesson from the Index is that governments are more willing to set expectations for the broader AI ecosystem than to subject their own use of AI to equivalent scrutiny. Public-sector AI is AI used through and with the authority of the state and therefore requires an even higher level of accountability.

If governments do not disclose their own use of AI, provide routes for accountability, and create meaningful pathways for public participation, their role as custodians of human rights and democracy is called into question. They cannot credibly govern responsible AI while leaving their own deployments opaque or weakly accountable.

The future of responsible AI governance will depend on whether governments are willing to make their own AI systems visible, contestable, and subject to independent oversight. Public registers, procurement transparency, disclosure of AI-mediated decisions, participatory governance mechanisms, and accessible redress should become baseline expectations for democratic AI governance.

## Responsible AI must govern the material conditions of AI

Too much of the responsible AI agenda treats AI as a technical system, rather than as a material, economic, social, and political one.

AI development depends on physical infrastructures: energy, water, land, minerals, data centres, hardware supply chains, and human labour. Its deployment depends on institutions, workplaces, public authorities, and markets willing to adopt it. The harms of AI therefore arise not only from automated decisions but from the material and social systems that make AI possible.

Environmental and labour protections remain at the edges of global responsible AI governance despite growing evidence that AI is already affecting workers, communities, and ecosystems. Data centre expansion is placing more and more pressure on water, energy, land, and communities. AI supply chains rely on workers who label data, moderate content, maintain systems, and increasingly work under algorithmic management.

The next phase of responsible AI governance must directly regulate these conditions. That means requiring disclosure of AI's energy and water use, environmental impact reporting, enforceable efficiency standards, and consultation with affected communities. It also means protecting the workers who build and maintain AI systems through fair pay, safe working conditions, limits on opaque algorithmic management, and rights to organise, bargain collectively, and challenge harmful workplace AI systems.

Environment and labour are not simply resources for AI development. They are basic conditions of life, livelihood, and social flourishing and must be treated as core concerns of responsible AI governance.

## Preparing people for AI cannot substitute for effectively governing AI systems

Across several findings, a wider pattern emerges: governments are often more comfortable preparing people to adapt to AI than requiring AI systems to be safe, fair, accountable, and rights-respecting.

AI literacy, reskilling, and capacity-building are important. They can help people understand and use AI systems, participate in AI-enabled economies, and develop skills for changing labour markets, but they cannot replace protection. Teaching people to navigate AI does not remove the obligation to ensure that AI systems do not exploit workers, harm children, discriminate against marginalised groups, distort public information, or deny people access to services without explanation or remedy.

The next phase of responsible AI governance must therefore resist shifting the burden of adaptation onto individuals. Responsible AI cannot mean asking children, workers, women, minorities, or low-income communities to become more resilient to systems that remain weakly governed. It must mean changing the conditions under which AI is designed, deployed, monitored, and contested through clear protections, particularly for those most vulnerable to AI harms.

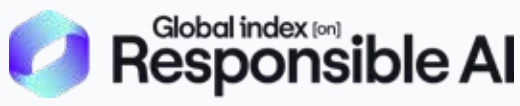

## Inclusion must become enforceable

The findings show that inclusion is increasingly part of the language of responsible AI. Gender, children's rights, cultural and linguistic diversity, and marginalised communities appear more often in policy frameworks than before. This is important, but recognition is not the same as protection.

The next task is to make inclusion enforceable. That means moving from naming groups in policy documents to building obligations, funding, oversight, reporting systems, participatory processes, and redress mechanisms that respond to the harms those groups actually face.

When marginalised groups are named in policy but not effectively protected, their harms risk becoming further invisibilised. This happens because policy recognition can create the appearance that a problem has been addressed while failing to recognise the concrete ways in which harms are experienced: through exclusion from services, discriminatory profiling, online abuse, deepfake sexual content, linguistic misrepresentation, unsafe systems, or the absence of accessible routes to challenge harm.

This is particularly important because AI harms are not distributed evenly. Some people and communities, including women and gender-diverse people, racialised minorities, migrants, low-income workers, and many in the Global South are more exposed to surveillance, misinformation, online abuse, labour exploitation, environmental burdens, or poor system performance. Responsible AI must therefore be judged by whether it protects those most likely to bear these harms, not only by whether it adopts inclusive language.

## Global AI governance needs a rights-based floor

The Index points to a fragmented global governance landscape. Some fragmentation is legitimate; countries and regions face different priorities, capacities, histories, risks, and development needs. Responsible AI should not mean a single model imposed everywhere.

But tackling fragmentation cannot simply default to familiar calls for interoperability or shared global frameworks. If these processes are led mainly by Global North countries and powerful technology companies, they risk reproducing the same inequalities that already shape the global AI ecosystem. Common frameworks may then reflect the priorities of those with the greatest regulatory capacity and market power, while leaving Global South countries to adopt standards they had limited role in shaping and may lack the resources to enforce.

Nor can AI governance remain entirely fragmented across national boundaries, voluntary, or industry-led. The risks and harms of AI systems are too serious, too cross-border, and too unevenly distributed for self-regulation to be sufficient. It is also unrealistic to assume that the people and communities most exposed to AI harms will always have the power, information, or resources to demand accountability on their own.

The task, then, is to build global alignment around minimum rights-based protections while preserving space for different regions to address their own priorities. That baseline should include expectations on transparency, oversight, redress, non-discrimination, labour protection, environmental accountability, public-sector disclosure, and meaningful participation.

Without such a foundation, interoperability may serve markets before it protects people. Data, models, platforms, standards, and services may move across borders more easily than rights, remedies, safeguards, or public accountability. A rights-based foundation is therefore essential to ensure that global AI governance matures into a basis for more just and accountable AI futures rather than remaining a patchwork of soft commitments and unequal protection.

### The role of civil society and public-interest actors must be strengthened

Where implementation is weak, civil society often becomes essential to making harms visible, representing affected communities, challenging misuse, and shaping the meaning of responsible AI from the ground up. Civil society cannot, however, substitute for state accountability, and it cannot play this role without resources, access to information, and institutional channels for influence.

The next phase of responsible AI governance should therefore invest in public-interest capacity: CSOs, universities, independent researchers, labour organisations, digital rights groups, journalists, public institutions, and affected communities. These actors are needed not only to monitor government and industry but to help define what responsible AI should mean in different social, cultural, linguistic, economic, and political contexts.

This is particularly important in Global South countries, where governments may face resource constraints, dependence on external technology providers, and pressure to adopt AI quickly for development goals and market growth. South–South coordination can help shift the balance of influence by supporting shared positions on public procurement, data governance, labour protections, local-language AI, infrastructure dependency, environmental safeguards, and accountability demands for global AI companies.

## Priorities for the Next Phase of Responsible AI

The findings of the 2nd Edition suggest that responsible AI governance must now move in several directions at once.

### 1. Hold public-sector AI to a higher standard

Governments should disclose when and how AI is used in public services and rights-bearing domains. Public algorithmic registers, procurement transparency, disclosure of AI-mediated decisions, and accessible routes for challenge should become baseline requirements.

## 2. Build institutions as well as frameworks

Only 29 countries have established independent oversight bodies. Responsible AI requires independent and adequately resourced oversight bodies, monitoring and evaluation systems, audit capacity, procurement safeguards, enforcement mechanisms, and public reporting. Frameworks without institutions will not protect people, just as institutions without legally binding frameworks to ensure sustained protection of rights will be ineffective.

## 3. Establish binding floors in high-risk areas

Voluntary commitments have limits. Binding obligations are needed where harms are foreseeable and serious, including public-sector AI, children's rights, gendered harms, AI-facilitated misinformation and violence, environmental impact, labour protections, cultural and linguistic diversity, transparency, and redress.

## 4. Make redress central to responsible AI

Responsible AI should be judged by whether people can challenge AI-related decisions and obtain remedy when harm occurs. Redress should not be an afterthought. It is the point at which responsibility becomes meaningful to those affected.

## 5. Bring labour and environmental harms into AI governance

Governments should treat labour conditions, data work, platform work, algorithmic management, energy use, water demand, e-waste, data centre infrastructure, and community impact as core responsible AI issues.

## 6. Fund independent public-interest capacity and knowledge production

AI governance capacity should not depend primarily on the companies whose systems are being governed. Universities, regulators, public institutions, CSOs, labour groups, digital rights organisations, and independent researchers should be resourced to support accountable AI governance.

### 7. Create the conditions for making participation meaningful

Affected communities should have influence over AI governance rather than only occasional consultation. Workers, children's rights advocates, women's rights organisations, disability groups, linguistic minorities, communities affected by AI infrastructure, public service users, and digital rights groups need formal pathways to shape decisions.

### 8. Build global convergence around rights, not only systems

Global AI governance needs a shared rights-based foundation to ensure that data, models, standards, and services do not move across borders faster than protections.

### 9. Treat monitoring and evidence as governance infrastructure

Only 25 countries have monitoring and evaluation reports or platforms. Public registers, incident reporting systems, environmental reporting, implementation reviews, access-to-information rules, and independent observatories are not secondary research tools. They are core governance infrastructure. Without evidence, harms remain hidden; without disclosure, they remain difficult to contest.

### 10. Advance South-South cooperation on AI governance

Global South countries should approach international AI governance fora with the position that interoperability is something to negotiate, not import. South-South coordination on AI governance, including on shared positions for global negotiations, is currently underdeveloped and should be prioritised and supported. Such coordination would materially shift the imbalance of influence.

# Conclusion

The 2nd Edition shows that responsible AI governance is growing, but growth alone is not enough. The next phase must be judged by whether responsible AI can protect people in the places where AI is already shaping rights, opportunities, dignity, environments, and everyday life.

# Annexures

# Methodology

The 2nd Edition of the GIRAI assesses the extent to which countries are building the frameworks, institutions, practices, and enabling conditions needed to govern AI responsibly.

The Index is grounded in a human rights-based understanding of responsible AI. It draws on the values, principles, and policy areas set out in the UNESCO Recommendation on the Ethics of Artificial Intelligence as well as wider international human rights, democratic governance, and sustainable development commitments. Its purpose is to track how far countries are moving from high-level commitments on responsible AI to concrete governance action.

Data for the Index was collected by a global research network of country-level experts. Researchers assessed publicly available evidence on government frameworks, implementation activity, and civil society engagement. Researchers further documented cases of government misuse of AI. This primary data was complemented by selected secondary data sources measuring broader enabling conditions for responsible AI, such as rule of law, equality, cybersecurity, labour rights compliance, digital skills, and access to information.

The 2nd Edition builds on the methodology developed for the 1st Edition, while introducing a more detailed assessment of the gap between framework existence and implementation. This distinction is central to the updated methodology. The Index does not ask only whether a country has adopted a law, policy, strategy, guideline, or other AI governance framework. It also assesses whether there is evidence that these commitments are being operationalised through institutions, oversight mechanisms, programmes, monitoring systems, public consultations, budgets, or other forms of implementation.

**The Index is structured around five dimensions of responsible AI governance:**

1. Inclusion and Diversity
2. Ethics and Sustainability
3. Labour and Skills
4. Trust and Safety
5. AI Use in Public Service

Each dimension contains a set of indicators that capture key components of responsible AI. These include, among others, Gender Equality, Children's Rights, Fairness and Non-Discrimination, Transparency and Explainability, Human Oversight and Determination, Environmental Impact, Labour Protections, AI Literacy, Safety and Security, Access to Redress and Remedy, Public Sector Skills Development, and Public Disclosure of Government Algorithmic Systems.

For government-led indicators, the Index assesses three main types of evidence.

First, it assesses the **existence and quality of government frameworks**. Frameworks include laws, regulations, policies, strategies, guidelines, and draft instruments adopted or developed by national or federal governments. The Index evaluates not only whether such frameworks exist but also their status, enforceability, reach, scope, exclusions, and thematic coverage.

Second, it assesses the **implementation of government frameworks**. This captures whether adopted frameworks are being translated into concrete action. Evidence may include implementation bodies, operational programmes, dedicated budget lines, regulatory sandboxes, algorithmic audits, impact assessments, AI registers, monitoring and evaluation systems, public tenders, training programmes, research centres, standards, certification schemes, public awareness campaigns, advisory groups, AI observatories, or independent oversight bodies.

Third, where no formal framework exists, the Index captures **government-led initiatives** that show movement towards responsible AI governance. This allows the Index to recognise early-stage action in countries that may not have adopted formal AI governance instruments but are undertaking relevant work through consultations, working groups, public programmes, research, training, or awareness-raising activities.

The Index also assesses **civil society engagement** across the responsible AI ecosystem. CSOs play a critical role in promoting accountability, representing affected communities, advancing public interest concerns, and scrutinising the use of AI by both state and non-state actors. The Index captures CSO activity through evidence of policy engagement, litigation, capacity building, public awareness, research, monitoring, and participation in governance processes.

A further indicator captures **documented cases of government misuse of AI**. This includes credible evidence of AI systems deployed by governments in ways that pose unacceptable risks to human rights, rule of law, or democratic governance, such as mass biometric surveillance, AI-based predictive policing, social scoring, or other high-risk public-sector uses. Country-level researchers were asked to provide robust evidence of government use of AI systems within the period under study — 1 November 2023 to 30 September 2025 — across the following categories: AI-driven cyberattacks, disinformation, or influence operations; criminal justice and law enforcement misuse; discriminatory public service systems; mass biometric surveillance (e.g., facial recognition in public spaces); social scoring systems; and AI weapons or the use of AI in armed conflicts. Researchers could record up to one instance per category. In practice, the highest number of categories documented for any single country was four.

The 2nd Edition also revises how contextual conditions are incorporated into the Index. In the 1st Edition, contextual indicators were used as coefficients to adjust pillar scores. In the 2nd Edition, contextual indicators are embedded within the relevant thematic dimensions. This change improves transparency and interpretability while ensuring that responsible AI governance is assessed in relation to the broader conditions that shape whether rights-respecting AI governance is possible in practice.

Countries score points across the five dimensions, each with equal weight. Within each dimension, scores come from indicators from the three pillars: AI Policy (17 indicators that include government frameworks and government-led initiatives); CSO Engagement (five indicators), and Enabling Conditions (15 indicators). In addition, where Unacceptable Risks AI Systems (URAI Systems) indicator is assessed separately and applied as an accountability penalty to the final country score.

All data underwent a structured quality assurance process to assess the consistency, relevance, and evidentiary strength of the information submitted. The Index relies on publicly verifiable evidence wherever possible and is designed to support comparative analysis across countries and regions.

The Index should be understood as an evidence-based assessment of responsible AI governance ecosystems. It does not measure all AI activity within a country, nor does it claim to capture every informal or undocumented initiative. Instead, it provides a systematic, comparable, and globally representative picture of how countries are advancing responsible AI through frameworks, implementation, civil society engagement, and enabling governance conditions.

# How GIRAI Scores Are Computed

GIRAI scores are derived from 38 indicators: 37 are organised across five equally weighted dimensions, each combining three pillars: AI Policy, CSO Engagement, and Enabling Conditions. The 38th - the Unacceptable Risk AI Systems (URAI) indicator- is applied separately as a penalty to the final country score.

Scores are computed in several steps. First, qualitative evidence submitted by researchers is coded into variables by assigning numerical values to the different types of evidence. Higher values reflect stronger responsible AI governance outcomes. For example, binding frameworks receive higher scores than non-binding instruments, broader thematic coverage receives higher scores than partial.

For AI Policy indicators, frameworks are coded by variables such as type, scope, stakeholder consultation, operationalisation, thematic coverage, reach, and defence or security exemptions. Government-led initiatives are coded by number and linked to the relevant indicator.

CSO Engagement evidence is coded according to the number of CSO-led initiatives and the policy areas to which they contribute, while in the fifth dimension the pillar captures government mechanisms for CSO inclusion in AI policy and governance.

Enabling Conditions indicators are based on secondary sources and treated separately. They are reviewed for missing values, outliers, and directionality, so that higher values consistently indicate better performance. All indicators are then normalised to a common 0–100 scale.

Dimension/pillar scores are first computed as the average of the relevant indicator scores.

Each dimension score is then calculated using the pillar weights: 60% for AI Policy, 10% for CSO Engagement, and 30% for Enabling Conditions. The raw GIRAI score is the average of the five dimension scores.

Finally, the Unacceptable Risks AI Systems indicator is applied as a penalty to the raw GIRAI score. The penalty is based on the number of documented categories of unacceptable-risk AI systems deployed by national governments. A country with no documented cases receives no penalty; one case applies a 0.96 multiplier; two cases apply 0.93; three cases apply 0.91; and four or more cases apply 0.90. The final score is therefore computed as:

GIRAI = GIRAI_raw × urai_penalty

Further details on variables, coding rules, normalisation, imputation, and aggregation are available in the full methodology document[73].

# How the 2nd Edition of GIRAI Evolved

The 2nd Edition of the GIRAI builds on the 1st Edition but introduces a stronger distinction between framework existence and implementation.

The Index does not only assess whether countries have adopted responsible AI laws, policies, strategies, or guidelines. It also examines whether those commitments are being put into practice through institutions, oversight mechanisms, programmes, standards, budgets, monitoring systems, public consultations, or other concrete actions. Another key update to the Index approach and structure is the reshaping of its dimensions and pillars. Thematic indicators were regrouped to better reflect current debates on responsible AI. The three initial dimensions — Responsible AI Capacities, Human Rights and AI, and Responsible AI Governance — were replaced by a more thematic structure: Inclusion and Diversity, Ethics and Sustainability, Labour and Skills, Trust and Safety, and AI Use in Public Service. This change allows more straightforward interpretation, enabling the evidence to support concrete policy action to protect human rights across these five priority areas.

In this new matrix structure, pillars are no longer calculated within each thematic area. Instead, each indicator clearly belongs to a specific pillar, and aggregation begins at the dimension level. More concretely, the first two pillars of the 1st Edition were merged into one: AI Policy indicators now cover both government frameworks and government-led initiatives, with the latter assessed as evidence of implementation where a framework exists and evidence of government activity where no framework has been put in place. All dimensions include AI Policy indicators, each focused on a distinct key topic of responsible AI.

The third pillar of the 1st Edition — Non-State Actors — shifted from the indicator level to the dimension level. Each dimension now includes a dedicated indicator to track civil society engagement on the relevant topic. This change, along with a narrowing of scope to CSOs only — no longer tracking academia or the private

73. https://global-index.ai/methodology

sector — was decided primarily to reduce the data collection burden on country-level researchers, while still capturing key actors of the responsible AI ecosystems.

The 2nd Edition also changes how contextual conditions are incorporated. Unlike the 1st Edition, in which contextual indicators were used as coefficients to adjust scores, the 2nd Edition embeds contextual indicators within the relevant thematic dimensions. This improves interpretability while recognising that responsible AI governance depends not only on formal AI policies but also on the broader institutional and rights environments in which they operate. This pillar is built entirely from secondary data sources — that is, existing measurements developed by other organisations whose scores are incorporated into the GIRAI scoring system.

At the indicator level, the 2nd Edition also introduces greater detail in the assessment of government frameworks. Whereas the 1st Edition tracked only the existence and enforceability of frameworks, the 2nd Edition captures and scores multiple additional variables: the thematic breadth of frameworks in addressing specific issues, the inclusion of budget allocations, and references to external consultations, among others. At the same time, more information on government-led activities is captured for each indicator, strengthening the assessment at the implementation level.

# Key Data and Figures

Figure A1: **Country-Level GIRAI 2026 Scores: Overall, Dimension, and Pillar**

Overall, Dimension, and Pillar scores for all 135 countries and jurisdictions assessed in GIRAI 2026, with regional and World Bank income group classifications.

| Colored by WB Income Group | Colored by group |
|---|---|
| High | Global North |
| Upper middle | Global North |
| Lower middle | Global South |
| Low | Global South |

| # | Country | Region | GIRAI Score | DIMENSION SCORES | | | | | PILLARS SCORES | | |
|---|---|---|---|---|---|---|---|---|---|---|---|
| | | | | Inclusion & Diversity | Ethics & Sustain-ability | Labour & Skills | Trust & Safety | AI in Public Service | AI Policy | CSO Engage-ment | Enabling Condi-tions |
| 1 | Norway | Europe | 75.26 | 76.1 | 75.08 | 75.15 | 75.75 | 74.2 | 71.2 | 42.94 | 94.14 |
| 2 | Italy | Europe | 72.71 | 68.37 | 73.59 | 71.71 | 79.49 | 70.4 | 73.9 | 51.74 | 77.33 |
| 3 | Ireland | Europe | 71.39 | 70.83 | 67.77 | 76.45 | 76.6 | 65.28 | 66.49 | 45.64 | 89.76 |
| 4 | France | Europe | 70.32 | 74.74 | 75.9 | 74.76 | 74.55 | 78.12 | 72.82 | 57.65 | 87.19 |
| 5 | Netherlands | Europe | 69.51 | 81.6 | 80.51 | 72.25 | 83.67 | 68.14 | 76.5 | 57.89 | 85.15 |
| 6 | Germany | Europe | 68.98 | 82.98 | 72.03 | 77.56 | 76.86 | 61.43 | 72.65 | 46.09 | 86.58 |
| 7 | United Kingdom of Great Britain and Northern Ireland | Europe | 67.27 | 77.77 | 74.77 | 78.26 | 66.01 | 64.86 | 70.7 | 54.44 | 81.56 |
| 8 | Slovenia | Europe | 66.13 | 67.79 | 65.57 | 67.93 | 73.48 | 55.88 | 63.62 | 36.1 | 81.15 |
| 9 | Latvia | Europe | 65.18 | 72.61 | 56.93 | 73.12 | 74.26 | 49 | 64.43 | 12.52 | 84.24 |
| 10 | Estonia | Europe | 64.89 | 76.51 | 56.47 | 58.73 | 72.3 | 60.44 | 62.35 | 12.53 | 87.42 |
| 11 | Brazil | South & Central America | 63.3 | 72.01 | 63.42 | 76.27 | 61.49 | 67.14 | 68.56 | 52.21 | 72.36 |
| 12 | Spain | Europe | 63.27 | 74.15 | 67.99 | 75.41 | 73.8 | 38.2 | 62.98 | 30.39 | 83.6 |
| 13 | Greece | Europe | 62.36 | 68.89 | 61.84 | 66.17 | 71.39 | 66.97 | 67.46 | 47.89 | 72.63 |

| # | Country | Region | GIRAI Score | Inclusion & Diversity | Ethics & Sustain-ability | Labour & Skills | Trust & Safety | AI in Public Service | AI Policy | CSO Engage-ment | Enabling Condi-tions |
|---|---|---|---|---|---|---|---|---|---|---|---|
| 14 | Chile | South & Central America | 61.91 | 71.92 | 56.56 | 74.03 | 59.65 | 60.29 | 64.91 | 23.48 | 77.32 |
| 15 | Bulgaria | Europe | 61.61 | 66.01 | 59.09 | 66.98 | 77.71 | 38.28 | 59.56 | 40.79 | 72.66 |
| 16 | Belgium | Europe | 60.49 | 55.54 | 69.65 | 55.93 | 73.74 | 47.6 | 52.64 | 22.72 | 88.79 |
| 17 | Poland | Europe | 60.36 | 67.31 | 46.77 | 63.93 | 73.5 | 50.28 | 57.99 | 31.21 | 74.8 |
| 18 | Portugal | Europe | 60 | 72.24 | 61.87 | 66.76 | 73.78 | 37.86 | 57.44 | 31.44 | 82.98 |
| 19 | Switzerland | Europe | 59.63 | 35.93 | 76.4 | 47.48 | 71.48 | 66.83 | 46.91 | 38.27 | 92.18 |
| 20 | Austria | Europe | 58.16 | 57.59 | 56.1 | 71.15 | 70.08 | 35.88 | 48.65 | 20.14 | 89.84 |
| 21 | Lithuania | Europe | 57.69 | 67.83 | 56.08 | 60.91 | 75.36 | 28.28 | 51.06 | 12.96 | 85.86 |
| 22 | Uruguay | South & Central America | 56.99 | 50.6 | 61.66 | 49.95 | 64.32 | 70.29 | 53.58 | 20.43 | 83.9 |
| 23 | Australia | Asia & Oceania | 56.4 | 64.2 | 48.69 | 59.83 | 66.56 | 54.45 | 48.81 | 47.5 | 82.36 |
| 24 | Japan | Asia & Oceania | 55.66 | 61.95 | 55.84 | 51.93 | 63.5 | 56.66 | 48.66 | 47.62 | 80.06 |
| 25 | Canada | Northern America | 55.09 | 58.92 | 57.97 | 46.13 | 53.1 | 70.81 | 46.33 | 35.9 | 86.66 |
| 26 | Colombia | South & Central America | 54.74 | 53.84 | 46.33 | 64.04 | 54.9 | 54.6 | 55.11 | 35.16 | 60.54 |
| 27 | United States of America | Northern America | 53.34 | 56.26 | 45.78 | 63.4 | 59.21 | 62.11 | 51.08 | 48.05 | 72.99 |
| 28 | Costa Rica | South & Central America | 52.3 | 62.82 | 54.92 | 51.96 | 39 | 52.77 | 47.92 | 10.09 | 75.11 |
| 29 | Hungary | Europe | 52.07 | 62.23 | 49.62 | 49.52 | 69.1 | 29.9 | 51.06 | 3.96 | 70.14 |
| 30 | Romania | Europe | 51.69 | 48.27 | 50.07 | 58.12 | 68.03 | 33.96 | 48.6 | 13.84 | 70.49 |
| 31 | Kyrgyz Republic | Asia & Oceania | 51.64 | 63.24 | 66.61 | 41.97 | 61.53 | 35.62 | 54.75 | 34.15 | 58.43 |
| 32 | Slovakia | Europe | 50.25 | 51.65 | 52.78 | 49.52 | 69.96 | 27.37 | 42.1 | 17 | 77.65 |
| 33 | Serbia | Europe | 50.17 | 54.38 | 47.14 | 57.76 | 41.6 | 50 | 50.28 | 11.35 | 62.91 |
| 34 | Peru | South & Central America | 50.06 | 44.2 | 57.25 | 55.66 | 51.99 | 41.19 | 48.48 | 22.03 | 62.56 |

| # | Country | Region | GIRAI Score | Inclusion & Diversity | Ethics & Sustain-ability | Labour & Skills | Trust & Safety | AI in Public Service | AI Policy | CSO Engage-ment | Enabling Condi-tions |
|---|---|---|---|---|---|---|---|---|---|---|---|
| 35 | Croatia | Europe | 48.12 | 52.22 | 51.62 | 45.14 | 69.42 | 22.19 | 41.69 | 5.71 | 75.1 |
| 36 | Dominican Republic | Caribbean | 47.92 | 51.44 | 40.96 | 48.05 | 47.15 | 51.97 | 48.56 | 14.21 | 57.85 |
| 37 | South Korea | Asia & Oceania | 47.69 | 35.83 | 51.86 | 54.49 | 63.52 | 42.69 | 42.45 | 12.67 | 76.47 |
| 38 | Nigeria | Africa | 45.93 | 52.06 | 49.63 | 40.85 | 63.45 | 23.65 | 51.18 | 26.94 | 41.76 |
| 39 | New Zealand | Asia & Oceania | 45.81 | 40.78 | 53.2 | 50.12 | 38.82 | 46.16 | 29.94 | 25.05 | 84.48 |
| 40 | Singapore | Asia & Oceania | 45.21 | 59.06 | 43.53 | 39.23 | 55.18 | 29.05 | 36.76 | 3.3 | 76.08 |
| 41 | Israel | Middle East | 43.75 | 52.17 | 47.96 | 45.02 | 44.07 | 51.14 | 39.54 | 63.84 | 59.89 |
| 42 | China | Asia & Oceania | 43.58 | 43.97 | 53.43 | 49.23 | 64.16 | 31.29 | 50.9 | 33.27 | 48.51 |
| 43 | China Hong Kong Special Administrative Region | Asia & Oceania | 42.74 | 47.67 | 29.28 | 41.66 | 54.75 | 40.34 | 35 | 14.89 | 67.51 |
| 44 | Ukraine | Europe | 42.17 | 38.52 | 42.7 | 53.09 | 58.55 | 33.85 | 33.05 | 66.75 | 62.79 |
| 45 | Republic of Moldova | Europe | 42.04 | 34.9 | 36.83 | 55.73 | 43.55 | 39.19 | 35.17 | 27.98 | 60.46 |
| 46 | India | Asia & Oceania | 41.77 | 54.18 | 39.6 | 40.55 | 56.94 | 33.3 | 40 | 44.98 | 54.73 |
| 47 | Viet Nam | Asia & Oceania | 41.56 | 33.98 | 30.96 | 52.33 | 41.78 | 48.76 | 34.58 | 38.11 | 56.68 |
| 48 | Egypt | Africa | 41.26 | 37.21 | 39.54 | 50.28 | 47.99 | 31.29 | 39.97 | 40.42 | 44.14 |
| 49 | Jordan | Middle East | 40.46 | 27.1 | 30.82 | 61.91 | 39.03 | 43.44 | 35.96 | 24.17 | 54.88 |
| 50 | Kenya | Africa | 39.53 | 22.88 | 52.91 | 41.89 | 40.28 | 47.94 | 38.24 | 23.2 | 53.05 |
| 51 | Pakistan | Asia & Oceania | 38.93 | 22.61 | 32.15 | 38.97 | 61.15 | 47.86 | 42.7 | 20.27 | 43.01 |
| 52 | Qatar | Middle East | 38.88 | 35.31 | 40.62 | 40.09 | 50.63 | 27.76 | 36.27 | 4.62 | 55.52 |
| 53 | Indonesia | Asia & Oceania | 38.83 | 39.62 | 34.74 | 50.72 | 43.77 | 25.3 | 31.45 | 24.12 | 58.5 |
| 54 | Malaysia | Asia & Oceania | 38.54 | 33.95 | 31.16 | 50.49 | 42.67 | 42.45 | 32.7 | 16.03 | 63.07 |
| 55 | United Arab Emirates | Middle East | 38.54 | 62.14 | 32.48 | 36.15 | 42.82 | 27.12 | 35.89 | 1.76 | 61.44 |
| 56 | Ghana | Africa | 38.43 | 27.37 | 47.87 | 41.19 | 33.86 | 49.86 | 36.28 | 13.98 | 56.2 |

| # | Country | Region | GIRAI Score | Inclusion & Diversity | Ethics & Sustain-ability | Labour & Skills | Trust & Safety | AI in Public Service | AI Policy | CSO Engage-ment | Enabling Condi-tions |
|---|---|---|---|---|---|---|---|---|---|---|---|
| 57 | Nepal | Asia & Oceania | 37.69 | 40.85 | 51.71 | 37.02 | 36.81 | 29.93 | 34.62 | 22.42 | 54.17 |
| 58 | North Macedonia | Europe | 37.18 | 27.94 | 28.7 | 58.87 | 37.88 | 32.48 | 20.85 | 52.89 | 64.58 |
| 59 | Benin | Africa | 37.01 | 21.77 | 44.27 | 46.44 | 35.91 | 36.64 | 35.79 | 17.78 | 45.85 |
| 60 | Thailand | Asia & Oceania | 36.99 | 21.85 | 33.36 | 49.67 | 36.99 | 43.06 | 29.57 | 19.55 | 57.62 |
| 61 | El Salvador | South & Central America | 36.66 | 29.17 | 59.84 | 30.53 | 36.2 | 27.58 | 33.51 | 4.62 | 53.66 |
| 62 | Oman | Middle East | 36.21 | 44.87 | 38.61 | 16.4 | 43.55 | 37.59 | 34.68 | 6.5 | 49.16 |
| 63 | Morocco | Africa | 35.62 | 27.16 | 29.43 | 46.21 | 52.23 | 23.05 | 29.39 | 22.56 | 52.41 |
| 64 | Cote d'Ivoire | Africa | 35.18 | 39.33 | 40.44 | 31.37 | 31.37 | 33.38 | 31.37 | 9 | 51.52 |
| 65 | Argentina | South & Central America | 34.31 | 29.76 | 33.31 | 31.53 | 41.2 | 35.74 | 20.41 | 26.72 | 64.63 |
| 66 | Rwanda | Africa | 33.56 | 12.67 | 33.71 | 44.88 | 30.76 | 45.76 | 28.47 | 6 | 52.92 |
| 67 | Saudi Arabia | Middle East | 33.36 | 22.04 | 39.25 | 41.47 | 48.77 | 22.24 | 29.82 | 1.54 | 55.68 |
| 68 | Philippines | Asia & Oceania | 32.98 | 29.41 | 33.74 | 30.69 | 37.39 | 33.65 | 25.46 | 13.76 | 54.41 |
| 69 | Georgia | Europe | 32.43 | 18.32 | 49.91 | 20.56 | 54.81 | 18.55 | 20.33 | 5.64 | 65.55 |
| 70 | Kazakhstan | Asia & Oceania | 32.34 | 48.63 | 17.22 | 44.91 | 29.32 | 21.62 | 23.79 | 12.89 | 55.92 |
| 71 | Albania | Europe | 32.29 | 36.93 | 42.6 | 29.74 | 25.94 | 26.22 | 17.8 | 10.76 | 68.43 |
| 72 | Sri Lanka | Asia & Oceania | 32.08 | 25.48 | 32.85 | 33.42 | 40.6 | 28.04 | 22.09 | 23.29 | 54.99 |
| 73 | Ethiopia | Africa | 31.33 | 17.92 | 47.75 | 43.08 | 32.81 | 15.1 | 33.56 | 0.88 | 37.02 |
| 74 | Senegal | Africa | 31.22 | 15.85 | 19.74 | 43.28 | 31.06 | 46.2 | 26.17 | 6 | 49.73 |
| 75 | Bahrain | Middle East | 30.49 | 17.95 | 39.88 | 32.71 | 39.08 | 22.84 | 25.12 | 0 | 51.4 |
| 76 | Azerbaijan | Asia & Oceania | 29.95 | 31.36 | 17.06 | 45.46 | 23.77 | 32.11 | 22.12 | 13.64 | 51.05 |
| 77 | South Africa | Africa | 29.68 | 30.44 | 19.99 | 31.88 | 40.13 | 25.94 | 10.81 | 43.73 | 62.72 |
| 78 | Libya | Africa | 29.24 | 35.36 | 25.47 | 42.96 | 20.29 | 22.14 | 31.99 | 2.93 | 32.53 |

| # | Country | Region | GIRAI Score | Inclusion & Diversity | Ethics & Sustain-ability | Labour & Skills | Trust & Safety | AI in Public Service | AI Policy | CSO Engage-ment | Enabling Condi-tions |
|---|---|---|---|---|---|---|---|---|---|---|---|
| 79 | Zambia | Africa | 29.12 | 11.46 | 39.63 | 40.25 | 32.45 | 21.79 | 19.91 | 17.32 | 51.46 |
| 80 | Gambia | Africa | 28.19 | 27.07 | 24.61 | 37.75 | 29.64 | 21.87 | 27.42 | 4.1 | 37.75 |
| 81 | Mexico | South & Central America | 27.95 | 41.88 | 17.49 | 33.1 | 32.22 | 20.88 | 15.57 | 29.52 | 56.07 |
| 82 | Montenegro | Europe | 27.65 | 26.3 | 27.02 | 34.86 | 28.65 | 27.18 | 9.22 | 27.93 | 68.26 |
| 83 | Uzbekistan | Asia & Oceania | 26.82 | 26.99 | 17.11 | 37.14 | 25.86 | 26.99 | 20.15 | 6.15 | 47.04 |
| 84 | Mauritius | Africa | 26.29 | 19.87 | 27.95 | 35 | 26.59 | 22.06 | 10.45 | 0 | 66.74 |
| 85 | Paraguay | South & Central America | 26.14 | 22.08 | 35.36 | 20.2 | 31.48 | 21.59 | 8.31 | 18.39 | 64.39 |
| 86 | Trinidad and Tobago | Caribbean | 26.14 | 25.57 | 20.33 | 34.14 | 25.35 | 25.32 | 13.05 | 13.49 | 56.54 |
| 87 | Panama | South & Central America | 25.56 | 25.18 | 25.45 | 25.39 | 27.84 | 23.92 | 7.63 | 13.72 | 65.34 |
| 88 | Jamaica | Caribbean | 24.25 | 22.14 | 31.15 | 26.66 | 19.57 | 21.72 | 10.53 | 7.69 | 57.21 |
| 89 | Cambodia | Asia & Oceania | 24.09 | 27.67 | 16.3 | 28.23 | 21.95 | 26.31 | 15.51 | 28.57 | 39.76 |
| 90 | Kuwait | Middle East | 24.03 | 20.71 | 20.67 | 28.11 | 38.62 | 17.02 | 11.08 | 18.82 | 54.99 |
| 91 | Bhutan | Asia & Oceania | 23.35 | 31.96 | 25.82 | 12.1 | 27.02 | 19.86 | 7.93 | 0 | 61.99 |
| 92 | Botswana | Africa | 23.05 | 18.7 | 28.34 | 19.83 | 32.39 | 16.01 | 10.23 | 4.1 | 55.02 |
| 93 | United Republic of Tanzania | Africa | 22.96 | 28.9 | 19.9 | 18.11 | 29.29 | 18.6 | 11.47 | 12.16 | 49.53 |
| 94 | Lesotho | Africa | 22.8 | 28.24 | 30.02 | 15.8 | 22.33 | 17.6 | 9.94 | 1.75 | 55.52 |
| 95 | Armenia | Asia & Oceania | 22.51 | 23.43 | 18.73 | 20.68 | 31.07 | 18.63 | 5.74 | 5.13 | 61.82 |
| 96 | Ecuador | South & Central America | 21.75 | 18.72 | 22.16 | 19.98 | 23.25 | 24.63 | 6.25 | 11.85 | 56.04 |
| 97 | Mongolia | Asia & Oceania | 20.68 | 19.17 | 14.38 | 22.65 | 22.74 | 24.45 | 6.82 | 2.05 | 54.6 |
| 98 | State of Palestine | Middle East | 20.52 | 28.71 | 10.3 | 27.09 | 11.26 | 25.22 | 12.12 | 0.5 | 43.98 |
| 99 | Barbados | Caribbean | 20.18 | 24.24 | 11.94 | 27.03 | 15.4 | 22.27 | 2.27 | 9.54 | 59.53 |

| # | Country | Region | GIRAI Score | Inclusion & Diversity | Ethics & Sustain-ability | Labour & Skills | Trust & Safety | AI in Public Service | AI Policy | CSO Engage-ment | Enabling Condi-tions |
|---|---|---|---|---|---|---|---|---|---|---|---|
| 100 | Malawi | Africa | 19.71 | 6.99 | 35.76 | 14.72 | 28.64 | 12.43 | 9.43 | 4.1 | 45.46 |
| 101 | Bangladesh | Asia & Oceania | 19.58 | 19.21 | 11.71 | 17.38 | 32.51 | 21.18 | 12.9 | 5.85 | 40.24 |
| 102 | Angola | Africa | 19.41 | 9.13 | 19.43 | 16.41 | 24.96 | 27.14 | 8.56 | 9.59 | 44.39 |
| 103 | Kosovo | Europe | 18.59 | 22.81 | 9.15 | 16.16 | 25.3 | 19.53 | 0 | 3.08 | 60.94 |
| 104 | Tajikistan | Asia & Oceania | 18.42 | 4.74 | 26.74 | 23.76 | 23.81 | 13.04 | 10.56 | 6.59 | 38.09 |
| 105 | Belize | Caribbean | 17.79 | 19.24 | 28.31 | 16.64 | 13.19 | 11.59 | 4.56 | 0 | 50.18 |
| 106 | Antigua and Barbuda | Caribbean | 17.65 | 25.96 | 11.35 | 24.89 | 12.9 | 13.16 | 1.06 | 0 | 56.72 |
| 107 | Uganda | Africa | 17.01 | 4.36 | 28.24 | 9.15 | 33.23 | 13.59 | 5.62 | 4.18 | 46.41 |
| 108 | Bolivia (Plurinational State of) | South & Central America | 16.65 | 16.71 | 14.2 | 14.39 | 21.59 | 19.84 | 2.95 | 10.77 | 48.32 |
| 109 | Saint Lucia | Caribbean | 16.46 | 21.55 | 10.15 | 18.53 | 11.94 | 20.13 | 0.61 | 0 | 53.66 |
| 110 | Tunisia | Africa | 16.37 | 19.22 | 9.61 | 13.46 | 23.52 | 16.04 | 0 | 16.04 | 49.22 |
| 111 | Belarus | Europe | 15.64 | 16.67 | 11.43 | 25.53 | 23.06 | 9.25 | 2.78 | 20.15 | 45.01 |
| 113 | Zimbabwe | Africa | 15.18 | 15.19 | 16.77 | 14.6 | 16.49 | 12.83 | 1.21 | 15.38 | 43.04 |
| 114 | Sierra Leone | Africa | 14.91 | 13.53 | 19.22 | 10.3 | 14.8 | 16.72 | 1.21 | 10.26 | 43.87 |
| 115 | Lebanon | Middle East | 14.49 | 14.6 | 15.23 | 18.35 | 11.14 | 13.11 | 0.86 | 12.16 | 42.51 |
| 116 | Lao People's Democratic Republic | Asia & Oceania | 14.45 | 13.36 | 16.79 | 15.94 | 19.07 | 7.11 | 0.61 | 2.05 | 46.28 |
| 117 | Togo | Africa | 14.4 | 9.99 | 11.65 | 17.78 | 19.21 | 13.35 | 1.82 | 4.62 | 42.81 |
| 118 | Guinea | Africa | 13.72 | 11.75 | 15.1 | 10.48 | 18.51 | 12.76 | 0.61 | 0 | 44.52 |
| 119 | Cameroon | Africa | 13.53 | 15.18 | 14.41 | 12.95 | 14.46 | 10.66 | 0.61 | 2.05 | 43.21 |
| 120 | Mozambique | Africa | 13.49 | 10.95 | 16.16 | 11.28 | 19.31 | 9.77 | 3.18 | 3.3 | 37.51 |
| 121 | Burkina Faso | Africa | 13.11 | 3.24 | 7.35 | 10.33 | 29.48 | 15.15 | 2.86 | 16.06 | 32.62 |
| 122 | Liberia | Africa | 13.01 | 11.7 | 12.99 | 8.22 | 16.52 | 15.62 | 0 | 0 | 43.36 |
| 123 | Algeria | Africa | 12.84 | 10.61 | 5.61 | 17.06 | 19.26 | 11.68 | 1.21 | 0 | 40.38 |

| # | Country | Region | GIRAI Score | Inclusion & Diversity | Ethics & Sustain-ability | Labour & Skills | Trust & Safety | AI in Public Service | AI Policy | CSO Engage-ment | Enabling Condi-tions |
|---|---|---|---|---|---|---|---|---|---|---|---|
| 124 | Somalia | Africa | 12.05 | 6.36 | 11.73 | 5.34 | 28.92 | 7.88 | 9.6 | 0 | 20.95 |
| 125 | Niger | Africa | 12.01 | 13.18 | 5.02 | 9.2 | 24.63 | 8.02 | 3.67 | 2.93 | 31.72 |
| 126 | Myanmar | Asia & Oceania | 11.71 | 9.15 | 8.6 | 17.52 | 15.93 | 9.79 | 1.82 | 3.52 | 35.85 |
| 127 | Turkmenistan | Asia & Oceania | 11.31 | 18.43 | 2.86 | 18.46 | 9.48 | 7.32 | 1.21 | 3.08 | 34.26 |
| 128 | Democratic Republic of the Congo | Africa | 11.21 | 5.01 | 17.09 | 8.69 | 13.09 | 12.18 | 0 | 1.76 | 36.78 |
| 129 | Central African Republic | Africa | 9.44 | 3.21 | 18.31 | 10.61 | 6.71 | 8.35 | 0.61 | 8.21 | 27.51 |
| 130 | Haiti | Caribbean | 8.96 | 10.44 | 4.96 | 10.43 | 8.34 | 10.61 | 1.21 | 0 | 27.43 |
| 131 | Congo | Africa | 8.73 | 3.87 | 6.84 | 13.84 | 13.83 | 5.29 | 0.61 | 5.64 | 26.02 |
| 132 | Burundi | Africa | 8.7 | 1.61 | 14.59 | 12.27 | 7.45 | 7.6 | 0 | 4.1 | 27.64 |
| 133 | Chad | Africa | 6.51 | 0.82 | 3.56 | 5.7 | 17.3 | 5.18 | 0.61 | 0 | 20.49 |
| 134 | Afghanistan | Asia & Oceania | 6.02 | 3.69 | 14 | 4.23 | 4.74 | 3.45 | 0 | 0 | 20.08 |
| 135 | South Sudan | Africa | 4.04 | 0.19 | 1.11 | 5.4 | 5.07 | 8.41 | 0 | 2.05 | 12.77 |

## Figure A2: GIRAI Score Distribution by Region

Distribution of GIRAI scores across regions, showing median, interquartile range, and outliers for each. Regions are ordered from highest to lowest mean score.

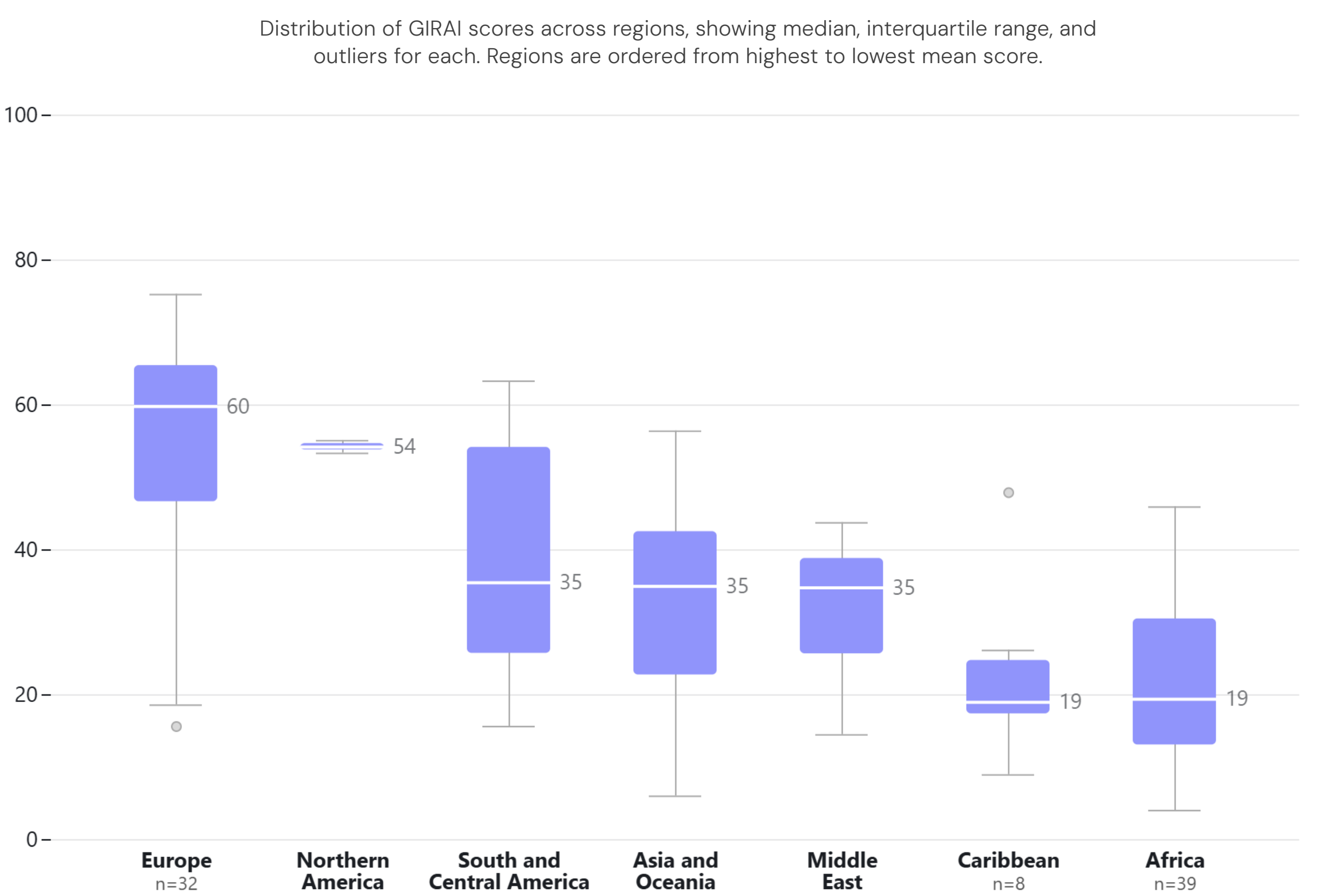

*Source: Global Index on Responsible AI (GIRAI) Global Survey, 2026.*

## Figure A3: Descriptive Statistics: GIRAI 2026 Score by Region

Summary statistics (mean, median, standard deviation, min, max, Q1, Q3, and IQR) of GIRAI scores for each region, based on the 135 countries assessed in the 2026 edition.

| Grouping | N countries | Mean | Median | Std Dev | Min | Q1 | Q3 | Max | IQR |
|---|---|---|---|---|---|---|---|---|---|
| Europe | 32 | 54.23 | 59.81 | 15.85 | 15.64 | 46.63 | 65.42 | 75.26 | 18.79 |
| Northern America | 2 | 54.22 | 54.22 | 1.24 | 53.34 | 53.78 | 54.65 | 55.09 | 0.88 |
| South and Central America | 14 | 38.85 | 35.48 | 17.15 | 15.64 | 25.7 | 54.13 | 63.3 | 28.43 |
| Asia and Oceania | 30 | 32.98 | 34.98 | 13.52 | 6.02 | 22.72 | 42.5 | 56.4 | 19.78 |
| Middle East | 10 | 32.07 | 34.78 | 9.57 | 14.49 | 25.64 | 38.8 | 43.75 | 13.15 |
| Caribbean | 8 | 22.42 | 18.98 | 11.55 | 8.96 | 17.35 | 24.72 | 47.92 | 7.37 |
| Africa | 39 | 21.79 | 19.41 | 11.12 | 4.04 | 13.06 | 30.45 | 45.93 | 17.39 |

*Source: Global Index on Responsible AI (GIRAI) Global Survey, 2026.*

## Figure A4: Average GIRAI 2026 Score by Dimension (95% CI)

Average scores across all 135 countries for each of the five GIRAI dimensions, with 95% confidence intervals reflecting cross-country variation within each dimension.

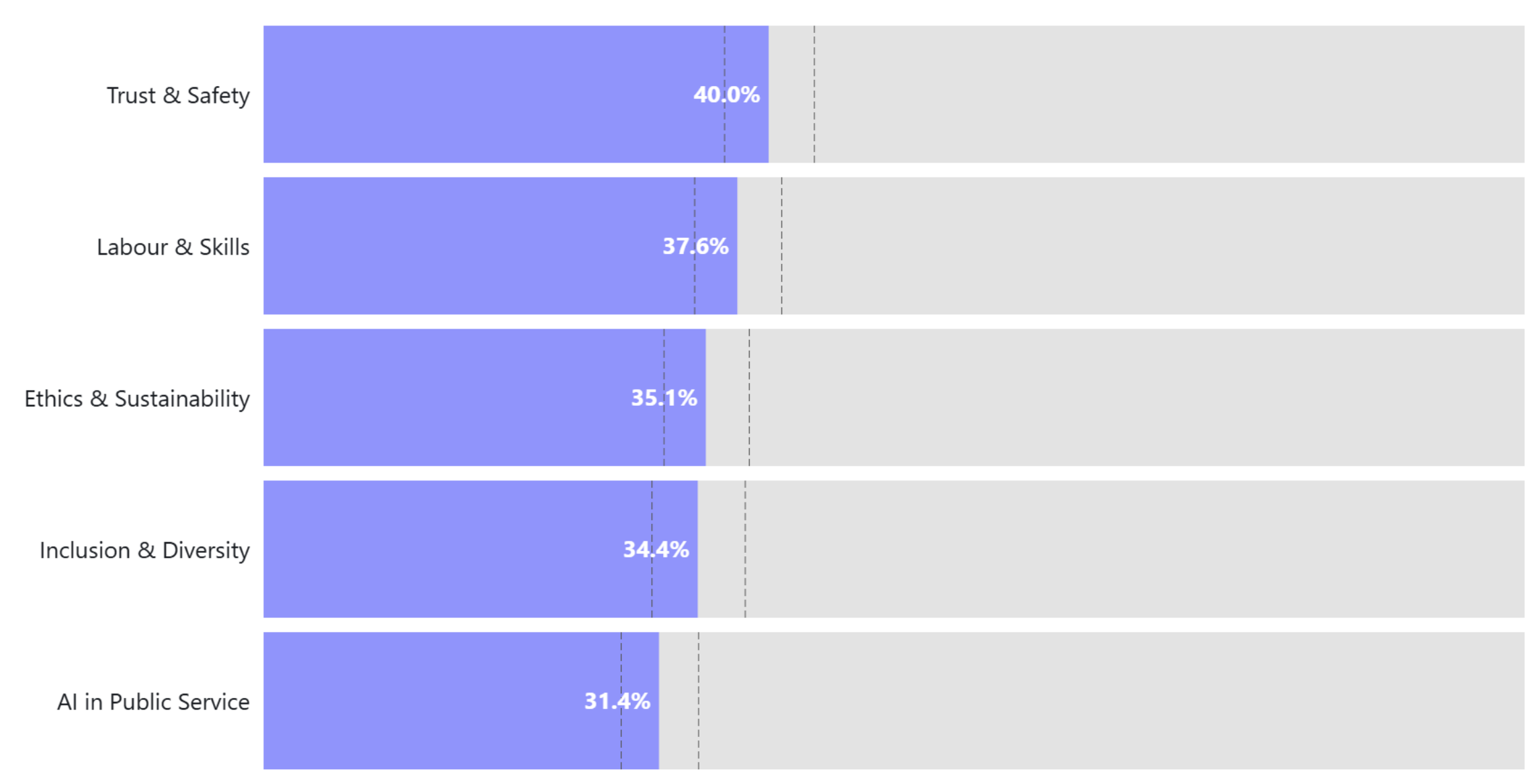

*Source: Global Index on Responsible AI (GIRAI) Global Survey, 2026.*

## Figure A5: Average GIRAI 2026 Score by Dimension and Region

Average scores for each of the five GIRAI dimensions across all regions, including a Global Average row. Darker shading indicates higher average scores.

| Region | Inclusion & Diversity | Ethics & Sustainability | Labour & Skills | Trust & Safety | AI in Public Service |
|---|---|---|---|---|---|
| Europe | 56.44 | 54.81 | 57.98 | 63.43 | 44.61 |
| Northern America | 57.59 | 51.88 | 54.76 | 56.16 | 66.46 |
| South and Central America | 39.61 | 40 | 40.26 | 40.04 | 38.5 |
| Asia and Oceania | 33.21 | 31 | 36.03 | 38.45 | 30.02 |
| Middle East | 32.56 | 31.58 | 34.73 | 36.9 | 28.75 |
| Africa | 16.68 | 23.38 | 23.3 | 26.36 | 19.73 |
| Caribbean | 25.07 | 19.89 | 25.8 | 19.23 | 22.1 |
| **Global Average** | **34.43** | **35.07** | **37.57** | **40.05** | **31.36** |

*Source: Global Index on Responsible AI (GIRAI) Global Survey, 2026.*

## Figure A6: Average Scores by AI Policy Indicator, Overall and by Region

Average scores for each of the 17 AI Policy indicators, disaggregated by region.
Indicators are ranked from highest to lowest overall score.

| Indicator | Africa | Asia and Oceania | Caribbean | Europe | Middle East | Northern America | South and Central America | Global Average |
|---|---|---|---|---|---|---|---|---|
| Access to Redress and Remedy | 20.70 | 23.20 | 7.91 | 54.97 | 21.96 | 34.29 | 14.92 | **28.32** |
| AI Literacy | 27.82 | 39.00 | 12.93 | 52.26 | 29.76 | 34.85 | 38.55 | **36.58** |
| AI-facilitated Misinformation and Violence | 6.59 | 25.50 | 13.71 | 58.39 | 12.01 | 52.56 | 26.94 | **26.68** |
| Children's Rights | 8.54 | 27.78 | 0.00 | 59.03 | 15.80 | 50.76 | 31.09 | **27.78** |
| Cultural and Linguistic Diversity | 12.61 | 25.91 | 9.04 | 41.79 | 30.00 | 24.62 | 21.89 | **24.70** |
| Environmental Impact | 7.12 | 17.95 | 8.95 | 28.51 | 13.65 | 23.16 | 25.18 | **17.30** |
| Fairness and Non-discrimination | 21.48 | 26.05 | 16.86 | 60.09 | 36.25 | 41.18 | 38.32 | **34.51** |
| Gender Equality | 8.64 | 24.51 | 11.41 | 51.14 | 25.23 | 62.62 | 40.39 | **27.73** |
| Human Oversight and Determination | 24.16 | 26.23 | 16.76 | 57.82 | 36.67 | 62.71 | 31.08 | **34.38** |
| Impact Assessments | 12.17 | 27.88 | 1.42 | 60.45 | 35.68 | 54.10 | 40.16 | **31.73** |
| Labour Protections | 2.84 | 7.70 | 2.27 | 53.88 | 6.25 | 23.87 | 29.68 | **19.33** |
| Public Disclosure of Government Algorithmic Systems | 0.00 | 11.14 | 0.00 | 21.52 | 4.09 | 69.43 | 22.97 | **11.29** |
| Public Procurement | 6.48 | 14.75 | 10.37 | 25.29 | 17.44 | 74.75 | 17.74 | **16.00** |
| Public Sector Skills Development | 19.67 | 31.99 | 13.78 | 41.99 | 37.52 | 47.33 | 34.76 | **30.64** |
| Reskilling and upskilling initiatives | 25.38 | 41.43 | 21.40 | 56.04 | 49.43 | 75.57 | 36.07 | **39.61** |
| Safety and Security | 14.25 | 33.77 | 11.65 | 60.69 | 38.15 | 39.26 | 37.61 | **34.00** |
| Transparency and Explainability | 18.11 | 33.51 | 19.12 | 64.50 | 40.96 | 48.59 | 41.34 | **37.14** |
| **Average** | **13.92** | **25.78** | **10.45** | **49.90** | **26.52** | **48.21** | **31.10** | **28.10** |

*Source: Global Index on Responsible AI (GIRAI) Global Survey, 2026.*

Figure A7: **Global AI Policy Framework Coverage Summary**

Overall framework coverage rate and number of countries with no adopted AI policy framework, across the 135 countries assessed. Framework coverage is defined as the share of AI Policy indicators covered by at least one adopted framework, excluding drafts.

42%
Framework coverage

31
Countries with no framework
(23%)

Figure A8: **AI Policy Framework Coverage by Region**

Average percentage of AI Policy indicators covered by at least one adopted framework, by region.

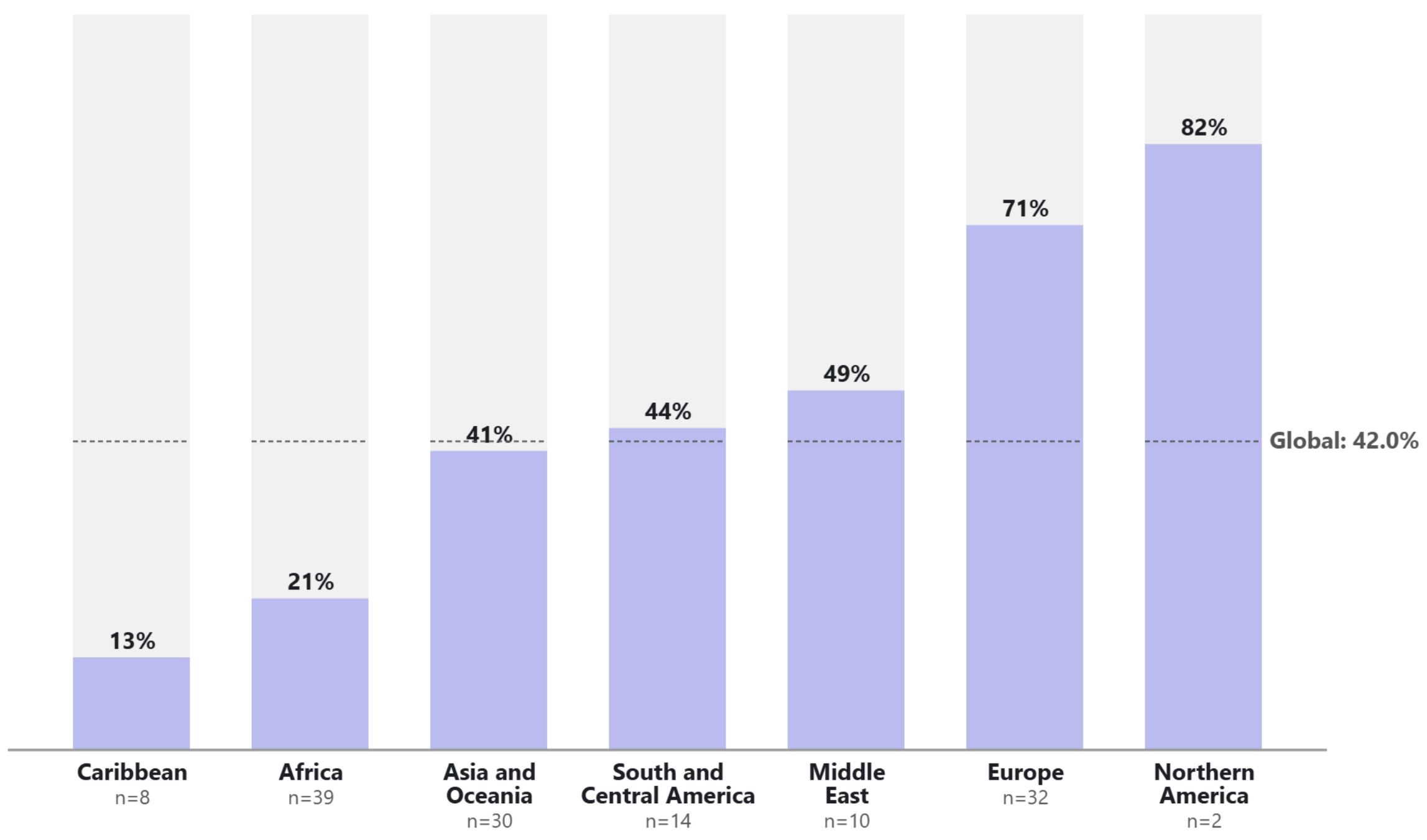

*Source: Global Index on Responsible AI (GIRAI) Global Survey, 2026.*

## Figure A9: Framework Coverage Across AI Policy Indicators, by Region and Enforceability

Share of AI Policy indicators covered by binding frameworks, non-binding frameworks, or no framework, by region.

| Region | Binding coverage | Non-binding coverage | No framework |
|---|---|---|---|
| Africa | 4.68% | 15.84% | 79.49% |
| Asia and Oceania | 8.24% | 32.35% | 59.41% |
| Caribbean | 2.21% | 10.29% | 87.50% |
| Europe | 45.77% | 25.55% | 28.68% |
| Middle East | 2.94% | 45.88% | 51.18% |
| Northern America | 55.88% | 26.47% | 17.65% |
| South and Central America | 14.71% | 28.99% | 56.30% |

*Source: Global Index on Responsible AI (GIRAI) Global Survey, 2026.*

## Figure A10: Framework Coverage by Dimension and Region

Average scores for each of the five GIRAI dimensions across all regions, including a Global Coverage row. Darker shading indicates higher average scores.

| Region | AI Use in Public Service | Ethics and Sustainability | Inclusion and Diversity | Labour and Skills | Trust and Safety |
|---|---|---|---|---|---|
| Africa | 12.82% | 28.21% | 14.53% | 24.79% | 19.87% |
| Asia and Oceania | 33.33% | 43.33% | 40.00% | 44.44% | 40.83% |
| Caribbean | 8.33% | 21.88% | 8.33% | 12.50% | 9.38% |
| Europe | 46.88% | 75.00% | 70.83% | 78.13% | 81.25% |
| Middle East | 40.00% | 62.50% | 40.00% | 40.00% | 55.00% |
| Northern America | 100.00% | 75.00% | 66.67% | 66.67% | 100.00% |
| South and Central America | 35.71% | 48.21% | 42.86% | 45.24% | 44.64% |
| **Global Coverage** | **30.86%** | **47.59%** | **38.77%** | **44.94%** | **44.81%** |

*Source: Global Index on Responsible AI (GIRAI) Global Survey, 2026.*

## Figure A11: Framework Coverage Across AI Policy Indicators, Overall and by Region

Percentage of countries with at least one adopted framework covering each of the 17 AI Policy indicators, overall and by region. Indicators are ranked from highest to lowest overall coverage.

| Indicator | Africa | Asia and Oceania | Caribbean | Europe | Middle East | Northern America | South and Central America | Global Average |
|---|---|---|---|---|---|---|---|---|
| Access to Redress and Remedy | 33.33% | 30.00% | 12.50% | 78.13% | 50.00% | 100.00% | 21.43% | **42.96%** |
| AI Literacy | 35.90% | 63.33% | 12.50% | 78.13% | 40.00% | 50.00% | 50.00% | **52.59%** |
| AI-facilitated Misinformation and Violence | 7.69% | 36.67% | 12.50% | 75.00% | 20.00% | 100.00% | 42.86% | **36.30%** |
| Children's Rights | 15.38% | 46.67% | 0.00% | 78.13% | 30.00% | 50.00% | 42.86% | **40.74%** |
| Cultural and Linguistic Diversity | 15.38% | 36.67% | 12.50% | 59.38% | 50.00% | 50.00% | 28.57% | **34.81%** |
| Environmental Impact | 10.26% | 30.00% | 12.50% | 40.63% | 30.00% | 50.00% | 35.71% | **26.67%** |
| Fairness and Non-discrimination | 33.33% | 43.33% | 25.00% | 84.38% | 70.00% | 50.00% | 57.14% | **52.59%** |
| Gender Equality | 12.82% | 36.67% | 12.50% | 75.00% | 40.00% | 100.00% | 57.14% | **40.74%** |
| Human Oversight and Determination | 38.46% | 43.33% | 25.00% | 84.38% | 70.00% | 100.00% | 42.86% | **53.33%** |
| Impact Assessments | 17.95% | 43.33% | 0.00% | 87.50% | 80.00% | 100.00% | 57.14% | **48.89%** |
| Labour Protections | 5.13% | 13.33% | 0.00% | 78.13% | 10.00% | 50.00% | 42.86% | **28.89%** |
| Public Disclosure of Government Algorithmic Systems | 0.00% | 20.00% | 0.00% | 31.25% | 10.00% | 100.00% | 35.71% | **17.78%** |
| Public Procurement | 10.26% | 30.00% | 12.50% | 37.50% | 30.00% | 100.00% | 28.57% | **25.93%** |
| Public Sector Skills Development | 28.21% | 50.00% | 12.50% | 71.88% | 80.00% | 100.00% | 42.86% | **48.89%** |
| Reskilling and upskilling initiatives | 33.33% | 56.67% | 25.00% | 78.13% | 70.00% | 100.00% | 42.86% | **53.33%** |
| Safety and Security | 20.51% | 53.33% | 12.50% | 84.38% | 70.00% | 100.00% | 57.14% | **51.11%** |
| Transparency and Explainability | 30.77% | 56.67% | 25.00% | 90.63% | 80.00% | 100.00% | 57.14% | **57.78%** |
| **Average** | **20.51%** | **40.59%** | **12.50%** | **71.32%** | **48.82%** | **82.35%** | **43.70%** | **41.96%** |

*Source: Global Index on Responsible AI (GIRAI) Global Survey, 2026.*

Figure A12: **Global Government-Led Initiative Coverage Summary**

Average share of AI Policy indicators addressed by at least one government-led initiative, and number of countries with no evidence on government-led initiatives

33%
Initiative coverage

11
Countries with no initiatives
(8%)

Figure A13: **Government-Led Initiative Coverage by Region**

Average share of AI Policy indicators accompanied by at least one government-led initiative, by region and global.

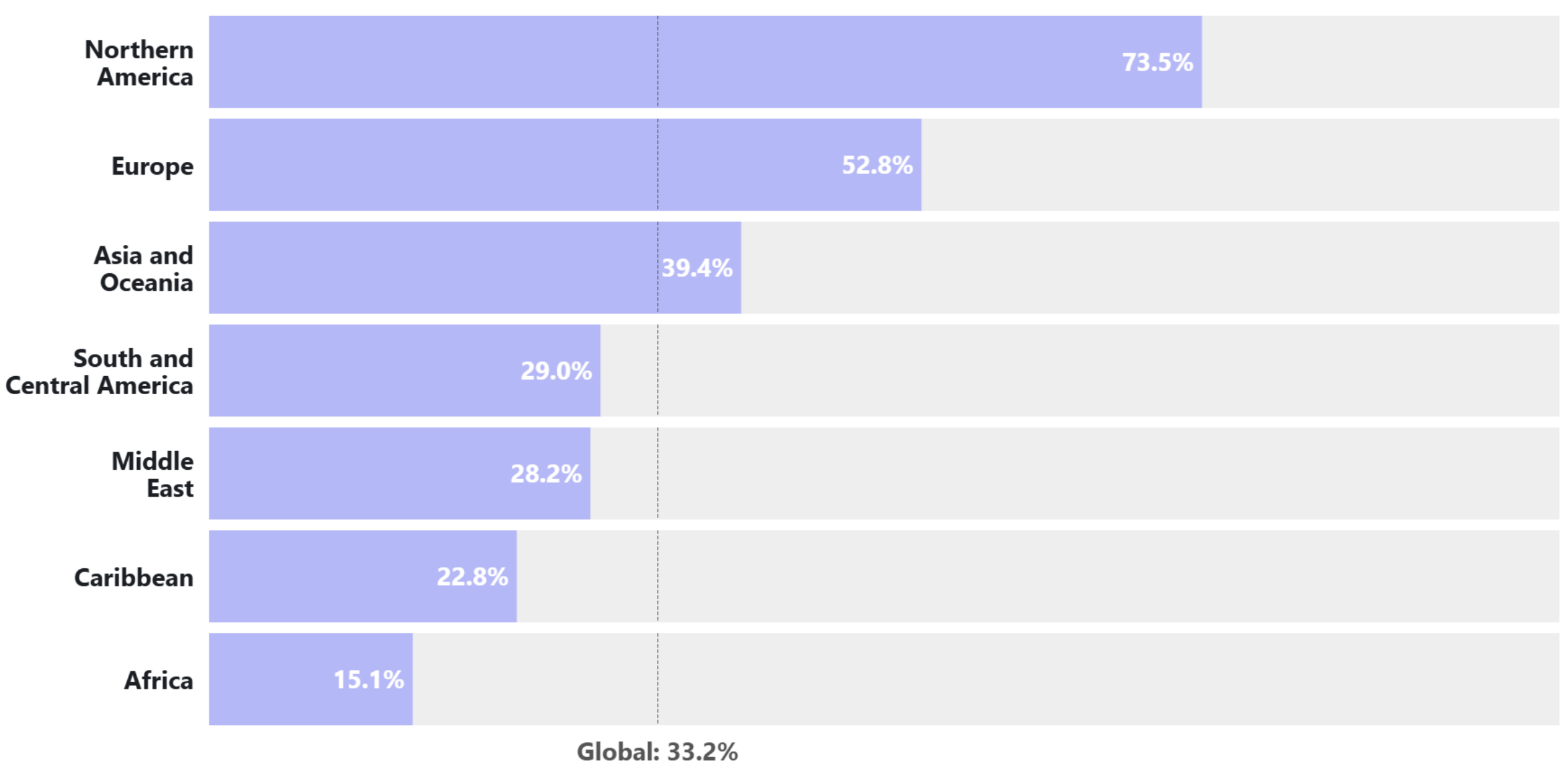

*Source: Global Index on Responsible AI (GIRAI) Global Survey, 2026.*

### Figure A14: Government-Led Initiative Coverage by Dimension and Region

Share of AI Policy indicators accompanied by at least one government-led initiative, disaggregated by GIRAI dimension and region. Darker shading indicates higher coverage.

| Dimension | AI Use in Public Service | Ethics and Sustainability | Inclusion and Diversity | Labour and Skills | Trust and Safety |
|---|---|---|---|---|---|
| Africa | 9.40% | 9.62% | 12.82% | 31.62% | 14.10% |
| Asia and Oceania | 36.67% | 30.00% | 47.78% | 51.11% | 35.83% |
| Caribbean | 25.00% | 18.75% | 12.50% | 41.67% | 18.75% |
| Europe | 46.88% | 46.09% | 48.96% | 63.54% | 58.59% |
| Middle East | 26.67% | 27.50% | 30.00% | 36.67% | 22.50% |
| Northern America | 100.00% | 87.50% | 50.00% | 66.67% | 62.50% |
| South and Central America | 35.71% | 25.00% | 23.81% | 57.14% | 10.71% |
| **Global Coverage** | **30.62%** | **27.41%** | **32.10%** | **47.65%** | **30.74%** |

*Source: Global Index on Responsible AI (GIRAI) Global Survey, 2026.*

### Figure A15: Government-Led Initiative Coverage Across AI Policy Indicators, Overall and by Region

Percentage of countries with at least one government-led initiative for each of the 17 AI Policy indicators, overall and by region. Indicators are ranked from highest to lowest overall coverage.

| indicator | Africa | Asia and Oceania | Caribbean | Europe | Middle East | Northern America | South and Central America | Global Average |
|---|---|---|---|---|---|---|---|---|
| Access to Redress and Remedy | 12.82% | 30.00% | 0.00% | 50.00% | 20.00% | 50.00% | 0.00% | **24.44%** |
| AI Literacy | 53.85% | 70.00% | 50.00% | 75.00% | 50.00% | 50.00% | 71.43% | **63.70%** |
| AI-facilitated Misinformation and Violence | 17.95% | 33.33% | 50.00% | 59.38% | 30.00% | 50.00% | 21.43% | **34.81%** |
| Children's Rights | 7.69% | 43.33% | 0.00% | 46.88% | 20.00% | 50.00% | 21.43% | **27.41%** |
| Cultural and Linguistic Diversity | 17.95% | 53.33% | 12.50% | 65.63% | 30.00% | 50.00% | 21.43% | **38.52%** |
| Environmental Impact | 5.13% | 30.00% | 12.50% | 37.50% | 20.00% | 100.00% | 7.14% | **21.48%** |
| Fairness and Non-discrimination | 12.82% | 36.67% | 12.50% | 59.38% | 30.00% | 100.00% | 35.71% | **34.07%** |
| Gender Equality | 12.82% | 46.67% | 25.00% | 34.38% | 40.00% | 50.00% | 28.57% | **30.37%** |
| Human Oversight and Determination | 12.82% | 30.00% | 25.00% | 37.50% | 30.00% | 50.00% | 21.43% | **25.93%** |
| Impact Assessments | 5.13% | 36.67% | 0.00% | 59.38% | 20.00% | 100.00% | 14.29% | **28.15%** |
| Labour Protections | 2.56% | 20.00% | 25.00% | 43.75% | 0.00% | 50.00% | 42.86% | **22.22%** |
| Public Disclosure of Government Algorithmic Systems | 0.00% | 16.67% | 0.00% | 34.38% | 0.00% | 100.00% | 28.57% | **16.30%** |
| Public Procurement | 0.00% | 16.67% | 25.00% | 34.38% | 20.00% | 100.00% | 14.29% | **17.78%** |
| Public Sector Skills Development | 28.21% | 76.67% | 50.00% | 71.88% | 60.00% | 100.00% | 64.29% | **57.78%** |
| Reskilling and upskilling initiatives | 38.46% | 63.33% | 50.00% | 71.88% | 60.00% | 100.00% | 57.14% | **57.04%** |
| Safety and Security | 20.51% | 43.33% | 25.00% | 65.63% | 20.00% | 50.00% | 7.14% | **35.56%** |
| Transparency and Explainability | 7.69% | 23.33% | 25.00% | 50.00% | 30.00% | 100.00% | 35.71% | **28.15%** |
| **Average** | **15.08%** | **39.41%** | **22.79%** | **52.76%** | **28.24%** | **73.53%** | **28.99%** | **33.16%** |

*Source: Global Index on Responsible AI (GIRAI) Global Survey, 2026.*

Figure A16: **From Policy Frameworks to Implementation: Global Average**

Global average coverage at three levels: adopted framework (L1), L1 plus framework with at least one operational provision (L2), and L2 plus at least one government-led initiative (L3), with percentage-point drops between levels.

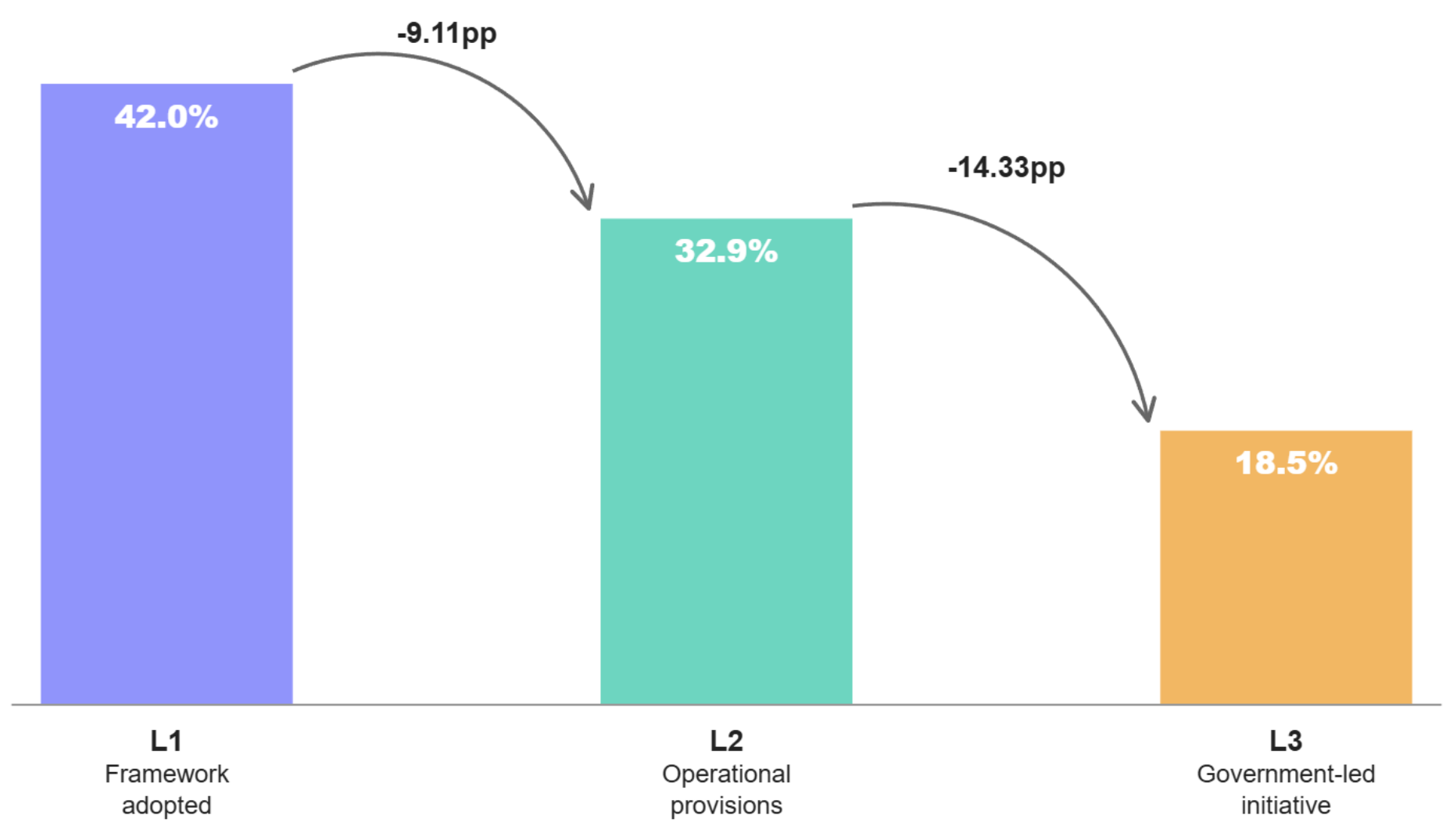

*Source: Global Index on Responsible AI (GIRAI) Global Survey, 2026.*

Figure A17: **From Policy Frameworks to Implementation, by Dimension**

Coverage at three levels — adopted framework (L1), L1 plus framework with operational provisions (L2), and L2 plus government-led initiative (L3) — for each of the five GIRAI dimensions.

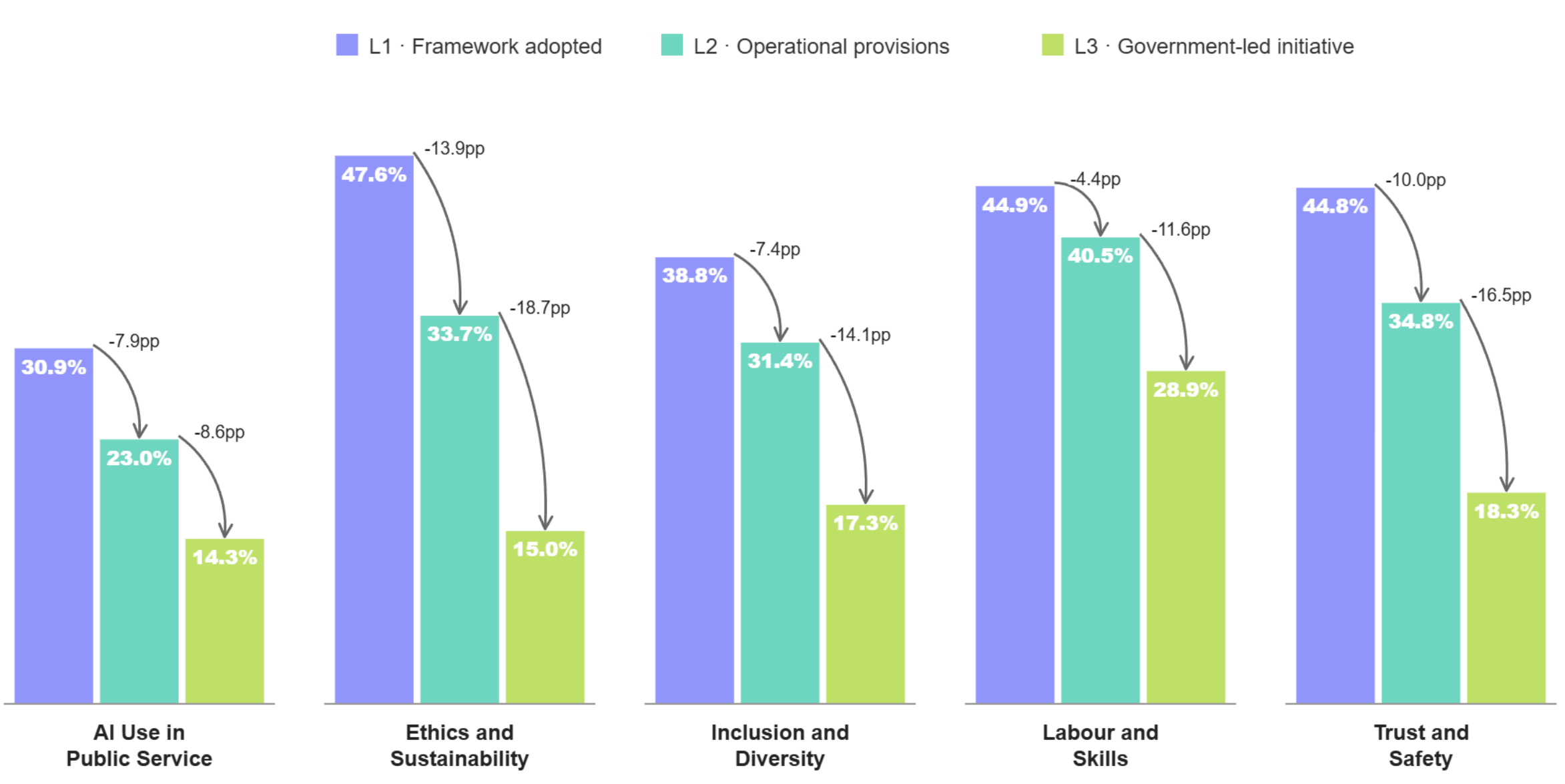

*Source: Global Index on Responsible AI (GIRAI) Global Survey, 2026.*

## Figure A18: From Policy Frameworks to Implementation, by Region

Coverage at three levels — adopted framework (L1), L1 plus framework with operational provisions (L2), and L2 plus government-led initiative (L3) — for each region.

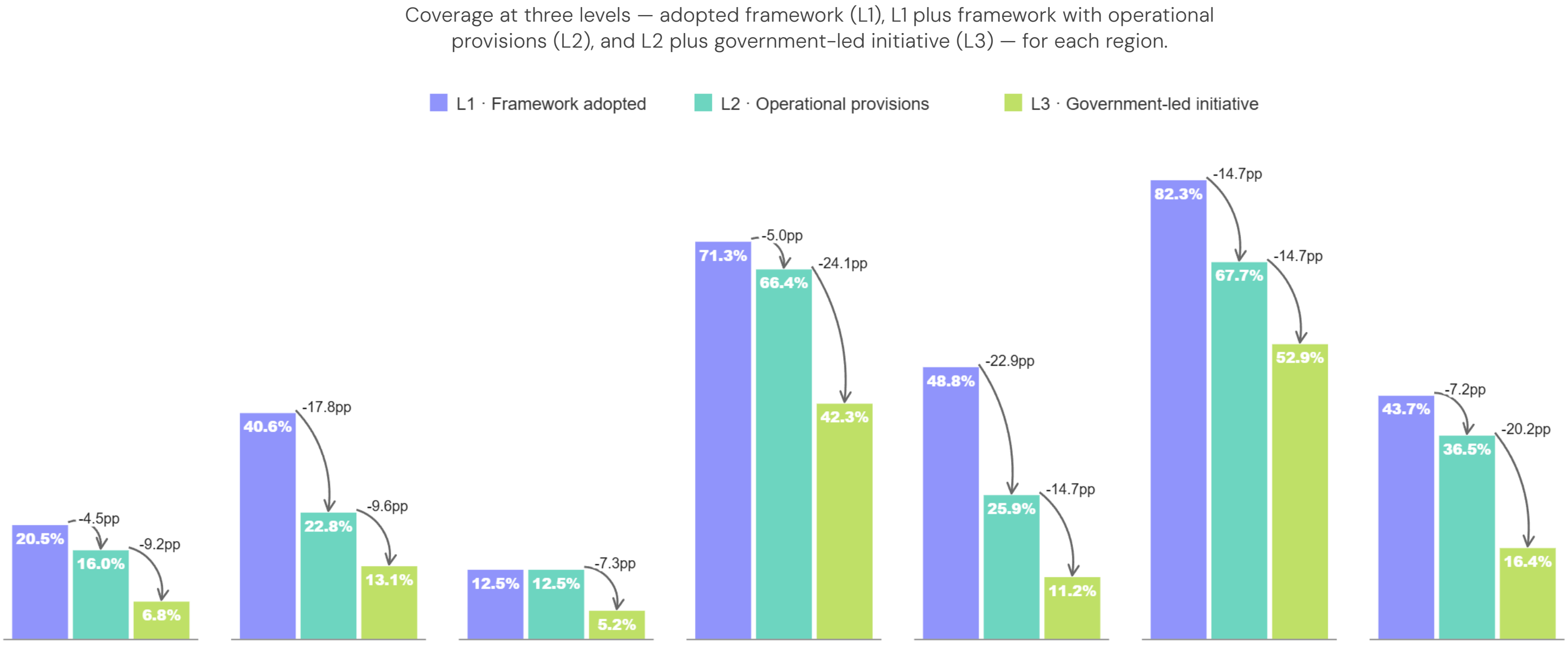

*Source: Global Index on Responsible AI (GIRAI) Global Survey, 2026.*

## Figure A19: Distribution of Civil Society Engagement Activities by Dimension

Share of total civil society engagement (CSE) activities across the four GIRAI dimensions covered by the CSE module, globally.

| Dimension | CSE Counts | % of CSE activities by dimension |
|---|---|---|
| Ethics and Sustainability | 169 | 25.11% |
| Inclusion and Diversity | 159 | 23.63% |
| Labour and Skills | 165 | 24.52% |
| Trust and Safety | 180 | 26.75% |
| **Grand Total** | **673** | **100.00%** |

*Source: Global Index on Responsible AI (GIRAI) Global Survey, 2026.*

# List of Figures

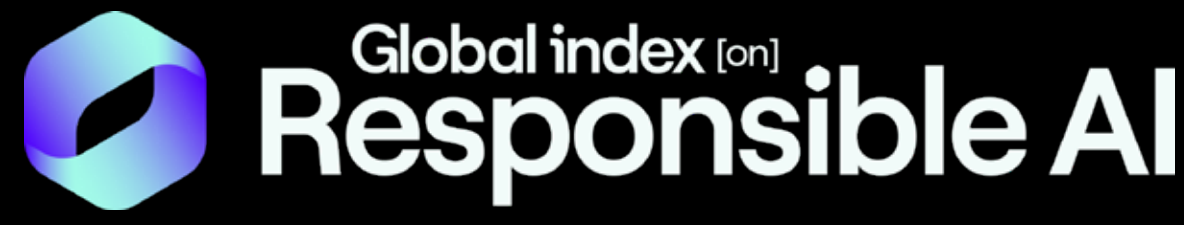

For more information, visit https://global-index.ai to explore the full dataset and all evidence assessed across the 135 countries included in this edition. All data are freely available for download in open data format. For enquiries, please contact us at girai_gcg@globalcenter.ai.